\pdfoutput=1
\documentclass{article}

\usepackage{template}

\usepackage[utf8]{inputenc} 
\usepackage[T1]{fontenc}    
\usepackage{hyperref}       
\usepackage{url}            
\usepackage{booktabs}       
\usepackage{amsfonts}       
\usepackage{nicefrac}       
\usepackage{microtype}      
\usepackage{lipsum}		
\usepackage{graphicx}
\usepackage{doi}
\usepackage{bbm}
\usepackage{amssymb}
\usepackage{amsmath}
\usepackage{hyperref}

\usepackage{algorithm}
\usepackage{algorithmic}

\DeclareMathOperator*{\argmax}{arg\,max}

\title{Reinforcement Learning and Bandits for Speech and Language Processing: Tutorial, Review and Outlook} 

\author{Baihan Lin\\
	Columbia University\\
	New York, NY 10027 \\
	\texttt{baihan.lin@columbia.edu} \\
}

\renewcommand{\headeright}{}
\renewcommand{\undertitle}{A Gentle Introduction}

\hypersetup{
pdftitle={Reinforcement Learning and Bandits for Language and Speech: Tutorial, Review and Perspectives},
pdfsubject={cs.AI, cs.LG, cs.CL},
pdfauthor={Baihan Lin},
pdfkeywords={Reinforcement learning, Bandits, Speech processing, Natural language processing, Survey, Perspective},
}

\begin{document}
\maketitle

\begin{abstract}

In recent years, reinforcement learning and bandits have transformed a wide range of real-world applications including healthcare, finance, recommendation systems, robotics, and last but not least, the speech and natural language processing. While most speech and language applications of reinforcement learning algorithms are centered around improving the training of deep neural networks with its flexible optimization properties, there are still many grounds to explore to utilize the benefits of reinforcement learning, such as its reward-driven adaptability, state representations, temporal structures and generalizability. In this survey, we present an overview of recent advancements of reinforcement learning and bandits, and discuss how they can be effectively employed to solve speech and natural language processing problems with models that are adaptive, interactive and scalable. 

\end{abstract}

\keywords{Reinforcement Learning \and Bandits \and Speech Processing \and Natural Language Processing \and Survey \and Perspective}

\section{Introduction}

As two cornerstones of modern day technologies, speech processing and natural language processing (NLP) are innately sequence learning problems to extract information from these linguistic or speech signals and provide insights into interactive systems to communicate in human understandable languages. The sequential and interactive nature of these problems can make them well-suited into the algorithmic framework of reinforcement learning (RL). In a reinforcement learning setting, an agent interacts with an environment through observations and actions, and based on the reward feedback attributed by the underlying reward function of this environment, the agent learns how to perform the task of interest through trials and errors. While the successful applications of reinforcement learning have been highlighted by a wide range of surveys in many real-world engineering domains such as robotics \cite{kober2013reinforcement}, vision \cite{le2021deep}, finance \cite{fischer2018reinforcement}, healthcare \cite{yu2021reinforcement}, linguistics \cite{uc2023survey}, and energy management \cite{zhang2019deep}, there have not been one for the rich community of both the speech and language domains. This is the first survey that emphasizes the synergy among the growing fields of the speech processing, natural language processing and the reinforcement learning. We aim to fill this gap by adopting a complete, timely and classical view of the reinforcement learning problems and their connections to speech and language processing. 

This survey distinguishes itself from previous ones by introducing, for the first time, the application of bandits and online learning to speech processing tasks. While deep reinforcement learning has gained popularity for reward-driven policy optimization in neural network architectures, other speech or language domains may be better suited to lightweight representations in the bandits and online learning framework. Additionally, over the past two years, there have been significant developments in different variants of reinforcement learning scenarios, such as inverse reinforcement learning, offline reinforcement learning, imitation learning, and behavioral cloning. Despite the successful applications outside the speech domain, these new problems can benefit from more exposure to the speech and language communities due to their unique perspectives and advantages. This survey provides a comprehensive overview of these novel problem settings as well as their state-of-the-art solutions (including the latest work in the field of large language models).

As this interdisciplinary field is still a fast growing research area at its early stages, the intended target audience for this survey are not just new researchers in the field, but also specialists of the speech and language domains that hopefully can gain insights and inspirations from the recent advances in reinforcement learning.
Accompanying the \textit{INTERSPEECH 2022} tutorial ``Reinforcement Learning and Bandits for Speech and Language Processing'',
the goal of this survey is not to provide an exhaustive review of the fields, but to first provide an applied tutorial of the methodologies in reinforcement learning, and then guide the readers through a series of prototypical examples of how reinforcement learning can be effectively applied to major speech and language tasks. We hope these case studies motivate the readers to rethink their daily tasks as reinforcement learning problems and encourage new discussions.

\tableofcontents

\textbf{Preview}. The survey is organized in the following way: First, we briefly introduce the basic concept of reinforcement learning and bandits, as well as the major variant problem settings in this machine learning domain. Second, we translate various speech and language tasks into the reinforcement learning problems and show the key challenges. Third, we introduce some reinforcement learning and bandit techniques and their varieties for speech and language tasks and their machine learning formulations. Fourth, we present several state-of-the-art applications of reinforcement learning in different fields of speech and language. Lastly, we will discuss some open problems in reinforcement learning and bandits to show how to further develop more advanced algorithms for speech and language research in the future. 

\section{Why do we want reinforcement learning in speech and language processing}

\begin{figure}[tb]
\centering
    \includegraphics[width=\linewidth]{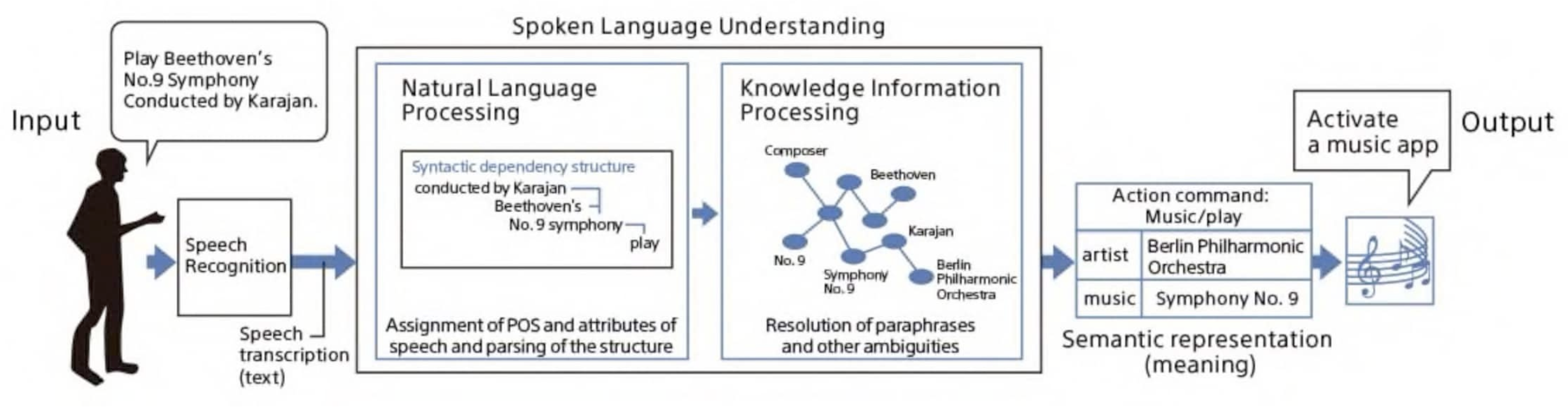}
\vspace{-1em}
\caption{Example of a speech and language processing system
}\label{fig:example}
\end{figure}

\begin{figure}[b]
\centering
    \includegraphics[width=\linewidth]{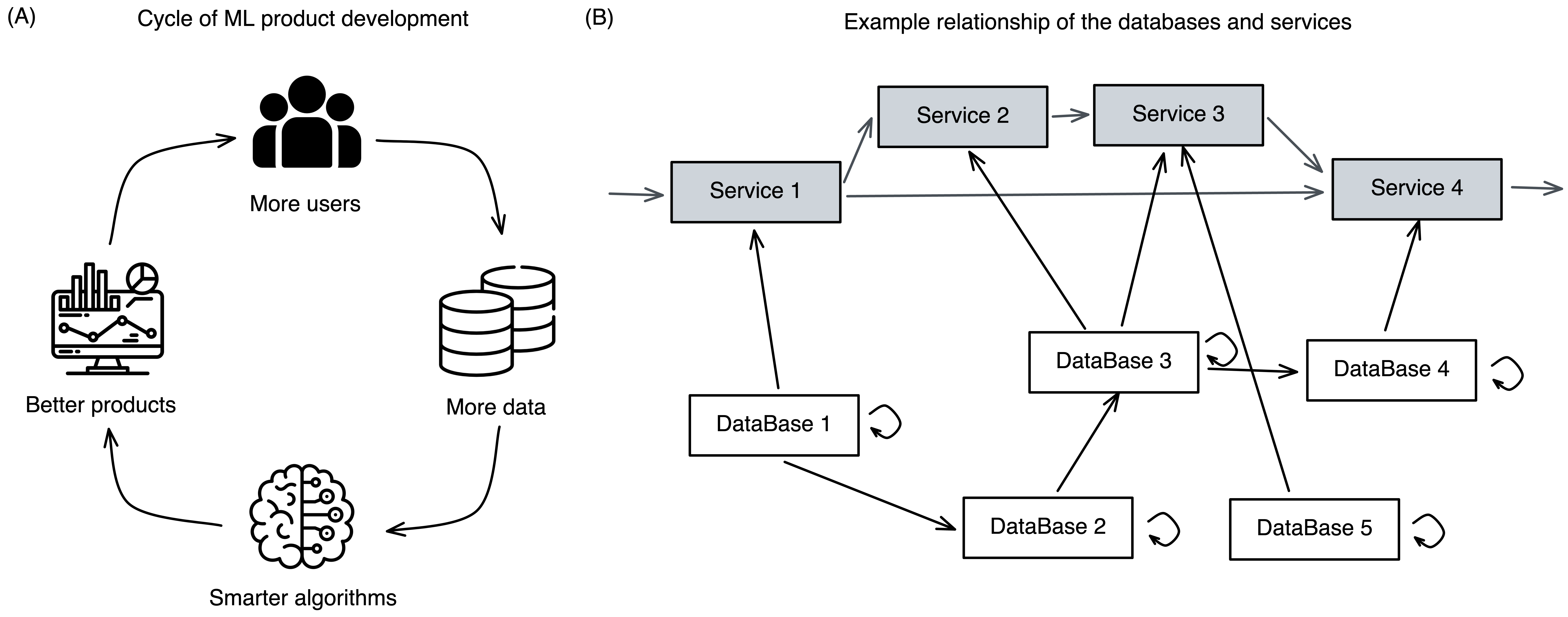}
\vspace{-1em}
\caption{(A) Cycle of the machine learning-based product development. (B) Example relationships among multiple services and multiple databases, each with their own update schedules or frequencies.
}\label{fig:cycle}
\end{figure}

A speech and natural language processing system usually involves many components. For instance, Figure \ref{fig:example} is an example of a voice command system to play music. It starts with a speech recognition engine to transcribe the speech into text. A speech language understanding components first uses natural language processing techniques to parse out the semantic structure, its intents (e.g. command actions) and then utilizes available knowledge graphs to extract machine understandable symbolic relationships for downstream information retrieval. Then the system will locate the entry in the databases that most closely matches the entity of query. Finally, the music app is activated playing the entry. These components are often data-rich machine learning models or data analytical tools pretrained on existing data. The iterative process of such a speech or language-based system is a cycle from more users to more data to smarter algorithms to better products and finally back to more users (Figure \ref{fig:cycle}A).

From a industrial product point of view, these system components (which are sometimes called services) interact with the underlying databases in various ways. Figure \ref{fig:cycle}B shows an example relationship between the services and the databases they train on. One service might depend on the information flow from another or several other services. One database might be accessed by one or more services, and also at the same time linked with another databases (e.g. in relational databases). These databases are usually updated in different frequencies. For instance, one database might updates regularly every 6 hours, while another database updates only sporadically whenever a new user signs up. One database might update its record everytime another database updates. Since we train our service models on these databases, one challenge of model training for different services is to determine when or how often they are trained given the updates in the databases. One might wish to train the service model everytime the available pool of historical data is updated, so as to get the best performing model. On the other hand, given the large size of the historical data and model parameters in industrial setting, iteratively retraining all the models at each database update would be too computationally expensive and environmentally irresponsible. As a result, we want our models to ideally be able to learn incrementally, and strategically given its performance or metric. Such a model would ideally deal with the uncertainty in the data: exploit what we have learned so far to make the best decision in deployment, while also exploring enough rounds in relatively unfamiliar knowledge domains to gain better understanding of all possible actions it can take. This strategy of dealing with uncertainty is usually called the \textit{Exploration vs. Exploitation dilemma tradeoff}.

\begin{figure}[tb]
\centering
    \includegraphics[width=\linewidth]{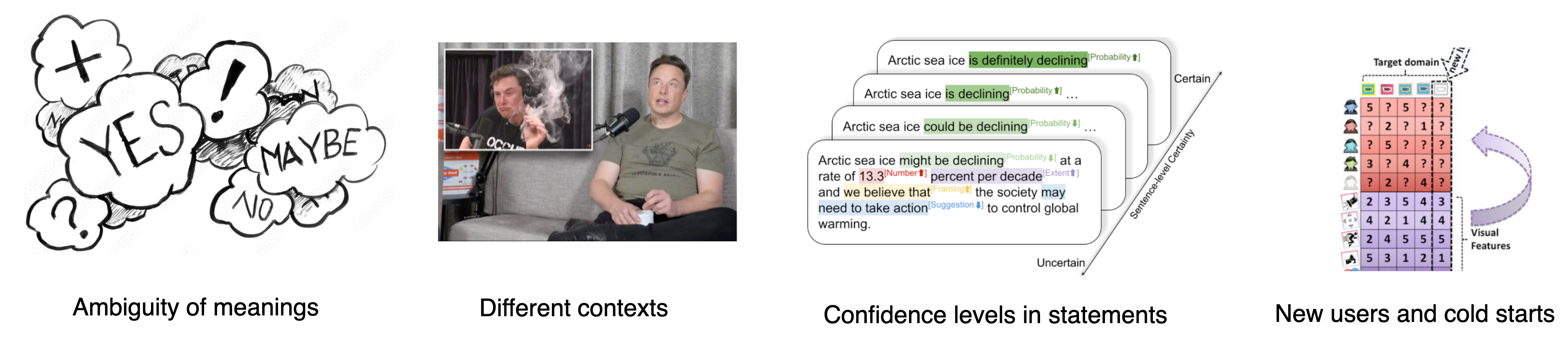}
\vspace{-1em}
\caption{Data uncertainty in speech and language processing.
}\label{fig:uncertainty}
\end{figure}

\begin{figure}[tb]
\centering
    \includegraphics[width=\linewidth]{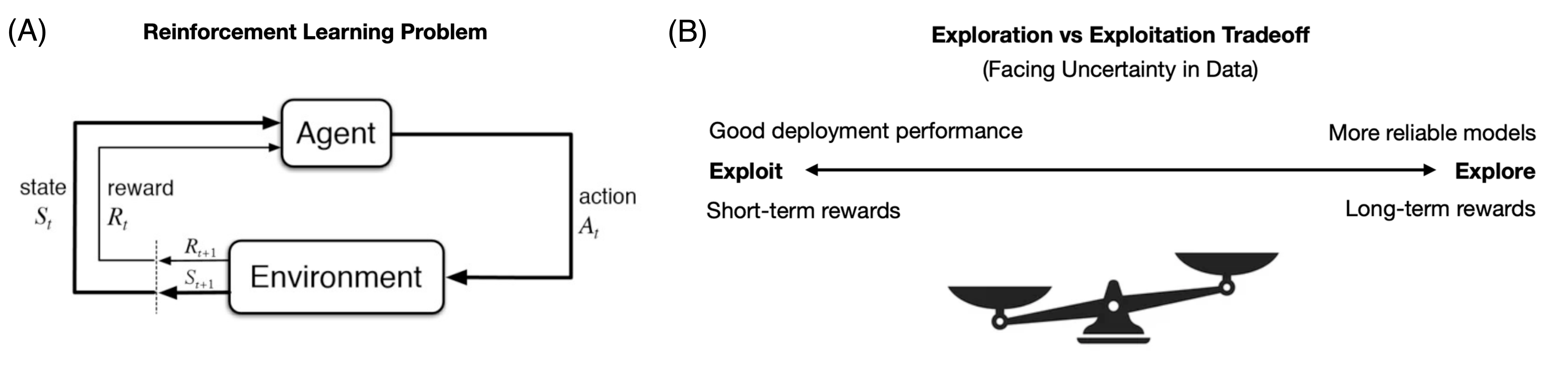}
\vspace{-1em}
\caption{(A) The reinforcement learning problem. (B) The exploration vs exploitation tradeoff.
}\label{fig:rlproblem}
\end{figure}

\textit{Uncertainty} is everywhere in speech and language domains. Figure \ref{fig:uncertainty} provides a few examples. For instance, the same words or sentences can have ambiguious meanings. The speech properties and language patterns can vary significantly when the contexts are different (e.g. a podcast interview vs. a commencement speech). The same statements can also have different levels of confidence, depending on their contexts and wording. New users or data streams might introduce distributional shifts and cold start issues. Other than innate noises or uncertainty in the data, there are also systematic uncertainty in machine learning systems. For instance, the model might not be flexible to handel the variability in real world situations. There might be errors and noises in the measurement systems, both in deployment and in training (e.g. errors in the labels in the training data). Since there are many hyperparameters to design the model, techniques like neural architecture search (and graduate student descent) can introduce uncertainty in model structures. Even with the same models, different training procedures can give rise to different performance and fidelity. Finally, during the inference phase, models can give prediction with different confidence levels. \cite{gawlikowski2021survey} covers different approaches to measure and quantify uncertainty in neural networks.

\textit{Reinforcement learning} is the learning of trials and errors to deal with uncertainty in data. Usually, a reinforcement learning problem involves two subjects: an agent and an environment. The agent interact with the environment and the environment provides reinforcement to help the agent learn. More formally, at each time step $t$, the \textit{agent} takes an action $A_t$ to affect the \textit{environment}, and the environment transitions to the next state $S_t$ and reveals a reward feedback $R_t$ to the agent for it to update its policy (Figure \ref{fig:rlproblem}A). The exploration vs. exploitation tradeoff then resides on a spectrum: on the one end, if we exploit, we obtain short-term rewards by reaching good deployment performance given what we have known so far; on the other end, if we explore, we obtain long-term rewards because we are building more reliable models by collecting knowledges in uncertain problem spaces (Figure \ref{fig:rlproblem}B). Reinforcement learning algorithms can navigate between these two ends to get a dynamic strategy according to the problem and performance. It is adaptive, so it can continually learn to accomodate for the changes in the data. It is about optimizing for the expected future rewards, so it can innately predict or plan ahead for future events. Since it is mostly driven by its reward feedback, it is also generalizable to new tasks or uncertainty patterns by modifying its reward structure. These algorithmic properties suggest that speech and language models may benefit from adopting a reinforcement learning approach.

One useful perspective to approach an applied task of interest is to think the following questions: what are the costs and benefits in this learning system? When should this system explore (as opposed to exploit)? And how to properly explore the problem space? These three questions can help us better formulate the applied problems into RL framework.

\section{A Concise Tutorial of Reinforcement Learning and Bandits}

\begin{figure}[tb]
\centering
    \includegraphics[width=\linewidth]{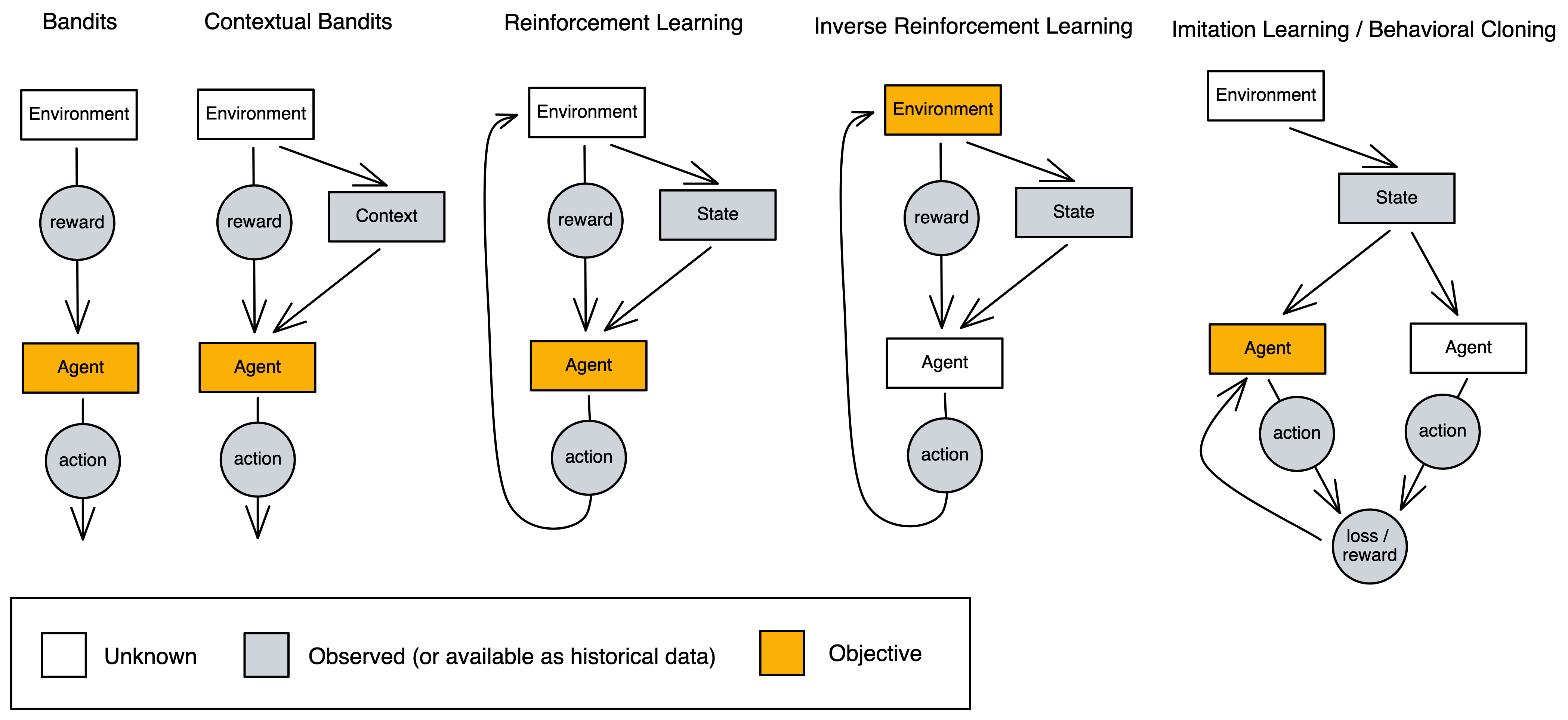}
\vspace{-1em}
\caption{Five classes of reinforcement learning-related problems. The blocks are models or systems. The circles are intermediate scalars or vectors. The white ones are unknown quantities or systems. The grey ones are observed variable, or sometimes, available historical data. The orange ones are the objectives to solve.
}\label{fig:rl_classes}
\end{figure}

In this section, we will give an applied tutorial of five classes of reinforcement learning-related problems. Figure \ref{fig:rl_classes} outlines the difference in their problem settings, where the circles are intermediate scalars or vectors, the white blocks are quantity or system that are unknown, the grey blocks are the observed variable or available historical data, and the orange blocks are the objectives that we are attempting to solve. We note that this visual representation is similar to a probabilistic graphical model, but that is not intended nor implied. 

\subsection{Preliminaries}

Most of the times, solving different classes of reinforcement learning can be modeled by a mathematical framework called the Markov decision processes (MDP) \cite{Sutton1998}. An MDP is defined by the tuple $(\mathcal{S}, \mathcal{A}, \mathcal{T}, \mathcal{R}, \gamma)$:

\begin{itemize}
    \item $\mathcal{S}$: a set of possible states
    \item $\mathcal{A}$: a set of actions
    \item  $\mathcal{T}$: a transition function, defined as $\mathcal{T}(s, a, s')=\Pr(s'\vert s,a)$, where $s, s'\in \mathcal{S}$ and $a\in \mathcal{A}$
    \item $\mathcal{R}:$ a reward function, $\mathcal{S}\times \mathcal{A} \times \mathcal{S}\mapsto \mathbb{R}$
\end{itemize}

This is a generalized formulation of the MDP. As we will see in the following the sections, some problems might only involve a subset of these notations (e.g. the bandits may not have states). Typically, the objective of the learning process is to maximize the long-term reward, assuming an infinite-horizon decision process. In other words, we want to find 
a \textit{policy} function, $\pi: \mathcal{S} \mapsto \mathcal{A}$, which specifies the action to take in a given state (which in the bandit scenario, might be a context or nothing), so that the cumulative reward is maximized.

\subsection{Multi-Armed Bandits (MAB)}

\begin{algorithm}[tb]

 \caption{Multi-Armed Bandits Problem}
 \label{alg:mab}
\begin{algorithmic}[1]
 \STATE {\bfseries }\textbf{for} t = 1, 2, 3, $\cdots$, T \textbf{do}
\STATE {\bfseries } \quad ${r}(t)$ is drawn according to $\mathbbm{P}_{r}$
\STATE {\bfseries }\quad  Player chooses an action $a_t =\pi_t(t)$
\STATE {\bfseries } \quad Feedback $r_{a_t,t}(t)$ for the arm $a_t$ is revealed
\STATE {\bfseries } \quad Player updates its policy $\pi_t$
\STATE {\bfseries } \textbf{end for}
 \end{algorithmic}
\end{algorithm}

Consider yourself at a casino, and in front of you, there are many bandit machines, each with an arm. For starter, we assume that they each have a static reward payoff, which is unknown to you. To gain better understanding of a certain bandit arm, you will need to pull that arm to gain a reward feedback revealed only for that arm. You want to play for $T$ rounds, and maximize your cumulative rewards. Back to our previous introduction to the Exploration vs Exploitation tradeoff: if we exploit entirely, we make long-term sacrifices by potentially missing unknown optimal action, because we are only choosing the best arm given the current information; if we explore entirely, we make short-term sacrifices by missing out known rewards, because we are always gathering more information. Due to the effectiveness of bandit algorithms in balancing the exploration vs exploitation tradeoff, they are widely applied in finance \cite{shen2015portfolio,charpentier2021reinforcement}, epidemic control \cite{lin2022optimal,lin2022evolutionary}, hyperparameter tuning \cite{li2017hyperband,parker2020provably}, recommendation systems \cite{yang2020exploring,wang2017biucb}, clinical prescription \cite{aziz2021multi,villar2015multi}, human behavioral modeling \cite{lin2020unified,lin2021models,bouneffouf2017bandit} and A/B testing \cite{satyal2018ab,xiang2022multi}.

As in algorithm \ref{alg:mab}, for each round, the player chooses an action $a_t$  given its policy $\pi_t$, and $r_a$, the feedback for only the chosen arm is revealed, which is often called \textit{bandit feedback}. The player then updates its policy $\pi_t$ given the feedback. If the player plays for $T$ trials, the goal of the agent is to maximize the total payoff (sum of rewards) $\sum_{t \in 1...T} r_t$. In another term, we can rephrase the performance metric to be minimizing total regret $\sum_{t \in 1...T} (r^* - r_t)$, where $r^*$ is the best arm reward. Other evaluation metrics include maximizing the average reward, and maximize the percentage of optimal action selection.

Here the agent is an algorithm, and the environment is the bandit task. Since they are both stochastic, the goal of this bandit task is empirically to maximize the expected total sum of rewards, i.e. $\mathbb{E}[\sum_{t \in 1...T} r_t]$. This learning process of continually updating its policy by sequentially interacting with the bandit environment is also called online learning. To report the results from an online learning environment, we can plot the above performance metrics over time steps (or trials). From this learning curve, we may compare different algorithms and select the best one.

We will introduce a few strategies how bandit algorithms deal with the exploration vs exploitation tradeoff. 

The first one is simply \textit{explore first} (by taking random actions for $N$ rounds) and then exploit (by choosing the arm with the highest estimated expected reward, the greedy approach). The estimated expected reward is defined by a value function called Q, where $\hat{Q}_t(a)$ is the estimated expected reward for action arm $a$ at time $t$. 
A simple formulation of the $\hat{Q}_t(a)$ is its average past observed rewards:

\begin{equation}
  \hat{Q}_t(a) := \hat{\mu}_a = \frac{r_1+r_2+...+r_{n_a}}{n_a}  
\end{equation}

where $r_i$ is the observed reward for action arm $a$, and $n_a$ is the number of times action arm $a$ has been played. $\hat{Q}_a^{t=0}=0$. There are many variants to this strategy by setting $N=f(N)$. For instance, one can explore once ($N=1$) for each action first, or the first $\epsilon$ proportion of rounds (i.e. the first $\epsilon \cdot T$ steps, where $0 \leq \epsilon \leq 1$). In the textbook ``Introduction to reinforcement learning'' \cite{Sutton1998}, a bandit simulation example of 10 arms with different average rewards trained over 1000 time steps compares six algorithms: Random (choosing randomly at each round), Greedy (choosing only the arm with the maximum Q value), Explore-First-20 (randomly choosing for only the first 20 rounds as pure exploration, and then choosing the arm with the maximum Q value), Explore-First-100, Explore-First-200, Explore-First-500. The random and purely greedy solutions both yield sub-optimal performance. Among the Explore-First-$N$ agents, the higher the $N$ (the exploration steps), the better the long-term cumulative rewards they get, but at the short-term initial rounds, especially when they are purely exploring, they receive rewards at the chance level, which might be a risky take for real-world problems.

A better strategy would be to dynamically switching between exploiting (taking the greedy action) and exploring (taking the random action). \textit{$\epsilon$-Greedy} algorithm describes a coin flipping process that, for a probability of $\epsilon$ ($0 \leq \epsilon \leq 1$), the agent take a random action at the current round, and for a probability of $1-\epsilon$, it takes the greedy action \cite{kaelbling1996reinforcement,cesa1998finite}. This approach doesn't explicitly specify a exploration phase and a exploitation phase, and generally yields a more balanced learning process. In the bandit simulation example, if we compare $\epsilon$-Greedy ($\epsilon=0$), $\epsilon$-Greedy ($\epsilon=0.01$) and $\epsilon$-Greedy ($\epsilon=0.1$), we observe a better cumulative long-term reward for larger $\epsilon$, i.e. more exploration. In the short-term, similarly, smaller $\epsilon$ can temporarily overtake larger ones by exploiting early, but they reach their plateau faster at lower payoff. 

Intuitively, at early rounds, the agent know little about the reward distribution of each arm, so it makes sense to explore more at early rounds. However, when we use methods like $\epsilon$-Greedy, the explored arms will have their Q values estimated in the first few steps larger than those of the unexplored arms (which are usually initialized zero), which discourages exploration. One strategy to encourage exploration in the early rounds, is to set all Q values with a large number to start with. This strategy of \textit{optimistic initial values}, can innately inject exploration (even in the greedy turns) to under-explored action arms \cite{kaelbling1996reinforcement,Sutton98}. Empirically (in bandit simulations), this trick yield better cumulative long-term rewards by trading off the rewards in very early rounds.

Another strategy would be to set a schedule for the exploration factor $\epsilon$ in $\epsilon$-Greedy, such that it reflects certain priors that can improve the learning. For instance, if we wish to encourage exploration in initial stages, we can use a \textit{decaying $\epsilon$} function (e.g. $\epsilon(t) \propto \frac{1}{t}$). Other variants of an \textit{$\epsilon$ schedule} can be a linear function, a step function, or oscillatory function if there are periodic changes in the environment. Despite proven poly-logarithmic bounds for these variants, it was reported limited advantages in these heuristics \cite{vermorel2005multi}. One can also potentially bind this scheduling with detection mechanism for paradigm shifts settings.

While useful, how to decide which scheduling function to use for the exploration? What if the environment is non-stationary and has irregular changes of reward distributions? And how best to motivate random exploration in a principle way? \textit{Probability matching} (PM) is a decision making strategy that choose actions given the probability of their reward in a stochastic setting \cite{luce2012individual,shanks2002re}. In other words, the probability of choosing arm $a$ at time $t$:

\begin{equation}
    p_t(a) \propto \frac{\hat{Q}_t(a)}{\sum_{a'\in A} \hat{Q}_t(a')}
\end{equation}

i.e. how likely the action arm is to be optimal. \textit{Boltzmann exploration} (or the SoftMax strategy) is such a formulation that samples a random choice according to the Gibbs distribution:

\begin{equation}
    p_t(a) = \frac{e^{ \frac{\hat{Q}_t(a)}{\tau}}}{\sum_{a'\in A} e^{ \frac{\hat{Q}_t(a')}{\tau}}}
\end{equation}

The hyperparameter $\tau$ is called temperature, which is a user-specified quantity to control the degree of exploration. If $\tau$ is set to be zero, it is full exploitation. If it is set to be infinity, it is full exploration. 

\begin{algorithm}[tb]

 \caption{Upper Confidence Bound (UCB) Algorithm}
 \label{alg:ucb}
\begin{algorithmic}[1]
 \STATE {\bfseries } Initialize $\hat{Q}_{a'}=0, n_{a'}=0, \forall a' \in A$
 \STATE {\bfseries }\textbf{for} t = 1, 2, 3, $\cdots$, T \textbf{do}
\STATE {\bfseries }\quad  Choose $a_t = $ $\begin{cases} \argmax_{a' \in A} \hat{Q}_{a'} + \sqrt{\frac{2\ln{t}}{n_{a'}}}, & \text{if } n_{a'} \neq 0 \\ -\infty, & \text{otherwise} \end{cases}$
\STATE {\bfseries } \quad Observe $r_{a_t,t}(t)$ for the arm $a_t$
\STATE {\bfseries } \quad $\hat{Q}_{a'} = \frac{\hat{Q}_{a'}\cdot n_{a_t} + r_{a_t,t}(t)}{n_{a_t} + 1}$
\STATE {\bfseries } \quad $n_{a_t} = n_{a_t} + 1$
\STATE {\bfseries } \textbf{end for}
 \end{algorithmic}
\end{algorithm}

Similar to $\epsilon$-Greedy, we can also build variants of Boltzmann exploration with schedules of the temperature $\tau$ (e.g. having the temperature decays with the number of rounds played). However, the dependency of the user to choose the exploration level (either $\epsilon$ or $\tau$) at each step can be problematic (e.g. unlucky initial experience) and a principled way to select them remains elusive. One strategy would be to use other quantity to guide exploration, such as adopting the \textit{uncertainty-guided exploration}. \cite{auer2002nonstochastic} proposes the family of \textit{Upper Confidence Bound} (UCB) algoritms as a elegant algorithmic implementation of the idea of \textit{optimism in the face of uncertainty} by \cite{LaiRobbins1985}. As in algorithm \ref{alg:ucb}, each action arm is represented by their estimated expected reward $\hat{Q}_a$ and a confidence bound $\sqrt{\frac{2\text{ln} N}{n_a}}$ as their uncertainty. The agent simply chooses the action that maximize the upper confidence bound:

\begin{equation}
  a_t = \argmax_{a' \in A} \hat{Q}_{a'} + \sqrt{\frac{2\ln{N}}{n_{a'}}}  
\end{equation}

where $N$ is the number of rounds played so far and $n_a$ is the number of time this arm $a$ has been chosen. We quickly notice that, the first term is the exploitation term and the second term is the exploration term. (Upper Confidence Bound-1) UCB1 is the most common implementation, which first let each arm to be played at least once before adopting the upper confidence bound action selection policy. Similar to $\epsilon$-Greedy and Boltzmann exploration, one can create variants of UCB1 by introducing a schedule parameter $C$ to the exploration: $\hat{Q}_{a} + C \cdot \sqrt{\frac{2\ln{N}}{n_{a}}}$.

\begin{algorithm}[tb]

 \caption{Thompson Sampling (TS) Algorithm}
 \label{alg:ts}
\begin{algorithmic}[1]
 \STATE {\bfseries } Initialize $S_{a'}=0, F_{a'}=0, \forall a' \in A$
 \STATE {\bfseries }\textbf{for} t = 1, 2, 3, $\cdots$, T \textbf{do}
 \STATE {\bfseries }\quad $\hat{Q}_t(a') \sim Beta(S_{a'}+1, F_{a'}+1) \forall a' \in A$
\STATE {\bfseries }\quad  Choose $a_t = \argmax_{a' \in A} \hat{Q}_t(a')$
\STATE {\bfseries } \quad Observe $r_{a_t,t}(t)$ for the arm $a_t$
\STATE {\bfseries } \quad $S_{a_t} = S_{a_t} + r_{a_t,t}(t)$
\STATE {\bfseries } \quad $F_{a_t} = F_{a_t} + (1 - r_{a_t,t}(t))$
\STATE {\bfseries } \textbf{end for}
 \end{algorithmic}
\end{algorithm}

The UCB algorithms are specified from a frequentist point of view, one can also adopt a Bayesian approach to this strategy. A \textit{Bayesian bandit} would represent the action value Q function as a probability distribution $p(\hat{Q}_t(a))$ and directly compute the probability distribution of the Q values using the Bayes rule: 

\begin{equation}
    p(\hat{Q}_t(a)|D_a) \propto p(D_a | \hat{Q}_t(a)) \cdot p(\hat{Q}_t(a))
\end{equation}

where $D_a$ is the observed data (past rewards revealed when this arm is selected): $D_a = \{r_1, r_2, r_3, \cdots, r_{n_a}\}$. The action selection strategy for Bayesian bandits would be, instead of using the Q values directly, the agent first sample from the probability distribution $p(\hat{Q}_t(a))$ of the Q value for each action arm. \textit{Thompson sampling} \cite{T33} (TS) is one such algorithm that shows a competitive performance with other approaches \cite{chapelle2011empirical}. 
In the Bernoulli bandit problem (where the rewards are drawn from a Bernoulli distribution: $R_a \sim Ber(p_a \in [0, 1])$), Thompson sampling is proven to be asymptotically optimal \cite{AgrawalG12}. The implementation is very straightforward. As in algorithm \ref{alg:ts}, we can express our uncertainty about $\hat{Q}_t(a)$ with:

\begin{equation}
    p(\hat{Q}_t(a)) \sim Beta(S_{a}+1, F_{a}+1)
\end{equation}

where $S_{a}$ and $F_{a}$ are two values stand for ``success'' and ``failure''. If the reward is 1, then we increment the ``success'' $S_{a}$ by 1. And if the reward is 0, then we increment the ``failure'' $F_{a}$ by 1. Since this is a class of probability matching method, we select the action based on $p(\hat{Q}_a| D_a)$ where $D_a = \{r \in [0, 1]\}$. In our case, we randomly sample an estimate $\hat{Q}_t(a')$ from the posterior, i.e. the Beta distribution of each arm parameterized by $S_{a}$ and $F_{a}$, and then choose the action arm $a_t$ that has the maximum $\hat{Q}_t(a')$.

\begin{table}[tb]
 \centering
\begin{tabular}{l|c|c}
Approach & Exploitation & Exploration \\ \hline
 Switching to random exploration & $a_t = \argmax_{a \in A} \hat{Q}_t{a}$ & $a_t = $ random action \\ \hline
 Uncertainty-guided exploration & \multicolumn{2}{c}{$a_t = \argmax_{a \in A} \hat{Q}_t{a} + \text{measure of uncertainty}$} \\ \hline
 Probability matching  & Use $\hat{Q}_t{a}$ to define $p_t(a)$ & Select $a_t$ according to $p_t(a)$ \\ 
\end{tabular}
\vspace{1em}
\caption{Summary of bandit action selection strategies}
\label{tab:recap_bandits}
\end{table}

As a recap, we cover five strategies: exploration first, optimistic initial values, parameter schedule, guided exploration and probability matching. We may represent the action value functions $\hat{Q}_t(a)$ in two approaches: the Frequentist approach ($\hat{Q}_t(a) := \hat{\mu}_a = \frac{r_1+r_2+...+r_{n_a}}{n_a}$) and the Bayesian approach (express the uncertainty about $\hat{Q}_t(a)$ with $p(\hat{Q}_t(a))$). As summarized in table \ref{tab:recap_bandits}, there are three action selection strategies. We can explicitly separate the exploration (by taking random action) and the exploitation (by taking greedy action) into different phases. We can use uncertainty to encourage optimism and guide exploration by combining the exploitation (the action estimate) and the exploration (the measure of uncertainty) together into the argmax criterion. Finally, in probability matching approach, we use $\hat{Q}_t{a}$ to define $p_t(a)$ (exploitation) and select $a_t$ according to $p_t(a)$ (exploration). Some effective algorithms include $\epsilon$-Greedy, upper confidence bound (UCB), and Thompson sampling (TS). 

We wish to point out that, the bandits by itself is a rich field, and there are many variants to these solutions. The reward distributions can be independent or correlated \cite{lazaric2014online}, stochastic or adversarial \cite{AuerC98,AuerCFS02,BouneffoufF16}, static or non-stationary \cite{garivier2008upper,lin2018contextual}. The action space can be discrete or continuous \cite{srinivas2009gaussian,trovo2016budgeted}, finite or infinite number of arms \cite{wang2008algorithms}, single or multi-dimensional (combinatorial) \cite{chen2013combinatorial,lin2021optimal}. The time horizon can be infinite or finite \cite{lattimore2016regret}. There can be a single agent or a population of bandit agents \cite{lin2022evolutionary}. The feedbacks can be bandit or partially full feedback \cite{kocak2014efficient}, always revealed or only sporadically revealed (sparse or missing feedback) \cite{berlinucb}, and sometimes even budgeted \cite{ding2013multi,lin2021optimal,badanidiyuru2018bandits}. As such, one can choose from many different subproblems of bandits, such as adversarial bandits, combinatorial bandits, dueling bandits, bandits with a budget or knapsack, multi-play bandits, bandits with dependent arms and finally the most important variant of all, the contextual bandits.
For readers with more interests, \cite{slivkins2019introduction} is a good introduction to different classes of bandit algorithms.


\subsection{Contextual bandits}

\begin{algorithm}[tb]

 \caption{Contextual Bandits (CB) Problem}
 \label{alg:cb}
\begin{algorithmic}[1]
 \STATE {\bfseries }\textbf{for} t = 1, 2, 3, $\cdots$, T \textbf{do}
\STATE {\bfseries } \quad $({x}(t), {r}(t))$ is drawn according to $\mathbbm{P}_{x,r}$
\STATE {\bfseries }\quad  Context ${x}(t)$ is revealed to the player
\STATE {\bfseries }\quad  Player chooses an action $a_t =\pi_t({x}(t))$
\STATE {\bfseries } \quad Feedback $r_{a_t,t}(t)$ for the arm $a_t$ is revealed
\STATE {\bfseries } \quad Player updates its policy $\pi_t$
\STATE {\bfseries } \textbf{end for}
 \end{algorithmic}
\end{algorithm}

The contextual bandits (CB) describe the scenario of multi-armed bandits that also receives side information as decision making contexts \cite{langford2008epoch}. As in algorithm \ref{alg:cb}, the contextual bandits problem is defined as follows. At each time point (or iteration) $t$, before choosing the arm $a_t \in A$, agent observes an $M$-dimensional context, $x_t$, as a feature vector of $M$ variables. The reward function $r(t)$ is drawn according to $\mathbbm{P}_{x,r}$, where $r_a(t) \in [0,1]$ is a reward at time $t$ associated with the arm $a \in A$ and the context at the current iteration $x_t$ (or alternatively, we can define a joint distribution over $(x,r)$). 
The agent uses this context as a side information, along with the rewards of the arms played in the past, to choose which arm to play in the current iteration.
The objective of this variant problem is to learn the relationship between the context and reward, in order to find the best arm-selection policy for maximizing cumulative reward over the time horizon.

\begin{algorithm}[tb]

 \caption{Linear Upper Confidence Bound (LinUCB) Algorithm}
 \label{alg:linucb}
\begin{algorithmic}[1]
 \STATE {\bfseries } \textbf{Initialize} $c_t \in \mathbbm{R}_+, {A}_a = {I}_d, {b}_a = {0}_{d \times 1}, \forall a \in A $
 \STATE {\bfseries }\textbf{for} t = 1, 2, 3, $\cdots$, T \textbf{do}
  \STATE {\bfseries } \quad Observe features ${x}_{t} \in \mathbbm{R}^d$
 \STATE {\bfseries } \quad \textbf{for all} $a \in A $ \textbf{do}
 \STATE {\bfseries }  \quad \quad  $\hat{{\theta}}_a = {A}_a^{-1} {b}_a$
 \STATE {\bfseries } \quad \quad $p_{t,a} = \hat{{\theta}}_a^{\top} {x}_{t}+c_t\sqrt{{x}_{t}^{\top}{A}_a^{-1} {x}_{t}}$
\STATE {\bfseries } \quad Choose arm $a_t=\argmax_{a \in A}p_{t,a}$ 
 \STATE {\bfseries }  \quad  Observe feedback $r_{a_t,t}$ 
\STATE {\bfseries }  \quad  ${A}_{a_t} = {A}_{a_t}+{x}_{t}{x}^{\top}_{t}$ 
\STATE {\bfseries }  \quad  ${b}_{a_t} = {b}_{a_t}+r_{a_t,t}{x}_{t}$ 
\STATE {\bfseries }\textbf{end for}
 \end{algorithmic}
\end{algorithm}

\begin{algorithm}[tb]
   \caption{Contextual Thompson Sampling (CTS) Algorithm}
\label{alg:cts}
\begin{algorithmic}[1]
 \STATE {\bfseries }\textbf{Initialize:} $B_a = I_d$, $ \hat{\theta}_a= 0_d, f_a = 0_d$, $\forall a \in A$.
 \STATE {\bfseries }\textbf{for } $t = 1, 2, 3, \cdots, T$ \textbf{do}
 \STATE \quad Receive context ${\bf x}_t$
 \STATE {\bfseries }\quad   \textbf{for } $\forall a \in A$,  sample $\tilde{\theta_{a}}$ from the $N(\hat{\theta}_a, v^2 B_a^{-1})$  
 \STATE {\bfseries }\quad Choose arm $a_t= \argmax_{a \in A} x(t)^\top \tilde{\theta_{a}} $
  \STATE {\bfseries }\quad Receive reward  $r_{a_t}$
 \STATE {\bfseries }\quad
 $B_{a_t}= B_{a_t}+ x_t  x_t^{T} $
 \STATE {\bfseries }\quad $f_{a_t} = f_{a_t} + x_t  r_{a_t}$
 \STATE {\bfseries }\quad $\hat{\theta_{a_t}} = B_{a_t}^{-1} f_{a_t}$
 \STATE {\bfseries }\textbf{end}
   \end{algorithmic}
\end{algorithm}

For introduction purpose, here we assume a Bernoulli bandit with binary reward, i.e. $r \in [0,1]$. We can adopt a linear assumption \cite{ChuLRS11}, that the expected reward is a linear function of the context: 

\begin{equation}
    \mathbb{E}[r_a|\textbf{x}_t] = \theta_a^\top \textbf{x}_t
\end{equation}

where $\mu_a$ is an unknown weight vector associated with the arm $k$. The Linear Upper Confidence Bound (LinUCB) algorithm is the contextual variants of the UCB algorithm \cite{ChuLRS11}. The action value function now becomes $\hat{Q}_t(x_t, a)$ and can be used to estimate $\mathbbm{E}[R_a|x_t]$

We can formulate the problem as a least-square ridge regression problem on its reward function: 

\begin{equation}
    R_a(x_t) = f_a(x_t) + \epsilon = x_t^\top \theta+a + \epsilon
\end{equation}

where $f_a$ is an arbitrary function that maps the context to the reward, $\theta_a \in \mathbbm{R}^d$ and $\epsilon$ is the noise or error. At each time step $t$ and given the context $x_t$, the weight for the action value solution becomes:

\begin{equation}
    \hat{Q}_t(x_t,a) = x^\top_t \hat{\theta_a} 
        \end{equation}
  \begin{equation}
    \hat{\theta_a} = (D_a^\top D_a + I_d)^{-1} D_a^\top C_a
\end{equation}

where $D_a$ and $C_a$ are the historical contexts and rewards observed when this arm $a$ is chosen: $D_a =\begin{bmatrix}
x_1\\
x_2 \\
\vdots \\
x_n \\
\end{bmatrix}, C_a =\begin{bmatrix}
r_1\\
r_2 \\
\vdots \\
r_n \\
\end{bmatrix}$.

The action selection will then depends on the estimated rewards with the mapping weight given the context:

\begin{equation}
    \hat{S}_t(x_t,a) = \sqrt{x_t^\top (D_a^\top D_a + I_d)^{-1} x_t} 
    \end{equation}
  \begin{equation}
    a_t = \argmax_{a \in A} (\hat{Q}_t(x_t, a) + \alpha \cdot S_t(x_t, a))
\end{equation}

where similar to the non-contextual version of UCB, here we have two terms in the argmax option, the first term $\hat{Q}_t(x_t, a)$ is for exploitation and the second term $S_t(x_t, a)$ is for the uncertainty-guided exploration, modulated by the exploration factor $\alpha$. 

This is the analogy to the linear regression problem, while empirically one can simply incrementally compute the solution based on online feedback as in algorithm \ref{alg:linucb}: we have matrix $A_a$ which characterizes the covariance of contexts, and $b_a$ which characterizes the reward mapping; and at the action selection stage, we take the argmax of $({A}_a^{-1} {b}_a)^{\top} {x}_{t}+c_t\sqrt{{x}_{t}^{\top}{A}_a^{-1} {x}_{t}}$; at the update stage, we simply incrementally update the covariance matrix $A_a$ and reward mapping $b_a$ for the selected arm $a_t$.

Similarly, we can formulate a linear contextual version of the Thompson sampling algorithm, the Contextual Thompson Sampling \cite{AgrawalG13} algorithm. We consider the general Thompson Sampling, where the reward $r_{a_t}$ for choosing arm $a_t$ at time $t$ follows a parametric likelihood function $p(r_t | \tilde{\theta}_a)$. The posterior distribution at time $t$:

\begin{equation}
    p(\tilde{\theta}_i | r_t) \propto Pr(r_t | \tilde{\theta}_i)
\end{equation}

To incorporate contextual information, $p(\tilde{\theta}_i)$ is given by a multivariate Gaussian distribution $\mathcal{N}(\hat{\theta}_a$, $v^2 B_a^{-1})$, where
$B_a(t)= I_d + \sum^{t-1}_{\tau=1} x_{\tau} x_{\tau}^\top$, and 
where $d$ is the size of the context vectors $x_t$, $v=R \sqrt{\frac{24}{\epsilon} d  \ln(\frac{1}{\gamma})}$ with $r>0$, $\epsilon \in [0,1]$, $\gamma \in [0,1]$, and $\hat{\theta}_a(t)=B_a(t)^{-1} (\sum^{t-1}_{\tau=1} x_{\tau} r_{\tau})$. At every step $t$ for each action arm $a$, the algorithm samples a $d$-dimensional context weight vector $\tilde{\mu_i}$ from
$\mathcal{N}(\hat{\theta_a}(t)$, $ v^2 B_a(t)^{-1})$. The agent would select the arm $a_t$ that maximizes $x_t^\top \tilde{\theta_a}$, and obtains reward $r_t$. 
\cite{zhou2015survey} is a survey of different contextual bandits algorithms.

\subsection{Reinforcement learning}

\begin{algorithm}[tb]

 \caption{Reinforcement Learning (RL) Problem}
 \label{alg:rl}
\begin{algorithmic}[1]
 \STATE {\bfseries }\textbf{for} t = 1, 2, 3, $\cdots$, T \textbf{do}
\STATE {\bfseries }\quad Agent makes an observation $o(t)$ to the state $s(t)$ 
\STATE {\bfseries }\quad Agent chooses an action given the current state or observation: $a_t =\pi_t({o}(t))$
\STATE {\bfseries } \quad Environment progresses to the next step $s_{t+1}$ given the action $a_t$
\STATE {\bfseries } \quad Agent receives a reward feedback $r_{a_t,s_t}(t)$ for the action $a_t$ taken on state $s_t$
\STATE {\bfseries } \quad Agent updates its policy $\pi_t$
\STATE {\bfseries } \textbf{end for}
 \end{algorithmic}
\end{algorithm}

Reinforcement learning is a general-purpose framework for decision-making. Reinforcement learning is for an agent with the capacity to act. Each action taken by the agent influences the agent’s future state. The success of this problem is measured by a scalar reward signal. As in the bandits and contextual bandits cases, the goal is to select actions to maximize cumulative future rewards. 

As hinted in the section of ``Preliminaries'', reinforcement learning defines a class of algorithms for solving problems modeled as Markov decision processes (MDP). Other than previously defined tuple $(\mathcal{S}, \mathcal{A}, \mathcal{T}, \mathcal{R})$, $\gamma \in [0,1]$ is a discounting factor for past or future rewards (the further into the past or future, the less impact of that reward has on current action choice). Typically, the objective is to maximize the discounted long-term reward, assuming an infinite-horizon decision process. In other words, we wish to find an optimal policy function $\pi: \mathcal{S} \mapsto \mathcal{A}$, which specifies the action to take in a given state, so that the cumulative reward is maximized: 

\begin{equation}
\max_{\pi} \sum_{t=0}^{\infty}\gamma^t \mathcal{r}(s_t, a_t, s_{t+1})
\end{equation}

The experience of a reinforcement learning agent is a sequence of observations, actions and rewards:

\begin{equation}
    o_1, a_1, r_1, \cdots, r_{t-1}, o_t, a_t, r_t 
\end{equation}

The state is a summary of the experience defined above:

\begin{equation}
    s_t = f(o_1, a_1, r_1, \cdots, r_{t-1}, o_t, a_t, r_t)
\end{equation}

which in a full observed environment, can be:

\begin{equation}
    s_t = f(o_t)
\end{equation}

A reinforcement learning agent usually involve one or more of three components: a \textit{policy} (which the agent's behavior function), a \textit{value function} (which  evaluates how good each state and/or action is), and a \textit{model} (which is the agent's representation of the environment). More specifically, the policy maps from the state to action, which can be deterministic ($a = \pi(s)$) or stochastic ($\pi(a|s)=p(a|s)$). The model is learned from experience, and acts as a proxy for the environment. With its model, the agent can make sense how world changes given agent's action in two ways: the transition or dynamics model predicts the agent's next state given an action
$
    p(s_{t+1}=s' | s_t=s, a_t=a)
$; and the reward model predicts the immediate reward at state given an action $r(s_t=s, a_t=a) = \mathbbm{E}[r_t|s_t=s, a_t=a]$.

The value function is a prediction of future reward, i.e. what rewards will the agent gets by taking an action $a$ in state $s$. Q-value functions gives the expected total reward from state $s$ via action $a$ under policy $\pi$ with discount factor $\gamma$:

\begin{equation}
    Q^\pi(s,a) = \mathbbm{E}[r_{t+1}+\gamma r_{t+2}+\gamma^2 r_{t+3}+\cdots | s, a]
\end{equation}

which can be decomposed into a Bellman equation:

\begin{equation}
    Q^\pi(s,a) = \mathbbm{E}_{s',a'}[r+\gamma Q^\pi(s', a') | s, a]
\end{equation}

where an optimal value function can be further obtained by taking the maximum achievable value:

\begin{equation}
    Q^*(s,a) = \max_\pi Q^\pi(s', a') = Q^{\pi^*} (s,a)
\end{equation}

With this optimal value function $Q^*$, the agent can act optimally:

\begin{equation}
    \pi^*(s) = \argmax_a Q^*(s,a)
\end{equation}

Maximizing over all possible decisions, $Q^*(s,a)$ can also be decomposed into a Bellman equation:

\begin{align}
    Q^*(s,a) &= r_{t+1}+\gamma \max_{a_{t+1}} r_{t+2}+\gamma^2 \max_{a_{t+2}} r_{t+3}+\cdots  \\
    &= r_{t+1}+\gamma \max_{a_{t+1}} Q^*(s_{t+1}, a_{t+1}) \\
    &= \mathbbm{E}_{s'}[r+\gamma \max_{a'} Q^*(s',a') | s, a]
\end{align}

\begin{figure}[tb]
\centering
    \includegraphics[width=\linewidth]{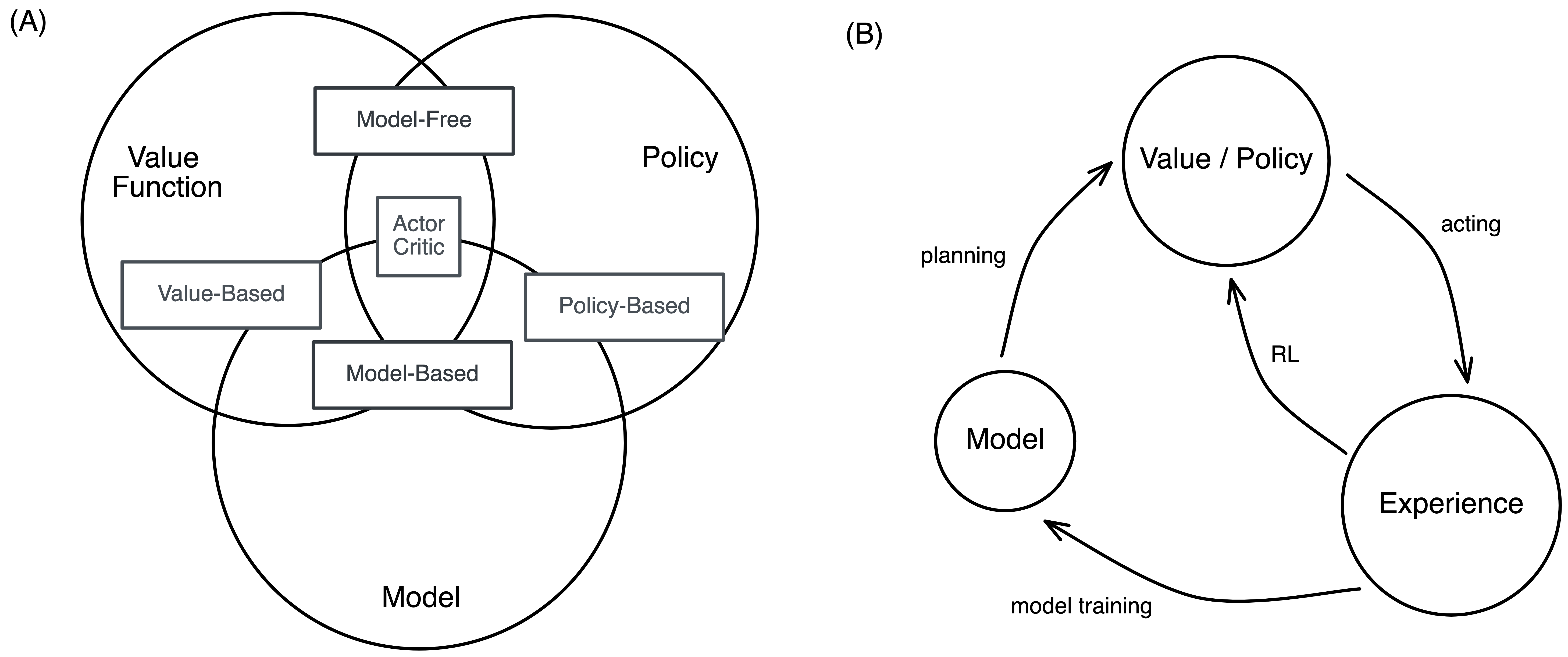}
\vspace{-1em}
\caption{(A) Categorization of reinforcement learning approaches. (B) Sub-processes of reinforcement learning. (Re-created from David Silver's lecture).
}\label{fig:rl}
\end{figure}

There are different ways to categorize reinforcement learning approaches. Figure \ref{fig:rl}A positions several approaches with respect to their spectrum on the three major model components. Figure \ref{fig:rl}B also includes experience as a landing point for sub-processes such as acting, model training and plannings. Overall there are \textit{model-based} reinforcement learning and \textit{model-free} reinforcement learning. Model-based methods construct a full model of the environment explicitly and then plan ahead by rolling out future steps using the model. It may or may not have an explicit policy and/or value function. Model-free methods, on the other hand, doesn't have a model, and has explicit functions for the value and/or policy. The model-free reinforcement learning can be further separated into \textit{value-based} and \textit{policy-based} reinforcement learning. Value-based reinforcement learning estimate the optimal value function $Q^*(s,a)$, which is the maximum value achievable under any policy. Policy-based reinforcement learning search directly for the optimal policy $\pi^*$, and use that policy to achieve the maximum future reward. There are methods which are both value-based and policy-based, such as the actor-critic algorithm, which we will cover in a bit.

\begin{algorithm}[tb]
 \caption{Q-Learning Algorithm}
\label{alg:ql}
\begin{algorithmic}[1]
  \STATE {\bfseries } {\bf Initialize:} $Q_{s,a} = 0 \forall s \in S, \forall a \in A$
  \STATE \textbf{for} each episode $e$ \textbf{do}
 \STATE {\bfseries } \quad Initialize state $s$
 \STATE {\bfseries } \quad \textbf{Repeat} for each step $t$ of the episode $e$
 \STATE {\bfseries } \quad \quad Take action $a = \arg \max_{a'}Q(s,a')$, and
  \STATE {\bfseries } \quad \quad Observe $s'\in S$, $r \in R(s)$ \\
  \STATE {\bfseries } \quad \quad $s = s'$ \\
 \STATE {\bfseries }  \quad \quad $Q(s,a) := \hat{Q}(s,a) + \alpha_t (r + \gamma \max_{a'}\hat{Q}(s',a')-\hat{Q}(s,a))$
 \STATE {\bfseries } \quad \textbf{until} s is the terminal state
 \STATE {\bfseries }\textbf{end for}
 \end{algorithmic}
\end{algorithm}

\textit{Q-learning} is a model-free value-based reinforcement learning algorithm.
The Q-value of a state-action pair, ${Q}(s,a)$, is key to this algorithm, representing the expected future discounted reward for taking action $a \in \mathcal{A}$ in state $s \in \mathcal{S}$. The action selection strategy is straightforward: $a_t = \argmax_a Q^\pi(s, a)$. Previously, we have shown that the optimal value function can be translated into a Bellman equation $Q^*(s,a) = \mathbbm{E}_{s'}[r+\gamma \max_{a'} Q^*(s',a') | s, a]$. Therefore, we can potentially approximate $Q^*$ by directly attempting to solve the optimal Bellman equation. To do so, we define the Bellman error as the update to our expected return when we observe the next state $s'$:

\begin{equation}
    r(s_t, a_t) + \gamma \max_{a} Q(s_{t+1},a) - Q(s_t, a_t)
\end{equation}

The first half of this error term, $r(s_t, a_t) + \gamma \max_{a} Q(s_{t+1},a)$ is the right hand side of the Bellman equation, and sometimes also related to the notion of the temporal difference (TD) target.
As in algorithm \ref{alg:ql}, Q-learning is the algorithm that repeatedly adjusts Q to minimize the Bellman error. Ideally, when the policy is converged to the optimal solution, the Bellman error would be zero. Each time, we sample consecutive states and actions to update the Q values:

\begin{equation}
    Q(s_t, a_t) = Q(s_t, a_t) + \alpha [r(s_t, a_t) + \gamma \max_{a} Q(s_{t+1},a) - Q(s_t, a_t)]
\end{equation}

where $\alpha$ is the learning rate. A common method to handle very large state spaces is to approximate the ${Q}$ function as a linear function of some features representing the observation or state space \cite{bertsekas1996neuro}. We denote $\psi(s,a)$  as the relevant features of the state-action pair $\langle s, a \rangle$. Assuming $Q(s,a) = \theta \cdot \psi(s,a)$, where $\theta$ is an unknown weight vector to be learned by interacting with the environment. Then each time step the agent takes action $a$ at state $s$, obtains immediate reward $r$ and arrives at the new state $s'$, the parameter $\theta$ is updated with the temporal:

\begin{equation} \label{eqn:q-update}
\begin{aligned}
\text{TD} &= (r + \gamma \max_{a'} {Q}(s',a')) - {Q}(s,a)\\
\theta_i &= \theta_i + \alpha \cdot \text{TD} \cdot \psi_i(s,a),
\end{aligned}
\end{equation}

As in the bandit strategies, a common approach to balance the exploration vs exploitation tradeoff here is to use the  $\epsilon$-Greedy strategy. The agent, using the $\epsilon$-Greedy to gradually update its weight parameter $\theta$ according until convegence or some pre-specified maximal number of steps.

\textit{Policy gradient} is a model-free policy-based method that learns the policy directly with a parameterized function respect to parameter $\theta$. The value of the reward (the objective) function depends on this policy $\pi(a|s)$, and then different algorithms can be applied to optimize $\theta$ for the best reward \cite{weng2018PG}. The modeled reward function as the optimization objective is defined as:

\begin{equation}
    J(\theta) = \sum_{s\in S} d\pi(s)V^\pi(s) = \sum_{s\in S} d^\pi(s) \sum_{a\in A} Q^\pi(s, a) \pi_\theta(a|s)
\end{equation}

where $V(s)$ is the state-value function measures the expected return of state, $V^\pi(s)$ is the value of state $s$ under policy $\pi$, $d^\pi(s)$ is the on-policy state under the policy $\pi$ given by the stationary distribution of the Markov transition model. Under the policy gradient theorem \cite{Sutton98}, the gradient $\nabla_\theta J(\theta)$ can be computed as:

\begin{align}
     \nabla_\theta J(\theta) &= \nabla_\theta \sum_{s\in S} d^\pi(s) \sum_{a\in A} Q^\pi(s, a) \pi_\theta(a|s) \\
     &\propto \sum_{s\in S} d^\pi(s) \sum_{a\in A} Q^\pi(s, a) \nabla_\theta \pi_\theta(a|s)  \\
     &= \sum_{s\in S} d^\pi(s) \sum_{a\in A} \pi_\theta(a|s) Q^\pi(s, a) \frac{\nabla_\theta \pi_\theta(a|s)}{\pi_\theta(a|s)} \\
     &= \mathbbm{E}^\pi_{s\sim d_\pi,a\sim \pi_\theta}[Q^\pi(s,a)\nabla_\theta \ln \pi_\theta(a|s)]
\end{align}

\begin{algorithm}[tb]
 \caption{REINFORCE Algorithm}
\label{alg:reinforce}
\begin{algorithmic}[1]
  \STATE {\bfseries } {\bf Initialize:} $\theta$ arbitrarily
  \STATE \textbf{for} each episode (on-policy trajectories) $\{s_1, a_1, r_1, \cdots, s_T, a_T, r_t\} \sim \pi_\theta$ \textbf{do} 
   \STATE \quad {\bfseries }\textbf{for} t = 1, 2, 3, $\cdots$, T \textbf{do}
 \STATE {\bfseries } \quad \quad $\theta = \theta + \alpha V^\pi(s) \nabla_\theta \ln \pi_\theta(a | s)$
  \STATE {\bfseries } \quad \textbf{end for}
  \STATE {\bfseries }\textbf{end for}
 \end{algorithmic}
\end{algorithm}

\cite{schulman2015high} is a nice introduction to policy gradient objective optimization, that covers general advantage estimation to control the variance and bias in this general form. More specifically, \textit{REINFORCE}, also known as Monte-Carlo policy gradient, is an effective algorithm to perform policy gradient reinforcement learning \cite{sutton1999policy}. As in algorithm \ref{alg:reinforce}, we can sometimes simplify the term $Q^\pi(s, a)$ in our gradient with the state value function $V^\pi(s)$, which is the value of state when we follow a policy $\pi$, assuming the chosen action $a_t$ is the optimal one so far that yields the best estimate of the expected future reward starting from this state $s_t$. Using episode samples, the algorithm applies Monte Carlo methods to estimate the expected return, i.e. the Q function $Q^\pi(s, a)$, to update the policy parameter $\theta$ given the gradient ascent update. A empirical strategy to improve the learning in REINFORCE is to use the advantage function $A(s, a) = Q(s,a) - V(s)$ instead of the value function in the gradient ascent update, because it reduces the variance of the gradient estimation while maintaining the bias necessary for learning \cite{weng2018PG}.

The \textit{actor-critic} algorithm is a model-free algorithm that is both policy-based and value-based \cite{konda1999actor}. On top of the policy gradient methods introduced above, the actor-critic algorithm also learns the value function in addition to its policy. This is another strategy to reduce the variance in the gradient in the general form of policy gradients. The critic updates the parameters $w$ of its value function (which can be either the action-value $Q_w(s, a)$ or the state-value $V_w(s)$). The actor updates the parameter $\theta$ for the policy $\pi_\theta(a|s)$ along the gradient direction suggested by the critic. As in algorithm \ref{alg:ac}, other than the update of the policy parameter $\theta$, we also compute the TD error as in Q-learning with function approximation, and update the weight $w$ for Q function given the TD error. We can define two learning rate, $\alpha_\theta$ and $\alpha_w$ separately for the value and policy updates.

\begin{algorithm}[tb]
 \caption{Actor-Critic Algorithm}
\label{alg:ac}
\begin{algorithmic}[1]
  \STATE {\bfseries } {\bf Initialize:} $\theta$ arbitrarily
  \STATE \textbf{for} each episode (on-policy trajectories) $\{s_1, a_1, r_1, \cdots, s_T, a_T, r_t\} \sim \pi_\theta$ \textbf{do} 
   \STATE \quad {\bfseries }\textbf{for} t = 1, 2, 3, $\cdots$, T \textbf{do}
 \STATE {\bfseries } \quad \quad Update policy parameter $\theta$: $\theta = \theta + \alpha_\theta Q_w(s,a) \nabla_\theta \ln \pi_\theta(a | s)$
  \STATE {\bfseries } \quad \quad Compute TD error $\delta$: $\delta_t=r_t+\gamma Q_w(s_{t+1},a_{t+1}) - Q_w(s,a)$
   \STATE {\bfseries } \quad \quad Update critic parameter $w$: $w=w+\alpha_w \delta_t \nabla_w Q_w(s,a)$
  \STATE {\bfseries } \quad \textbf{end for}
  \STATE {\bfseries }\textbf{end for}
 \end{algorithmic}
\end{algorithm}

There are many other variants of the policy gradient methods that have the equivalent forms for optimizations. Some examples are:

\begin{align}
     \nabla_\theta J(\theta) 
     &= \mathbbm{E}^\pi_{s\sim d_\pi,a\sim \pi_\theta}[\nabla_\theta \ln \pi_\theta(a|s) V^\pi(s)]  & REINFORCE \\
     &= \mathbbm{E}^\pi_{s\sim d_\pi,a\sim \pi_\theta}[\nabla_\theta \ln \pi_\theta(a|s)Q_w(s,a)] & Q Actor-Critic \\
     &= \mathbbm{E}^\pi_{s\sim d_\pi,a\sim \pi_\theta}[\nabla_\theta \ln \pi_\theta(a|s)A_w(s,a)] & Advantage Actor-Critic \\
     &= \mathbbm{E}^\pi_{s\sim d_\pi,a\sim \pi_\theta}[\nabla_\theta \ln \pi_\theta(a|s)\delta] & TD Actor-Critic
\end{align}

\begin{figure}[tb]
\centering
    \includegraphics[width=\linewidth]{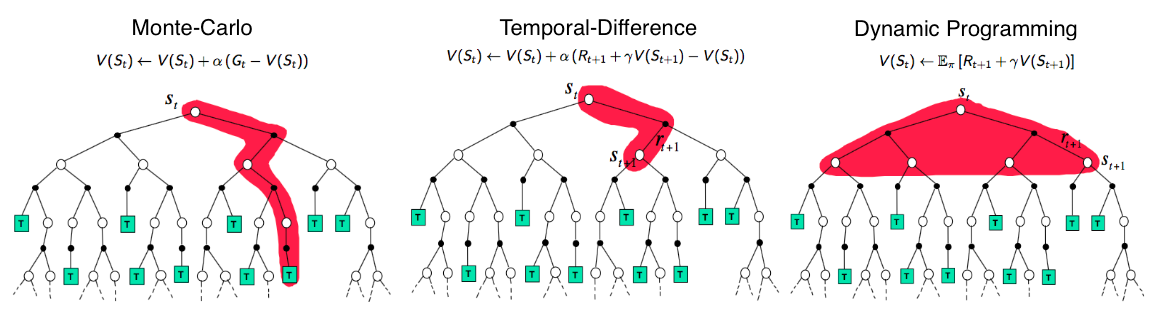}
\vspace{-1em}
\caption{Comparison of the rollout search space (or backup diagrams) of Monte Carlo (MC), Temporal Difference (TD) learning and Dynamic Programming (DP) for state value functions (from David Silver's RL lecture). Additional notation: $G_t$ is the final return, or expected future reward, $G_t=\sum_\tau \gamma^\tau r_{t+\tau+1}$ \cite{Sutton98}. 
}\label{fig:3rls}
\end{figure}

which can all be trained with stochastic gradient descent. In the variants that have a critic, policy evaluation involves using Monte Carlo or TD learning methods to estimate $Q^\pi(s, a), A^\pi(s, a), V^\pi(s)$.

There are other types of reinforcement learning algorithms worth noting. One categorization of reinforcement learning methods depend on the depth in exploration search space, which consists of Monte Carlo (MC), Temporal Difference (TD) learning and Dynamic Programming (DP) (Figure \ref{fig:3rls}). The \textit{Monte-Carlo} (MC) methods learns from episodes of
raw experience without modeling the
environmental dynamics. It computes
the observed mean return as an
approximation of the expected return.
Similar to MC, the Temporal-Difference
(TD) Learning, which we already covers several variants (e.g. Q-Learning, Actor-Critic), is model-free approach that learns from episodes of experience. However, unlike MC, TD learning can learn from incomplete
episodes and hence don’t need to
keep track of all the episodes all the way up to the termination stage. When the model is fully known or estimated, following Bellman equations, we can use Dynamic Programming methods (such as value iteration) to iteratively evaluate value function and improve policy. It is a model-based methods. For interested readers, \cite{Sutton98} is the most respected textbook on reinforcement learning that covers all these topics. 

\subsection{Inverse reinforcement learning}

\begin{algorithm}[tb]

 \caption{Inverse Reinforcement Learning (IRL) Problem}
 \label{alg:irl}
\begin{algorithmic}[1]
 \STATE {\bfseries Initialize} $w$, the parameter for the reward function 
\STATE \textbf{for} each episode (on-policy expert trajectories) in the demonstration policy $\{(s^*_1, a^*_1), \cdots, (s^*_T, a^*_T)\} \sim \pi^*$ \textbf{do} 
\STATE {\bfseries } \quad \textbf{while} not converged, or a maximal step is not reached
\STATE {\bfseries } \quad \quad Solve the MDP environment using the current reward function $R_w(s, a)$ to generate reinforcement learned policy $\pi$ or behaviors
\STATE {\bfseries } \quad \quad Optimize for reward function parameter $w$ by minimizing the inconsistency between the observed behavior $\pi^*$ (from the expert) and the reinforcement learned behavior $\pi$
\STATE {\bfseries } \quad \textbf{end while}
\STATE {\bfseries } \textbf{end for}
 \end{algorithmic}
\end{algorithm}

\begin{figure}[tb]
\centering
    \includegraphics[width=.8\linewidth]{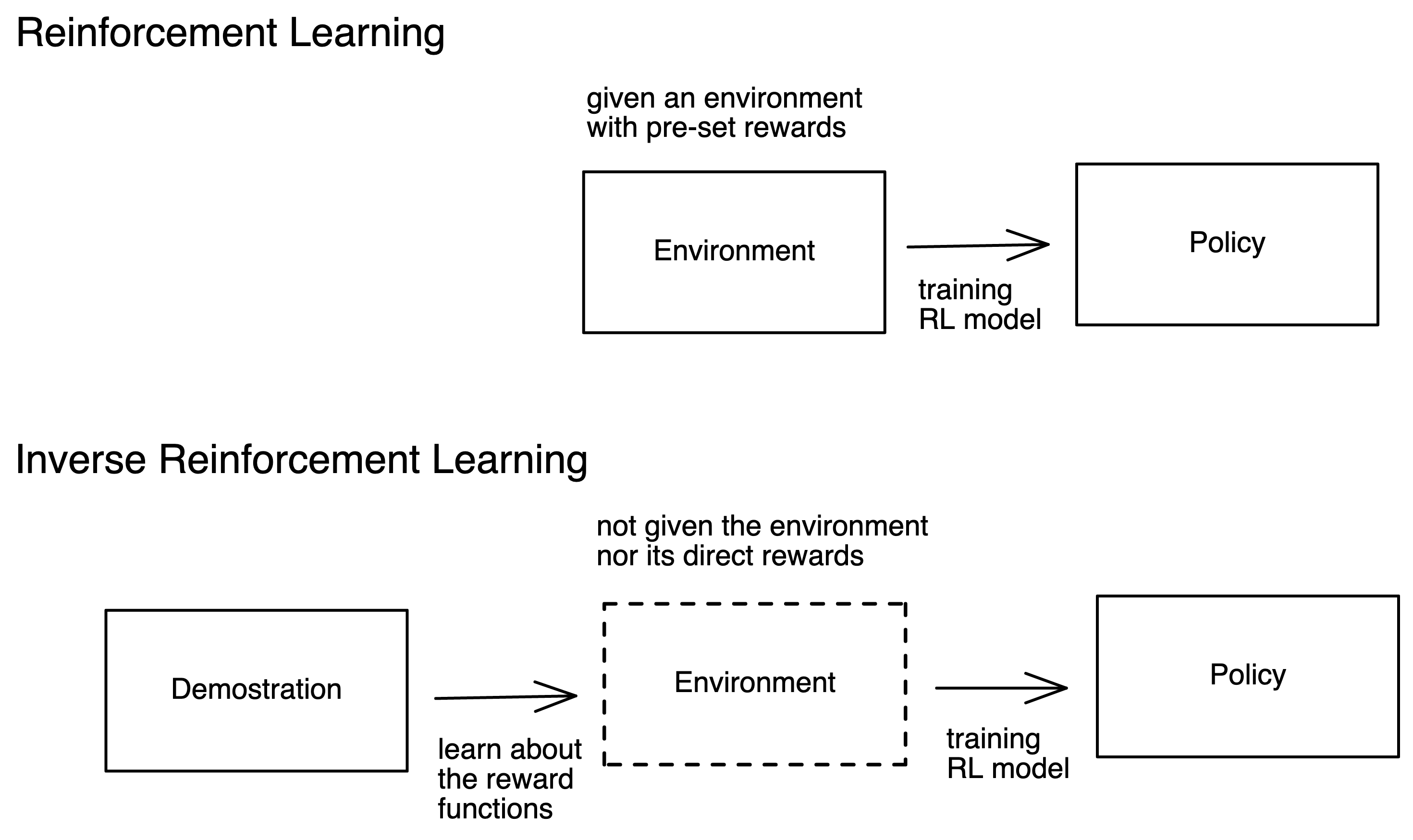}
\vspace{-1em}
\caption{Comparison of training reinforcement learning using the conventional reinforcement learning approaches vs using the inverse reinforcement learning.
}\label{fig:irl}
\end{figure}

If we compare the pipeline between the standard reinforcement learning and the inverse reinforcement learning in Figure \ref{fig:rl_classes}, they look very similar, both with the agent interacting with the environment with state emission, action taking and reward feedback. However, instead of given a simulation environment with pre-set reward functions for the agent to interact with, in inverse reinforcement learning, we only have the historical trajectories or experience of an expert interacting with an unknown environment, and the goal is to discover the reward functions of this environment (what is this world environment about? what motivates the agent to act the way it did? Can we reason about the goal or objective the expert is aiming for?). In addition, it is usually easier to obtain historical data of expert's action, rather than having an interactive systems to train forever. Thus, using these priors can potentially be powerful.

The inverse reinforcement learning is connected to the concept of inverse optimal control \cite{russell1998learning,ng2000algorithms}, a framework to model  or to ``imitate'' human behaviors. However, it is different from the standard imitation learning in the following way. standard imitation learning only copy the actions performed by the expert, without any reasoning about the outcomes or implications of the actions they imitated. It also doesn't necessarily characterize the salient properties of the behaviors. For instance, an expert can have multiple skills at different contexts. 
Inverse reinforcement learning, or here in this context, the ``human imitation learning'' aims to copy the intent of the expert, and thus, the policy it learns might take very different actions despite having the same intent.

We are given the state and action space, historical trajectories or samples from the expert policy $\pi^*$, and sometimes also its dynamic model, and the goal is to recover reward function, and sometimes also to use the learned reward to get an optimal policy (as in the reinforcement learning). Other than points covered above, here are several additional benefits. Since the policy is a mixture of the reward and the dynamics, to enable better transfer learning across tasks when learning from the demonstration, we want to decouple the goal from the underlying dynamics \cite{fu2020d4rl}. Inverse reinforcement learning might be a solution to extract out the reward part. Another user scenario would be to use inverse reinforcement learning to create additional training systems with learned rewards, as in Figure \ref{fig:irl}. If our end goal is to train a reinforcement learning agent that mimics the expert policy, but we don't have the same environment to train on, one way would be to first learn the reward functions from expert trajectories and then train our agent in the new environment we create using the learned reward functions.

The framework of inverse reinforcement learning is straightforward. 
As in algorithm \ref{alg:irl}, we model the observed behavior from the expert as the solution of an Markov Decision Process with unknown reward functions that is gradually optimized to converge to the underlying reward functions. We first initialize a parameter $w$ for the reward function $R_w$ we wish to optimize. This can be a linear weight, a neural net, a distribution over rewards, or any function approximation as we show before. Then, we iteratively optimize for $R_w$: at each step, we solve the MDP environment using the current reward function $R_w(s, a)$ to generate the reinforcement learned policy $\pi$ or behavioral trajectories; we compute the inconsistency between the observed behavior $\pi^*$ (from the expert) and the reinforcement learned behavior $\pi$ as a loss, for any optimization engine of $w$ to minimize against. We continue this two steps until it converged to a reasonable level of behavioral inconsistency.

Inferring the underlying reward functions directly from demonstration can be challenging. First, since many reward functions can explain the same behavior, it is an underspecified problem. Second, these historical trajectories of reward functions usually have unknown dynamics and large state-action space, which makes the optimization difficult. Third, it can be difficult to evaluate a learned reward, so we need to specify a reasonable metric. And lastly, the demonstrations may not be precisely optimal, and thus, it might be impossible to uniquely decompose the policy of irrational or suboptimal agents into the underlying reward function \cite{armstrong2018occam}.

\begin{algorithm}[tb]
 \caption{Maximum Entropy (MaxEnt) Algorithm}
 \label{alg:maxent}
\begin{algorithmic}[1]
 \STATE {\bfseries Initialize} $w$, the parameter for the reward function 
\STATE \textbf{for} each episode (on-policy expert trajectories) in the demonstration policy $\{(s^*_1, a^*_1), \cdots, (s^*_T, a^*_T)\} \sim \pi^*$ \textbf{do} 
\STATE {\bfseries } \quad \textbf{while} not converged, or a maximal step is not reached
\STATE {\bfseries } \quad \quad Solve for optimal policy $\pi(a|s)$ with value iteration in the environment parameterized by the current reward function $R_w(s, a)$.
\STATE {\bfseries } \quad \quad Solve for the state visitation frequencies $p(s | w)$
\STATE {\bfseries } \quad \quad Compute gradient $\nabla_w\mathcal{L} = \frac{1}{M} \sum_{\mathcal{T} \in \mathcal{D}} x_\mathcal{T} - \sum_{s\in S} p(s|w) x_\mathcal{T}$
\STATE {\bfseries } \quad \quad $w = w + \alpha \nabla_w\mathcal{L} $
\STATE {\bfseries } \quad \textbf{end while}
\STATE {\bfseries } \textbf{end for}
 \end{algorithmic}
\end{algorithm}

Despite these challenges, several strategies are proposed to solve inverse reinforcement learning. The first strategy is to optimize to \textit{maximize margin}. Similar to the intuition of support vector machines, the idea is to learn a reward function that better explains the demonstrated expert policy than alternative policies by a margin as large as possible. One simple formulation of the margin is the sum of the discrepancy between the expected value of the optimal action and the next best action over all states, given by \cite{ng2000algorithms}:

\begin{equation}
    \sum_{s\in S} Q^\pi(s, a^*) - \max_{\{a\in A|a \neq a^*} Q^\pi(s, a)
\end{equation}

where $a^*$ is the optimal action at state $s$. 
If we adopt the feature-based approach with function approximation, the margin is the difference between the expected value of the behavior from the observed expert trajectory and the largest expected values of the the behaviors of other trajectories that can be used to learn the feature weights $w$. We can obtain the expected value of a policy by multiplying the empirical state visitation frequency (SVF) from the demonstration data, which we denote $p(s|w)$, with the weighted features of the state obtained from the trajectory, which we denote $x_s$. As such, the margin becomes:

\begin{equation}\label{eq:margin}
     \sum_{(s, a) \in \mathcal{T}} p(s|w)w^\top x_s - \max_{\{\mathcal{T}'\in (S\times A)^l| \mathcal{T}' \neq \mathcal{T}} p(s|w)w^\top x_s
\end{equation}

where $\mathcal{T}$ denotes the trajectories, and $(S\times A)^l$ is the set of all possible trajectories of length $l$ defined by the state-action space. One can solve this optimization method with a linear program to retrieve the reward function that maximize this margin \cite{abbeel2004apprenticeship}. This margin optimization approach can produce the given policy as optimal output from the complete MDP \cite{arora2021survey}. Assuming that each demonstration trajectory is a distinct policy, the structured maximum margin prediction (MMP, \cite{ratliff2006maximum}) improves the optimization process by solving a quadratic program that is constrained to have positive margin value in equation \ref{eq:margin} and regularized with a loss term $l p(s|w)$ which measures the closeness between the demonstrated and alternative behaviors.

However, as in the overall challenges of IRL, in these maximum margin methods, matching of the feature counts can be ambiguous because a given policy can be optimal for many different reward functions, while the same feature counts can be obtained by many different policies.
This approach improves upon these limitations of the maximum margin approaches by solving the ambiguity of suboptimality with a maximum entropy principle to learn a distribution over behaviors parameterized by the reward function weights. The idea is to learn the reward function that provides the trajectory distribution constrained by the observed demonstrations with the maximum entropy:

\begin{equation}
    \max \sum_{\mathcal{T}\in(S \times A)^l} -p(\mathcal{T}) \log p(\mathcal{T})
\end{equation}

To avoid the exponential growth of the state-action search space, an alternative would be to find the reward function that maximizes the entropy of the distribution of all policies:

\begin{equation}\label{eq:entropy}
    \max \sum_{\pi\in(S \times A)^l} -p(\pi) \log p(\pi)
\end{equation}

A popular algorithm to optimize for this, is the Maximum Entropy (MaxEnt) \cite{ziebart2008maximum} algorithm, which introduces two constraints to equation \ref{eq:entropy}: the distribution of all demonstration trajectories should follow a probability distribution; and the expected feature count of the demonstration trajectories should satisfy:

\begin{equation}
    \sum_{\mathcal{T}\in \mathcal{D}} p(\mathcal{T})\sum_{t=1}^l \gamma^t x_s^t = \frac{1}{M}\sum_{i=1}^M \sum_{t=0}^\infty \gamma^t x_s^t
\end{equation}

where $\mathcal{D}$ is our available demonstration data, $x_s^t$ is the feature of the state $s$ at time $t$, and $M$ is the number of trajectories. There are two assumptions. First, the reward of the trajectory is a linear combination of the feature counts: 

\begin{equation}
    R_w^\mathcal{T} = w^\top x_\mathcal{T} = \sum_{s\in \mathcal{T}} w^\top x_s
\end{equation}

where $x_\mathcal{T}$ is the feature count of the trajectory $\mathcal{T}$. And the the probability of a demonstrated trajectory should be exponentially higher for higher rewards than lower rewards:

\begin{equation}
    p(\mathcal{T}) \propto e^{R_w^\mathcal{T}}
\end{equation}

As in algorithm \ref{alg:maxent}, the algorithm solves for a convex nonlinear optimization problem:

\begin{equation}
    \argmax_w \sum_{\mathcal{T}\in \mathcal{D}} \log p(\mathcal{T}; w), \text{where } p(\mathcal{T}; w) = \frac{e^{\sum_{(s,a) \in \mathcal{T}} w^\top x_s}}{Z(w)}
\end{equation}

\begin{figure}[tb]
\centering
    \includegraphics[width=\linewidth]{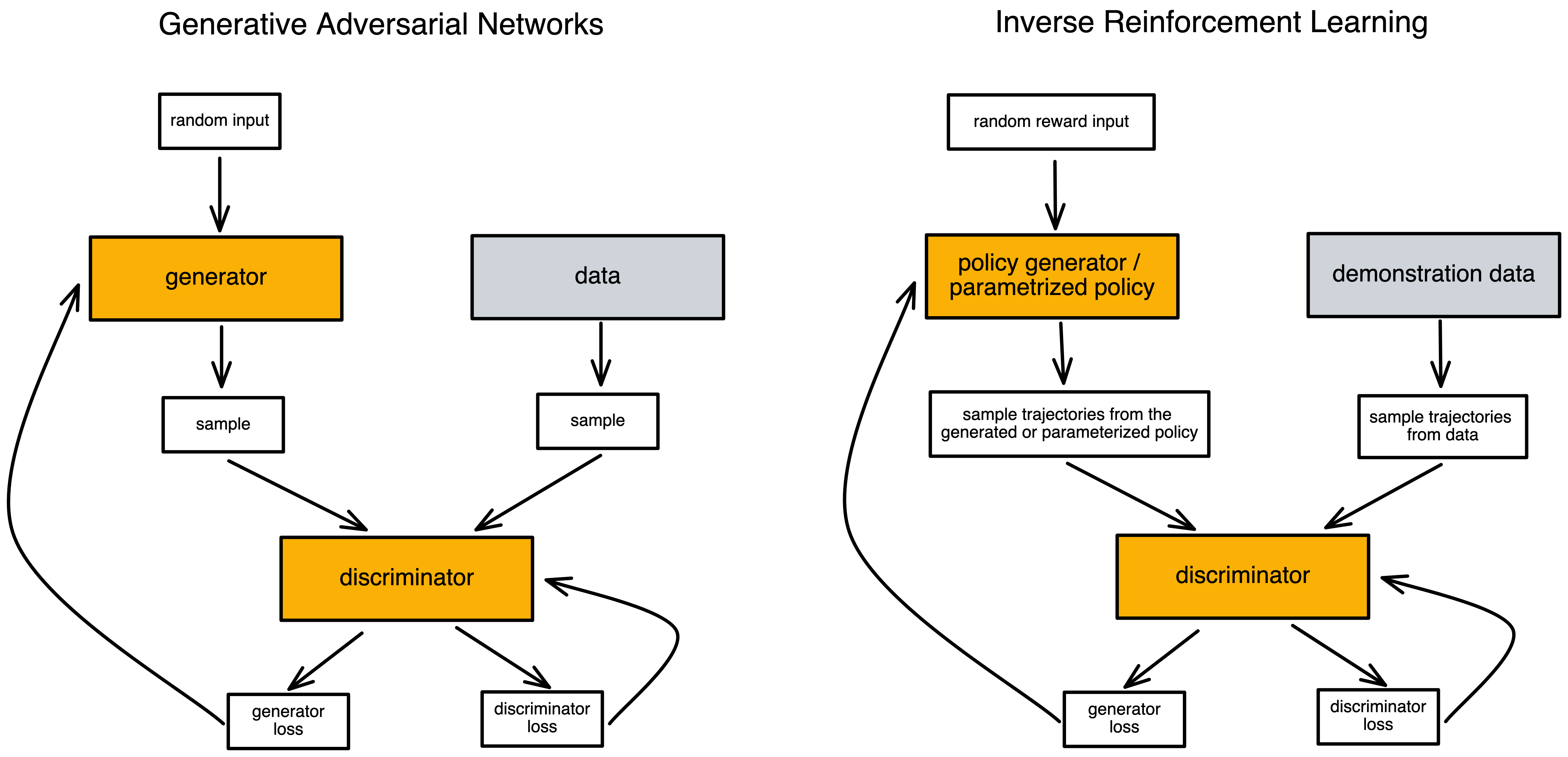}
\vspace{-1em}
\caption{Comparison of the generative adversarial networks and inverse reinforcement learning
}\label{fig:gan_irl}
\end{figure}

\begin{algorithm}[tb]
 \caption{Generative adversarial imitation learning (GAIL) Algorithm}
 \label{alg:gail}
\begin{algorithmic}[1]
 \STATE {\bfseries Initialize} $\theta_0$, $\psi_0$, the initial parameters for the policy and the discriminator
\STATE \textbf{for} $i=0, 1, 2, \cdots$ \textbf{do} 
\STATE {\bfseries } \quad Sample trajectories from the generator $\mathcal{T}_i \sim \pi_{\theta_i}$
\STATE {\bfseries } \quad Update discriminator from $\psi_i$ to $\psi_{i+1}$ with the gradient:
\begin{equation}\label{eq:d_grad}
    \hat{\mathbbm{E}}_{\mathcal{T}_i}[\nabla_\psi \log(D_\psi(s,a))] + \hat{\mathbbm{E}}_{\mathcal{T}_i}[\nabla_\psi \log(1- D_\psi(s,a))] 
\end{equation}
\STATE {\bfseries } \quad Update policy from $\theta_i$ to $\theta_{i+1}$ with the gradient: 
\begin{equation}\label{eq:p_grad}
    \hat{\mathbbm{E}}_{\mathcal{T}_i}[\nabla_\theta \log(\pi_\theta(a|s)Q(s,a))] - \lambda \nabla_\theta H(\pi_\theta)
\end{equation}
\STATE {\bfseries } \textbf{end for}
 \end{algorithmic}
\end{algorithm}

There is a connection between the maximum entropy IRL with the generative adversarial networks (GAN), which is a family of generative models which jointly and adversarially train a discriminator of synthetic and data samples and a generator of synthetic samples given a cost function trained by the discriminator  \cite{goodfellow2020generative}. In particular, there is a mathematical equivalence between sample-based maximum entropy algorithm and the generative adversarial networks, with respect to their similarity of both using the probability density of the generator as an evaluation metric and input to the dueling discriminator \cite{finn2016connection}. As in Figure \ref{fig:gan_irl}, similar to a GAN, inverse reinforcement learning attempts to optimize for a reward function that can generate or parameterize a policy which can sample out trajectories that is as indiscriminable to an expert trajectory sample from the demonstration as possible, and the evaluation metric for that has to be also trained as a discriminator which compares the samples under some principled or learned strategy (e.g. maximum entropy or margin). It involves a special form of discriminator, which in optimal case:

\begin{equation}
    D^*(\mathcal{T}) = \frac{p(\mathcal{T})}{p(\mathcal{T})+ q(\mathcal{T})}
\end{equation}

where we assume the generator is fixed with a density $q(\mathcal{T})$ and the $p(\mathcal{T})$ is the actual distribution of the demonstration data. Generative adversarial imitation learning (GAIL, \cite{ho2016generative}) and guilded cost learning (GCL,
\cite{finn2016guided}) are two IRL algorithms that optimize using a GAN-like setting. As in algorithm \ref{alg:gail}, for each iteration, we first sampel the trajectories from the generator $\mathcal{T}_i \sim \pi_{\theta_i}$, update the discriminator given the relative entropy (equation \ref{eq:d_grad}) and then update the policy using the Trust region policy optimization (TRPO) rule \cite{schulman2015trust} on the cost function $\log(D_\psi(s,a))$ (a KL-constrained natural gradient step as in equation \ref{eq:p_grad}).

For interested readers, \cite{arora2021survey} is a survey on various approaches of inverse reinforcement learning.

\subsection{Imitation learning and behavioral cloning}

The imitation learning or behavioral cloning aims to solve the policy learning with a supervised learning approach. As in algorithm \ref{alg:il}, we simply divide the expert demonstration into batches of state-action pairs, and treat these pairs as i.i.d. training examples for supervised learning using a loss function that characterizes the difference between the action taken by the expert and the action recommended by the supervised learning policy. Since the algorithmic space of this class of methods is largely in supervised learning, which can be highly customized to the specific problem setting and application domain, we will refer the readers to \cite{hussein2017imitation,osa2018algorithmic} for a survey on different types of supervised learning imitation learning methods. 

\begin{algorithm}[tb]

 \caption{Imitation Learning (IL) or Behavioral Cloning (BC) Problem}
 \label{alg:il}
\begin{algorithmic}[1]
 \STATE \textbf{Initialize} policy parameter $\theta$
\STATE \textbf{for} each episode (on-policy expert trajectories) in the demonstration data $\{(s^*_1, a^*_1), \cdots, (s^*_T, a^*_T)\} \sim D^*$ \textbf{do} 
   \STATE \quad {\bfseries }\textbf{for} batches of state-action pairs $(s^*, a^*)$ \textbf{do}
\STATE {\bfseries }\quad \quad Makes an observation $o^*(t)$ to the state $s^*(t)$ 
\STATE {\bfseries }\quad \quad Chooses an action given the current state or observation: $a_t =\pi_\theta({o^*}(t))$
\STATE {\bfseries } \quad \quad Update policy parameter $\theta$ using supervised learning by minimizing the loss function $D(a^*_t, a_t)$
\STATE {\bfseries } \quad \textbf{end for}
\STATE {\bfseries } \textbf{end for}
 \end{algorithmic}
\end{algorithm}

\begin{algorithm}[tb]

 \caption{Dataset Aggregation (DAgger) Algorithm}
 \label{alg:dagger}
\begin{algorithmic}[1]
 \STATE \textbf{Initialize} policy parameter $\theta$
\STATE \textbf{for} multiple episodes \textbf{do} 
\STATE {\bfseries } \quad train $\pi_\theta(a_t|s_t)$ from demonstration data $\{(s^*_1, a^*_1), \cdots, (s^*_T, a^*_T)\} \sim D^*$
\STATE {\bfseries } \quad run $\pi_\theta(a_t|s_t)$ to get new dataset $D_\pi = {s_1, \cdots, s_t}$
\STATE {\bfseries } \quad ask human annotators to label $D_\pi$ with actions $a_t$
\STATE {\bfseries } \quad aggregate $D^* = D^* \cup D_\pi$
\STATE {\bfseries } \textbf{end for}
 \end{algorithmic}
\end{algorithm}

\begin{algorithm}[tb]
 \caption{Behavioral Cloning with Demonstration Reward (BCDR) Algorithm}
 \label{alg:bcdr}
\begin{algorithmic}[1]
\STATE \textbf{for} each episode (on-policy expert trajectories) in the demonstration data $\{(s^*_1, a^*_1), \cdots, (s^*_T, a^*_T)\} \sim D^*$ \textbf{do} 
   \STATE \quad {\bfseries }\textbf{for} t = 1, 2, 3, $\cdots$, T \textbf{do}
\STATE {\bfseries }\quad \quad Agent makes an observation $o^*(t)$ to the state $s^*(t)$ 
\STATE {\bfseries }\quad \quad Agent chooses an action given the current state or observation: $a_t =\pi_t({o^*}(t))$
\STATE {\bfseries } \quad \quad Environment progresses to the next step $s^*_{t+1}$ given the action $a^*_t$
\STATE {\bfseries } \quad \quad Agent receives a reward feedback about both the environment reward and the consistency between its action $a_t$ and the expert's action $a^*_t$ at this state state $s^*_t$: $r_{a^*_t, a_t, s^*_t}(t) = r_{env}(a_t, s^*_t) + \lambda D(a^*_t, a_t)$
\STATE {\bfseries } \quad \quad Agent updates its policy $\pi_t$
\STATE {\bfseries } \quad \textbf{end for}
\STATE {\bfseries } \textbf{end for}
 \end{algorithmic}
\end{algorithm}

An innate challenge for imitation learning is the \textit{distribution shift} problem. As we are attempting to mimic the action trajectory, even a small change of actions in early rounds might lead to a large variance in later actions due to the amplification effect from the earlier actions. In other words, the $p_{data}(o_t)$ can be very different from $p_{\pi_\theta}(o_t)$.
To tackle this distributional shift problem, the \textit{Dataset Aggregation} (DAgger, \cite{ross2011reduction}) is an iterative algorithm to trains a deterministic policy with additional collections of dataset. The idea is simple, instead of trying to craft $\theta$ such that $p_{\pi_\theta}(o_t)$ can be as close to $p_{data}(o_t)$ as possible, we can simply collect more on-policy data such that $p_{data}(o_t)$ are closer to $p_{\pi_\theta}(o_t)$. As in algorithm \ref{alg:dagger}, the goal is to collect additional on-policy training data from $p_{\pi_\theta}(o_t)$ by simply running $\pi_\theta(a_t|o_t)$ and then asks human annotators to label $a_t$. These labeled actions combined with the state transition data collected from the learning policy $\pi_\theta$ are then appended or aggregated to the training dataset. Iteratively, the dataset collected would be closer and closer to an on-policy dataset that avoids the distributional shift issue.

However, we don't always have access to human annotations. One strategy is to provide goals as contexts.
Another strategy is to formulate supervision as rewards.
The \textit{Behavior Cloning with Demonstration Rewards }(BCDR, \cite{lin2022ipd}) is a reinforcement learning approach to directly optimize against the trajectory difference. In this setting, the agent first goes through a constraint learning phase. As in algorithm \ref{alg:bcdr}, during the learning, the agent is allowed to query the states and actions in the available demonstration data, and receive feedback about both the environment reward and whether or not the selected action $a_t$ matches the teacher's action $a^*_t$ (from the demonstration data):

\begin{equation}
    r_{a^*_t, a_t, s^*_t}(t) = r_{env}(a_t, s^*_t) + \lambda D(a^*_t, a_t)
\end{equation}

both and the consistency between its action $a_t$ and the expert's action at this state state $s^*_t$: 
During the deployment or testing phase, the goal of the agent is to maximize both the environment reward $r_{env}(a_t, s^*_t)$, and a unobserved $D(a^*_t, a_t)$, which models whether or not the chosen action matches which action the expert would have taken. Through the learning, the behavioral cloning algorithm aims to train reinforcement learning agents to mimic the expert behaviors in the demonstration.

For interested readers, \cite{hussein2017imitation} is a survey on various methods in imitation learning. \cite{kumar2021should} compares the imitation learning with offline reinforcement learning (a similar but different method we will introduce later in the emerging strategies section) and discusses the specific environments and dataset compositions to use one over the other.

\section{Reinforcement Learning Formulation for Speech and Language Applications}

In this section, we will cover a series of prototypical examples of how reinforcement learning can be effectively applied to major speech and language tasks, followed by a brief summary of other works in the specific task domain. Table \ref{tab:rlformulation} summarizes and compares the reinforcement learning formulations for several common natural language processing tasks. By examining the objective, reward, action space, and state space for each task, we can gain insight into the commonalities and differences between these tasks. For example, many of the tasks involve extracting meaning from natural language input, but the specific form of the input and output can vary widely. Additionally, different tasks have different reward signals and action spaces depending on the specifics of the problem. Overall, this table provides a useful reference for understanding how natural language processing tasks can be framed as reinforcement learning problems. We hope these case studies in the following sections motivate the readers to rethink their daily tasks as reinforcement learning problems and encourage discussions in these new research avenues. 

\begin{table}[tb]
\centering
\caption{The reinforcement learning formulations in different speech and language applications}
\label{tab:rlformulation}
\begin{tabular}{|p{0.15\linewidth}|p{0.15\linewidth}|p{0.15\linewidth}|p{0.15\linewidth}|p{0.15\linewidth}|}
\hline
\textbf{Task} & \textbf{Objective} & \textbf{Reward} & \textbf{Action Space} & \textbf{State Space} \\ \hline
Speech Recognition & Transcribe audio signal to text & Accuracy of transcription & Sequence of phonemes/words & Current audio input or transcription \\ \hline
Speaker Recognition / Diarization & Identify speakers and group speech into segments & Accuracy of speaker identification and segmentation & Speaker labels for each segment & Current audio input, speaker labels, speaker characteristics or speech features \\ \hline
Spoken Language Understanding & Extract meaning from spoken language and map to a machine-readable representation & Accuracy of extracted meaning & Set of possible meanings & Current audio input, dialogue history, external knowledge \\ \hline
Natural Language Understanding & Extract meaning from natural language text and map to a machine-readable representation & Accuracy of extracted meaning & Set of possible meanings & Current text input, dialogue history, external knowledge \\ \hline
Sequence Generation / Text-to-Speech Synthesis & Generate natural-sounding speech from text input & Naturalness and clarity of generated speech & Sequence of speech signals & Current text input, dialogue history, external knowledge \\ \hline
Natural Language Generation & Generate natural language text in response to a given prompt or input & Quality, coherence, and relevance of generated text & Set of possible natural language sentences & Current input prompt or dialogue history, external knowledge \\ \hline
Large Language Models & Improve language model behavior and responses & Relevance, engagement, diversity, and fluency & Set of possible words, phrases, or sentence structures & Context information, previous dialogue history, and generated text \\ \hline
Conversational Recommendation System & Provide personalized recommendations in response to user queries or preferences & Accuracy and relevance of recommendations, user engagement and satisfaction & Set of recommended items/actions & Current user preferences, search queries, historical data, user feedbacks \\ \hline
\end{tabular}
\end{table}

\subsection{Automatic speech recognition (ASR)}

\begin{figure}[tb]
\centering
    \includegraphics[width=\linewidth]{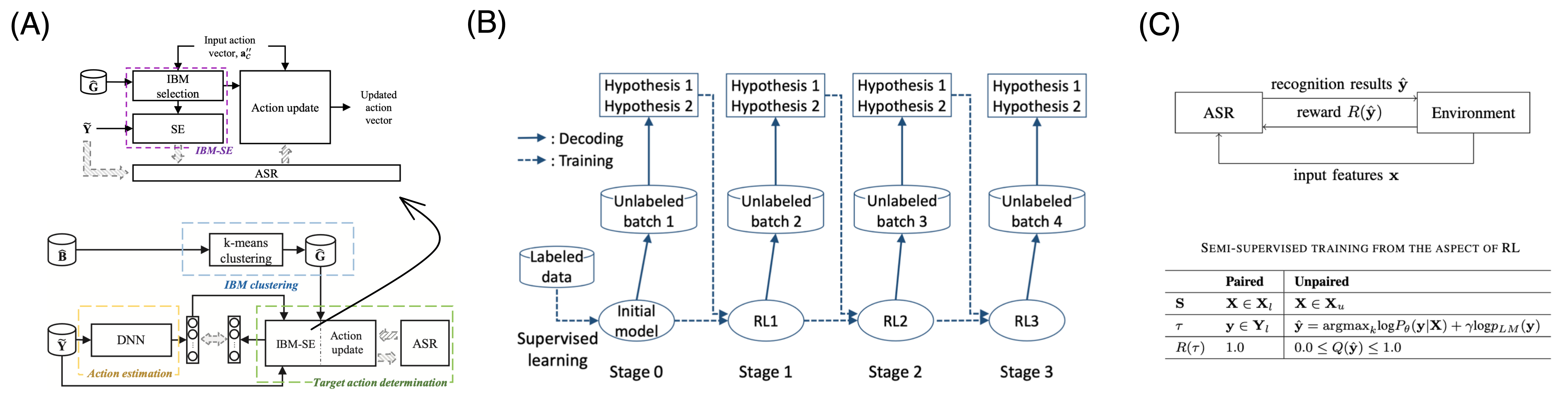}
\vspace{-1em}
\caption{Examples on speaker recognition: (A) speech enhancement using reinforcement learning \cite{shen2019reinforcement} (B) acoustic model selection using reinforcement learning \cite{kala2018reinforcement} (C) semi-supervised learning with reinforcement learning \cite{chung2020semi,rajapakshe2019pre}.
}\label{fig:asr}
\end{figure}





Automatic speech recognition (ASR) is an important component of many practical language processing systems. However, there are several challenges that can make it difficult to achieve accurate and reliable performance. One major challenge is the difficulty of training effective models for low-resource or marginalized languages. This is because there is often a lack of available data to train these models, which can result in poor performance and inaccurate results.
Another challenge is the impact of noisy environments on ASR systems. In real-world scenarios, speech signals can be distorted or obscured by background noise, making it difficult for the system to accurately recognize and transcribe spoken language. This can lead to errors and decreased performance, particularly in situations where high accuracy is critical.
Finally, collecting human feedback to improve ASR systems can be valuable, but can also be expensive and time-consuming. This can make it difficult to obtain large amounts of annotated data that can be used to train and improve the system.

Reinforcement learning can be a powerful tool for addressing some of these challenges in speech recognition. For example, one approach is to use curriculum learning to improve the effective learning from relatively small samples. This involves training the model on a gradually increasing level of difficulty, starting with simple examples and gradually moving towards more complex and challenging inputs. 
Another approach is to use speech enhancement techniques to generate cleaned and undistorted speech signals, which can improve the accuracy and robustness of the system in noisy environments. This can involve using signal processing techniques such as denoising and filtering, or training separate models to clean and preprocess the audio input before it is passed to the ASR system.
Finally, domain adaptation can be used to transfer learned representations from previous batches of audio signals to future ones. This can be particularly useful for low-resource or marginalized languages, where there may be limited data available to train the system. By using transfer learning techniques, it may be possible to leverage information from related languages or domains to improve the performance of the ASR system.
Here are some case studies:

\textbf{Case study} \cite{kuznetsova2021bandit}. In this work, the authors study the usage of curriculum learning to improve model training on automatic speech recognition on low-resource languages such as Turkish. Curriculum learning feed to the ASR model batches of data samples in a particular way such that they maximize the incremental learning performance at each batch. To do so, a bandit-based curriculum learning framework is proposed, where the reward feedback to the bandit is the prediction gain, $r_{PG} = L(x,\theta) - L(x, \theta')$, which is the loss before and after training on the current batch. By using an adversarial bandit algorithm, the Exponential-weight algorithm for Exploration and Exploitation (or more commonly known as EXP3) \cite{auer2002nonstochastic}, the curriculum generation procedure manage to effectively navigate among the exploration vs exploitation trade-off in using different batches of speech-text pairs which likely provides a performance boost in non-stochastic degree, and speed up the learning in ASR task with limited data.  

While curriculum learning with RL can help improve model training, the approach may be sensitive to the choice of reward function and the specific curriculum generation strategy. Designing an effective reward function that accurately measures learning progress can be challenging and require domain expertise. Moreover, the effectiveness of this approach may be limited by the complexity of the ASR task and the quality of the training data.

\textbf{Case study} \cite{shen2019reinforcement}. In this work, the authors introduce a reinforcement learning-based speech enhancement system for automatic speech recognition. As shown in Figure \ref{fig:asr}A, the raw audio signal can be noisy, and therefore, transforming them into the Fourier space, filtering the Fourier spectrum with a binary mask, and then applying inverse Fourier transform, can potentially generate a cleaned and un-distorted version of the speech signal. However, it is an innate challenge to find the ideal binary mask, since the audio data stream might come from different sources and thus contain different noise profiles. The solution is to first cluster the binary masks into major groups, and use a deep reinforcement learning agent to identify the ideal pool of binary masks to minimize the recognition errors. The reward signal here to update the agent is a scaled difference between the utterance-based error rates of the ASR result from the noisy original data and the ASR result from the speech enhanced samples spectrum-filtered by the binary mask picked by the reinforcement learning agent.

The RL-based speech enhancement approach can rely on the assumption that the pre-defined binary masks can effectively clean and enhance the speech signal. However, the performance of this approach heavily depends on the ability of the deep reinforcement learning agent to identify the optimal binary masks for various noise scenarios. Mismatched or incorrect mask choices could lead to suboptimal enhancement results and adversely affect ASR performance.

\textbf{Case study} \cite{kala2018reinforcement}. In this work, the authors propose a reinforcement learning-based hypothesis selection mechanism to pick the optimal hyperparameter for the acoustic processing embedding in batch-evaluated ASR systems. As in Figure \ref{fig:asr}B, they assume the data come in batches (e.g. batches of phone calls at different times of the day) and each batch may exhibit a slightly different voice pattern that would affect the effectiveness of model's domain adaption from one batch to another. As a result, they formulate the challenge as a reinforcement learning problem, where the actions are a few options of the update coefficient $\tau$ for a GMM-HMM acoustic embedding modulated by a deep learning-based MAP adaptation model. To specify the reward, they create two rival systems (one deep neural net that updates using reinforcement learning, and a baseline that doesn't). For each input utterance, a recognition hypothesis is sampled from each of two competing systems, and both of them are presented to the user. Then, the user provides a human-in-the-loop feedback by selecting the better hypothesis (which is the text prediction from the speech) among the two models. The reward feedback to the reinforcement learning agent is 1 if the deep reinforcement learning model is selected, and 0 if the baseline is selected by the user as providing the better hypothesis.

Using RL to select optimal hyperparameters for acoustic processing in ASR may introduce additional computational overhead, as the agent needs to explore the action space of hyperparameters. Moreover, the efficacy of reinforcement learning for hyperparameter selection can be influenced by the complexity of the task, the available labeled data, and the design of the reward signal. The performance gains achieved by this approach might be limited by the available labeled data and the diversity of voice patterns in different batches.

\textbf{Case study} \cite{chung2020semi,rajapakshe2019pre}. In these two works with similar ideas, the authors use the reinforcement learning as a semi-supervised pre-training mechanism to augment the ASR training with the unlabelled speech data. As in Figure \ref{fig:asr}C, traditional supervised learning ASR system needs paired data of the speech and their text transcripts in order to complete the supervised training. This component can also be accomplished using a reinforcement learning agent because, the supervised training procedure can be handled in terms of policy-based reinforcement learning because the gradient of cross-entropy loss for the one-hot target output of word tokens can be formulated as a special case of policy gradient:

\begin{equation}
    \nabla_\theta \log \pi_\theta(y_t | x_t) = \nabla_\theta \log \pi_\theta (a_t | s_t) \cdot 1.0
\end{equation}

On the other hand, the reinforcement learning can also handle unlabelled speech corpus without corresponding text labels. This is because reinforcement learning agents don't require speech-to-text paired corpus but requires a reward function, which can be a more relaxed condition. 
As a result, in face of unlabelled speech data, a reward function can still be effectively defined using a perplexity-based soft reward, which anti-correlates with the character error rate (CER) commonly used in ASR training.

Although RL offers a flexible way to use unlabelled speech data for pre-training ASR models, the definition of a reward function for unlabelled data may be challenging. The reward function needs to effectively correlate with the ASR task objective, such as minimizing character error rate (CER), without the direct supervision provided by labeled data. Additionally, the performance improvement achieved by using unlabelled data through reinforcement learning might be limited compared to fully supervised training.

\textbf{Other works}. \cite{tjandra2018sequence} trains a sequence-to-sequence model for ASR via policy gradient taking the negative Levenshtein distance as the reward. \cite{tjandra2019end} further improves this system by using token-level rewards instead of sentence-level rewards. \cite{karita2018sequence} uses policy gradient to train an encoder-decoder ASR system with sequence-level evaluation metric as its reward.

\textbf{Practicality and limitations}. 
Applying reinforcement learning to speech recognition faces challenges such as designing suitable reward functions, identifying optimal hyperparameters, and ensuring effective transfer of learned representations. The choice of reward function and the design of the reinforcement learning agent's action space can significantly impact the performance gains achieved. Moreover, reinforcement learning approaches may be limited by the quality and quantity of available data, making them more effective in scenarios with ample labeled or reward-driven data. Despite the potential benefits, these challenges underscore the need for careful consideration and evaluation when applying reinforcement learning to address the inherent complexities of ASR tasks.

\subsection{Speaker recognition and diarization}
 
\begin{figure}[tb]
\centering
    \includegraphics[width=\linewidth]{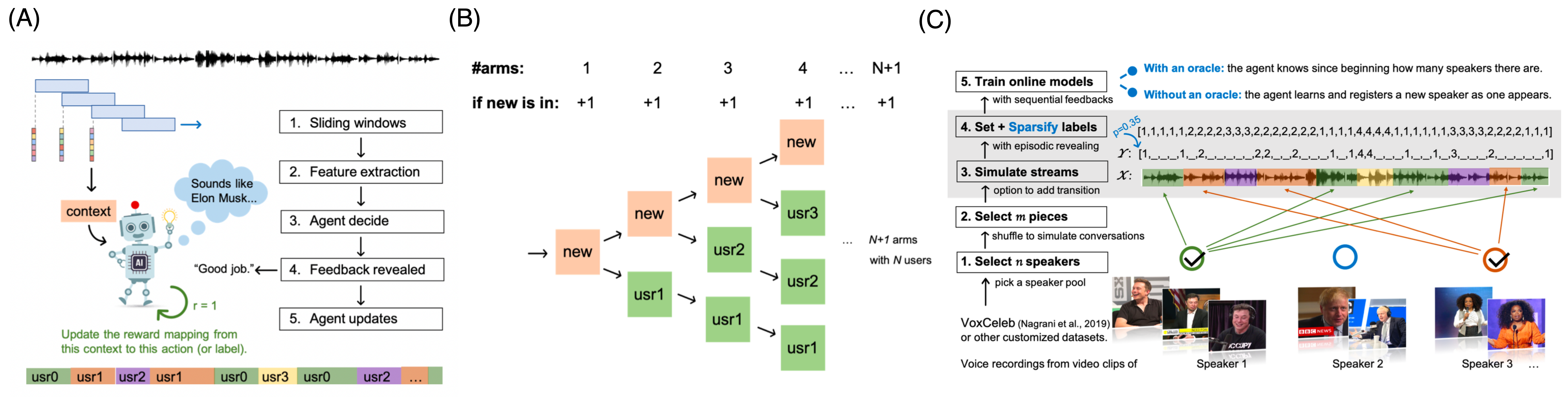}
\vspace{-1em}
\caption{Examples on speaker diarization: (A) The reinforcement learning problem \cite{lin2020voiceid} (B) the bandit solution for cold-start user problem \cite{lin2020speaker} (C) MiniVox: online speaker diarization benchmark for reinforcement learning agents \cite{lin2021speaker}.
}\label{fig:diarization}
\end{figure}





Speaker recognition and diarization are important tasks used in a wide range of applications such as speaker identification, speaker verification, and speech-based interaction. However, there are several challenges that can make it difficult to achieve accurate and reliable performance.
One major challenge is that existing speaker recognition models are usually pre-trained on large scale pre-registered user profiles. This can make it difficult to adapt these models to new or unknown speakers, which can result in poor performance in real-world scenarios. Additionally, modern multi-user teleconferences often have cold start new users, which can make it difficult for the system to accurately identify and track individual speakers.
Another challenge is that collecting human feedback to improve speaker recognition models can be difficult and expensive, particularly when the rewards are highly sparse. In addition, labelled data for low-resource populations may be very rare, which can make it difficult to train accurate models for these groups. Finally, diarization results can be hard to generalize to out-of-distribution environments, such as different contexts or acoustic conditions.

Reinforcement learning can be a useful tool for addressing some of these challenges in speaker recognition and diarization. For example, one approach is to use interactive learning with human feedback to improve the accuracy and performance of the system. This involves actively soliciting feedback from users and using it to update and refine the model in real-time.
Another approach is to make speaker recognition and diarization models lightweight and capable of learning on the fly. This can help improve the adaptability and flexibility of the system, particularly in scenarios with new or unknown speakers.
Finally, transfer learning or semi-supervised learning can be used to improve the accuracy and robustness of speaker recognition and diarization models, particularly in low-resource or marginalized populations. By leveraging information from related domains or data sources, it may be possible to improve the performance of the system even when labelled data is scarce.
Here are some case studies:

\textbf{Case study}  \cite{lin2020voiceid,lin2021speaker,lin2020speaker,lin2022voice}. In this work series, the authors consider the online speaker diarization task as a reinforcement learning (or fully online learning) problem, where at each time step, the agent uses the window-sliced acoustic features as its context or state to decide which user is speaking at the current moment (as its actions, Figure \ref{fig:diarization}A). The innate challenge for this problem is that, they assume the users can come and go at any moment, and the system can be deployed without any pretraining on users' voice profiles. The corresponding solution is to formulate it as a bandit problem with infinite arms, where one action arm always stands for a ``new user'', and some under-used action arms can demise with a recency principle (Figure \ref{fig:diarization}B). The online speaker diarization system updates the reinforcement learning policy with user feedbacks in a human-in-the-loop setting, which introduces an additional challenge that these reward feedback can be implicit and highly sparse. To effectively learn from sparse feedback, they propose to use the Background Episodic Reward LinUCB (or BerlinUCB \cite{berlinucb}), which is a semi-supervised learning bandit algorithm that uses clustering as self-supervision to provide pseudo-reward and update only contextual representations when the real reward is missing. To encourage the field to evaluate online speaker diarization as the reinforcement learning problem, they provide a testing benchmark called MiniVox (\cite{lin2021speaker,lin2020speaker}, Figure \ref{fig:diarization}B).

If historical data is available for pre-training, supervised and unsupervised approaches are still favored for their overall better performance, comparing to the on-the-fly reinforcement learning solution. A drawback of using RL for online speaker diarization is the challenge of collecting accurate reward feedback from users. Since the reward signal is based on implicit and sparse user feedback, there can be uncertainty in understanding the true reasons behind user preferences or decisions. This can affect the reliability of the reward signal used for updating the reinforcement learning policy, potentially leading to suboptimal learning outcomes. Moreover, designing a suitable reward function that accurately captures the quality of the diarization results and aligns with user preferences is critical, but can be challenging in dynamic and real-time scenarios.

\textbf{Other works}. It is a relatively new idea to apply reinforcement learning to solve speaker diarization directly. Other than the above works which use semi-supervised bandits on online speaker diarization problem, the follow-up work \cite{lin2022rldiarization} summarizes a complete spectrum of  reinforcement learning approaches in the speaker diarization problem.

\textbf{Practicality and limitations}.
Applying reinforcement learning to speaker recognition and diarization tasks offers potential benefits in addressing challenges such as adaptability to new speakers, handling cold start scenarios, and improving performance for low-resource or marginalized populations. However, a common drawback in these case studies is the challenge of obtaining accurate and reliable reward feedback, which is crucial for guiding the reinforcement learning process effectively. Sparse and implicit reward feedback can introduce uncertainty and potential biases, impacting the learning process and ultimately the system's performance. Despite the promise of reinforcement learning, these challenges highlight the need for carefully designed reward mechanisms and exploration strategies to ensure the success of the approach in speaker recognition and diarization applications.

\subsection{Spoken language understanding (SLU)}
 
\begin{figure}[tb]
\centering
    \includegraphics[width=\linewidth]{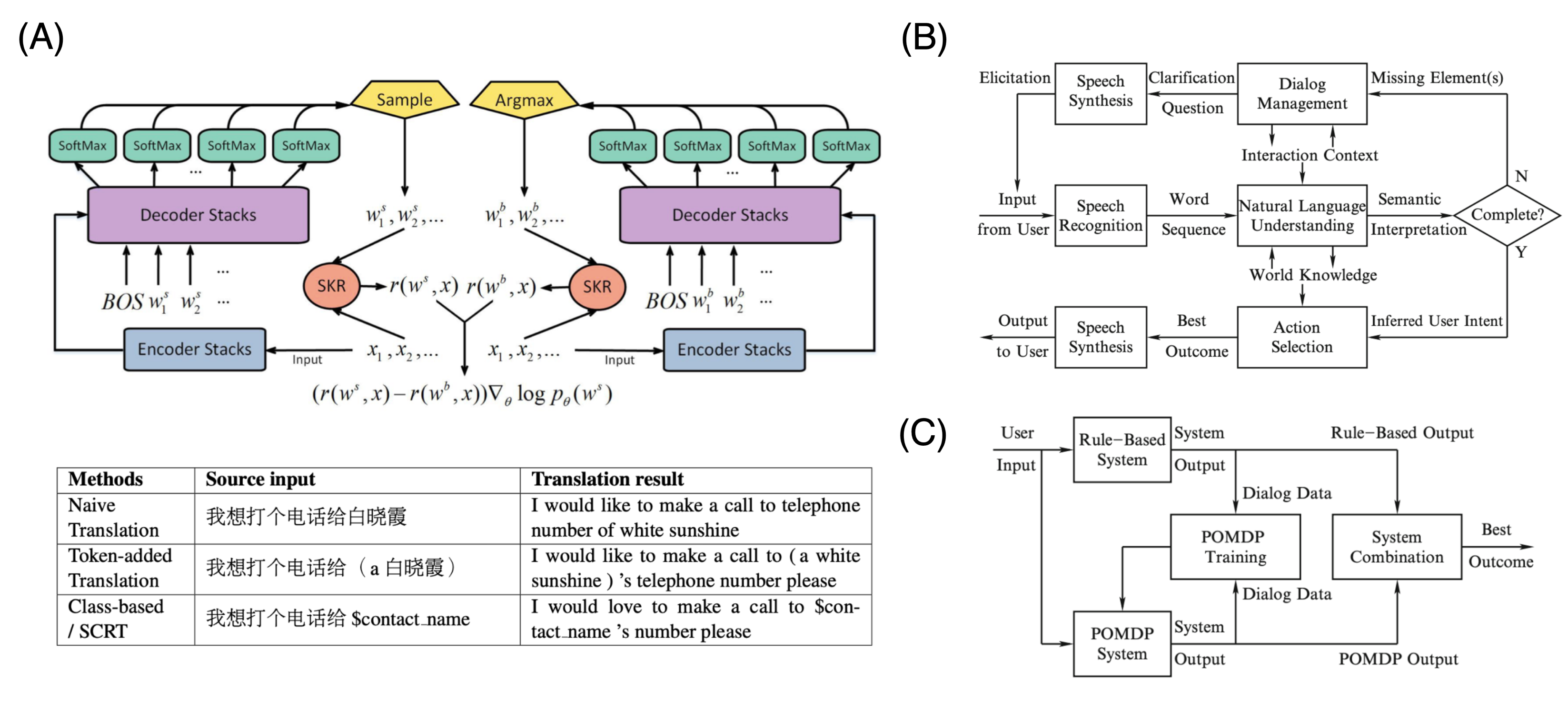}
\vspace{-1em}
\caption{Examples on spoken language understanding: (A) grammar slot matching with reinforcement learning in NMT language transference for SLU \cite{bai2018source}. (B) SLU components in Apple Siri \cite{bellegarda2014spoken} and (C) its training pipeline.
}\label{fig:slu}
\end{figure}



 

Spoken language understanding (SLU) is an essential component of modern day smart speakers, and is used to extract meaning and intent from spoken language input. However, there are several challenges that can make it difficult to achieve accurate and reliable performance.
One major challenge is the difficulty of transferring SLU models to new languages or domains. This can be particularly challenging if the grammatical structures of the new language differ significantly from the training data, which can result in poor performance and inaccurate results. Additionally, SLU is a complex task with many components, which can make modeling the state and action spaces very complex, particularly when the system is only partially observable.
Another challenge is that human refinement of SLU parsers can be interfering across different dialogue acts (DAs), which are utterances serving functions in the dialog. This can make it difficult to accurately assign and identify DAs, which can result in poor performance and inaccurate results.

Reinforcement learning can be a powerful tool for addressing some of these challenges in spoken language understanding. For example, one approach is to use reinforcement learning to auxiliarily optimize grammar slot matching. This involves training the model to identify and fill in missing or incorrect slots in the input, which can help improve the accuracy and completeness of the results.
Another approach is to consider the stages of training model components, and to use reinforcement learning to improve the performance of each component separately. This can involve training the ASR and NLU components separately and then integrating them using reinforcement learning to optimize performance and accuracy.
Finally, interactive learning can be used to adaptively assign dialogue acts, which can improve the accuracy and relevance of the results. This involves actively soliciting feedback from users and using it to update and refine the model in real-time, which can help improve the accuracy and effectiveness of the system.
Here are some case studies:

\textbf{Case study} \cite{bai2018source}. In this work, the authors adapts a neural machine translation (NMT) model to a new language using a policy-gradient reinforcement learning approach such that auxiliary requirements can be included as part of the reward signals (Figure \ref{fig:slu}A). In language transference between languages that are grammatically different, keeping the entity slots of word tokens aligned can be challenging. To tackle this issue, they propose a source-critical class-based method that directly measures the slot keeping ratio (SKR, a metric for evaluating the performance of slot transferring in SLU across language) as an incremental loss to bind to the reward function. More specifically, by optimizing over

\begin{equation}
    (r(w^s, x) - r(w^b, x)) \nabla_\theta \log p_\theta(w^s)
\end{equation}

where $x$ is the speech signal (source input to the NMT model), $w^b$ and $w_s$ are the word token generated by the baseline NMT model and the NMT model adapted with a reinforcement learning agent. The reinforced translation model updates given this final policy gradient by rewarding the translation candidates that generates a higher SKR over baseline. By using the relative SKR as the reward, the reinforcement learning agent adapts to obtain target language translations which can maintain both the semantic meanings and the slot information of the SLU-labeled sentences in the source language.

A drawback of using RL to adapt NMT models for new languages is the challenge of defining appropriate reward functions. Designing reward signals that accurately reflect slot keeping ratio (SKR) can be complex and require thorough understanding of the semantics and grammatical structures of both the source and target languages, as well as their innate ambiguity and dialects. An inaccurate or misaligned reward signal can lead to suboptimal adaptations and decreased translation quality.

\textbf{Case study} \cite{bellegarda2014spoken}. In this work, the author discusses the engineering choices made in the Siri system (Apple's personal intelligent assistant). The SLU system in Siri involves many critical components, ranging from speech recognition, natural language understanding, dialogue act recognition, dialogue management, action selection, to finally the speech synthesis to interact with the user using human-like voice feedback (Figure \ref{fig:slu}B). The training goal is to utilize statistical (e.g. deep learning-based) methods to the language understanding unit by integrating data-driven evidence collected from suitable training data in an optimal order, such that the intent of the user can be inferred efficiently and accurately. However, the feedback to the system are usually implicit and all computational components are proxies to the underlying inferred properties that can affect downstream tasks. As a result, this is a partially observable Markov decision process (POMDP) problem to the spoken dialogue systems. Performing a reinforcement learning training here is challenging because:
    (1) The internal state of the users that we wish to model is a complicated mixture of the user's goal, the user's input on the interface, and the historical dialogues and interactions.
    (2) This is further compounded by the speech-level uncertainty in the user's utterances and the systematic errors in upstream systems which can amplify and propagate the uncertainty into other computational entities.
    (3) For smooth user experience, the system should have a large enough action space that cover every possible system response, and as a result, the reinforcement learning policies should map from the complex and uncertain dialogue states to a large search space of possible actions.

One strategy they point out in this work is to first train rule-based system separately in order to gain more dialogue data to boost POMDP training for the reinforcement learning agent (Figure \ref{fig:slu}C).

\textbf{Case study} \cite{ferreira2016adversarial}. In this work, the authors aims to exploit the usage of user's annotating feedback for sequential training and refining of a zero-shot SLU semantic parser in an online way. The interactive system is formulated such that at each round of SLU, the user gets recommendations for their predicted intents, and they can be asked to do one of the three annotation tasks to help refine this SLU model: 1) Skip, meaning that the the user don't have to do anything; 2) YesNoQuestion, meaning that the user gets to confirm or negate the detected DAs in the best semantic hypothesis; and 3) AskAnnotation, meaning that the user is asked to annotate the incoming utterance. The reward signal to maximize is a mixture of the system improvement and the user effort.

A drawback of using RL for interactive SLU with user feedback is the potential uncertainty and variability in user annotations. The reliance on user annotations for refining the SLU model introduces noise and subjectivity in the reward signal, which can lead to challenges in accurately updating the reinforcement learning policy. Ensuring consistent and reliable user feedback is essential to avoid incorrect learning signals.

\textbf{Other works}. \cite{lin2022ispeak,lin2022ispeak_icassp} propose an interactive SLU system that uses a reinforcement learning agent to select both the optimal speech enhancement units to process the speech signals and the candidate intent prediction. The system can evaluate the correctness of the prediction by computing a similarity score between the intent prediction and the sparse annotation of the actual intent from the user. The reward signals to update the agent comes from the difference between the score of the agent and that of the baseline models that randomly sample predictions from the candidate pool. \cite{chen2020deep} uses a deep reinforcement learning approach to track speech dialogue state for online task-oriented spoken dialogue systems.

\textbf{Practicality and limitations}.
While reinforcement learning holds promise for improving SLU systems, these case studies highlight several common drawbacks. Designing appropriate reward functions that accurately reflect system performance or user preferences can be challenging, especially in scenarios involving complex language structures and user interactions. Additionally, the complexity of dealing with partially observable states in dialogue systems, as well as uncertainty and variability in user feedback, can introduce noise and errors in the reinforcement learning process. These challenges emphasize the need for careful reward engineering, exploration strategies, and handling uncertainties to ensure the success of reinforcement learning approaches in enhancing SLU performance.

 \subsection{Natural language understanding (NLU)}
 
\begin{figure}[tb]
\centering
    \includegraphics[width=\linewidth]{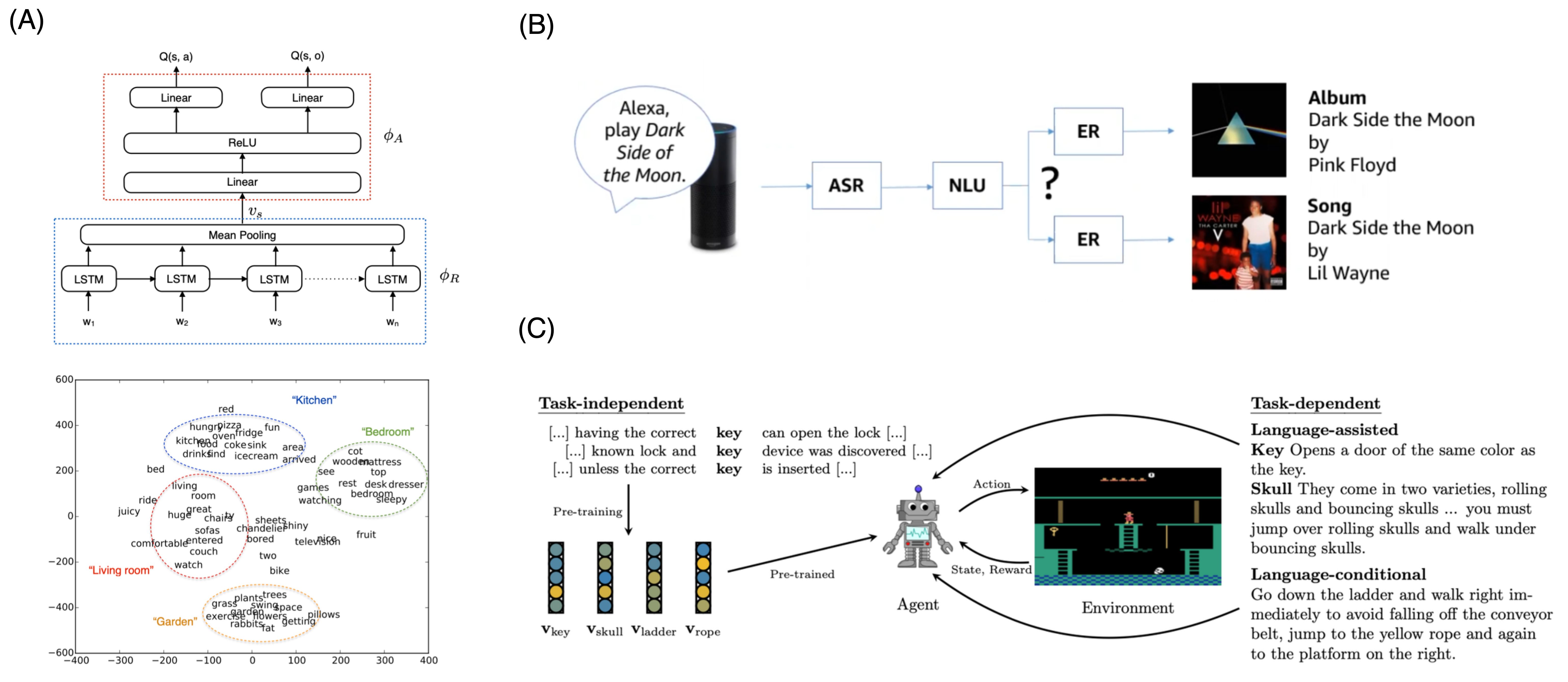}
\vspace{-1em}
\caption{Examples on natural language understanding: (A) NLU learned by deep reinforcement learning in text-based games \cite{narasimhan2015language}: top panel -- model architecture; bottom panel -- encoder word embedding. (B) NLU in recommendation systems with implicit feedback and ambiguous entities \cite{moerchen2020personalizing} (C) NLU-informed reinforcement learning \cite{luketina2019survey}.
}\label{fig:nlu}
\end{figure}





Natural language understanding (NLU) is a critical task used to extract meaning and intent from text-based information. However, there are several challenges that can make it difficult to achieve accurate and reliable performance.
One major challenge is that in many applied problems, the intents or interpretations from NLU are usually implicit or simply not given. This can make it difficult to accurately identify and extract the meaning and intent from the input, which can result in poor performance and inaccurate results. Additionally, NLU is usually a critical step to provide contexts for downstream tasks, and training it independently might not guarantee its performance in downstream tasks, as even a small variation can be amplified in later steps.

Reinforcement learning can be a useful tool for addressing some of these challenges in natural language understanding. For example, one approach is to train the whole system as a reinforcement learning problem and implicitly learn the natural language understanding task. This involves training the model to optimize a specific reward function, which can help improve the accuracy and effectiveness of the system in a wide range of applications and use cases.
Another approach is to provide contexts and personalization to natural language understanding training by training it together with downstream tasks in an end-to-end optimization. This can help improve the accuracy and robustness of the system, particularly in scenarios where the downstream tasks are closely related to the NLU task.
Finally, natural language understanding can in turn be used to improve reinforcement learning algorithms, by providing additional insights and context that can be used to improve the performance and accuracy of the system. This can help unlock the potential for natural language processing in a wide range of applications and use cases, by improving the efficiency and effectiveness of the algorithms and models used in these systems.
Here are some case studies:

\textbf{Case study} \cite{narasimhan2015language}. In this work, the authors study the neural network representation of deep reinforcement learning in text-based games and find that meaningful clusters of words emerge from it.
They propose a long short-term network (LSTM)-based deep Q network (DQN) that has a separate Representation Generator that takes a stream of words observed in the game's state $s$ as input and generates a vector representation, which is then fed into a an actor network consisting of scores for all actions and argument objects (Figure \ref{fig:nlu}A). The game states are hidden from the player, who only receives a varying textual description. In this reinforcement learning problem, the action space has two dimensions, the action and its argument object, whose policies are jointly trained with the representation generator given game rewards. The dimension reduction of the state encoder representation (Figure \ref{fig:nlu}A botton panel) shows that the words are grouped by contexts which can potentially be used for natural language understanding.

One potential drawback of using deep RL to extract neural network representation for specific tasks is the challenge of interpretability. While meaningful clusters of words may emerge from the representation, understanding the exact semantics and relationships between these clusters can be difficult, especially for deep RL-based decision making systems trained for high-stake scenarios such as finance, forensic or health-related data. 

\textbf{Case study} \cite{moerchen2020personalizing}. In this work, the authors use bandits to personalize the NLU given user features and implicit feedback. Consider a virtual assistant system, we give the system a speech command, it transcribes it using ASR, and detects intents (e.g. play music) with NLU and parse out necessary further information such as slots (e.g. album) and the slot values (Figure \ref{fig:nlu}B). Before executing the command of retrieving an entity (e.g. play a certain song), the system needs to perform the entity resolution (ER) process, which finds the best entity of the given type (e.g. album) an the slot values (e.g. ``dark side of the moon''). However, they can be ambiguous in all three levels (alternative ASR transcripts, multiple NLU intents, and multiple entities matching the same criterion). It is a bandit problem because we only receive implicit and bandit feedback (only revealed for the picked NLU interpretations). It is a contextual bandit problem, because the best NLU interpretation is dependent on the user's interest. We can also train it with feature-based reinforcement learning because the natural language understanding are usually trained with manually labelled utterances which can be additional signals to the agent.

Similar to the aforemnetioned bandit-based speaker diarization solution, the challenge of using bandit-based personalization for NLU is the uncertainty and subjectivity in sparse and implicit user feedback. Implicit feedback can be less informative and may not always accurately reflect the user's true intent or preferences. Incorrect or noisy feedback signals can lead to suboptimal personalization decisions and decreased performance in entity resolution and intent detection tasks.
 
\textbf{Case study} \cite{luketina2019survey}. In this work, the authors summarize different ways NLU can better inform reinforcement learning tasks (Figure \ref{fig:nlu}C). First, we have language-conditional reinforcement learning systems, where the text information are directly related to the task, such as following text instructions, parsing rewards from instructions, and reading languages in the observation or action spaces. Second, we have language-assisted reinforcement learning systems, where the languages are auxiliary but useful information not directly related to the task goal, such as communicating domain knowledge, and structuring policies. Finally, we have task-independent natural language understanding, such as finding real-world semantics and grammar, or finding the underlying storylines or intents of characters.

A potential challenge of using NLU to inform RL tasks is the complexity and diversity of natural language. Extracting meaningful and relevant information from natural language text can be challenging, especially when the language is ambiguous, context-dependent, or involves domain-specific jargon. Inaccurate or incomplete extraction of information can lead to incorrect or biased reinforcement learning decisions, such as in a medical or legal decision making scenario.

\textbf{Other works}. \cite{vogel2010learning} proposes a reinforcement learning solution that learns to execute navigation instructions expressed in natural language. \cite{narasimhan2015language} finds that LSTMs, if combined with reinforcement learning optimization, can learn a better representation in text understanding than bag-of-words approaches in capturing the underlying semantics of sentences.

\textbf{Practicality and limitations}.
While reinforcement learning offers promising approaches to address challenges in natural language understanding (NLU), these case studies highlight some common drawbacks. The complexity of language, potential ambiguity, and the challenge of designing accurate reward functions can limit the effectiveness of reinforcement learning in NLU tasks. Additionally, the uncertainty and subjectivity of implicit feedback and the lack of interpretability in neural network representations can introduce noise and uncertainty into the learning process. To successfully leverage reinforcement learning in NLU, careful consideration of these challenges and the development of strategies to mitigate them are crucial.

\subsection{Sequence generation and text-to-speech (TTS) synthesis}
 
\begin{figure}[tb]
\centering
    \includegraphics[width=\linewidth]{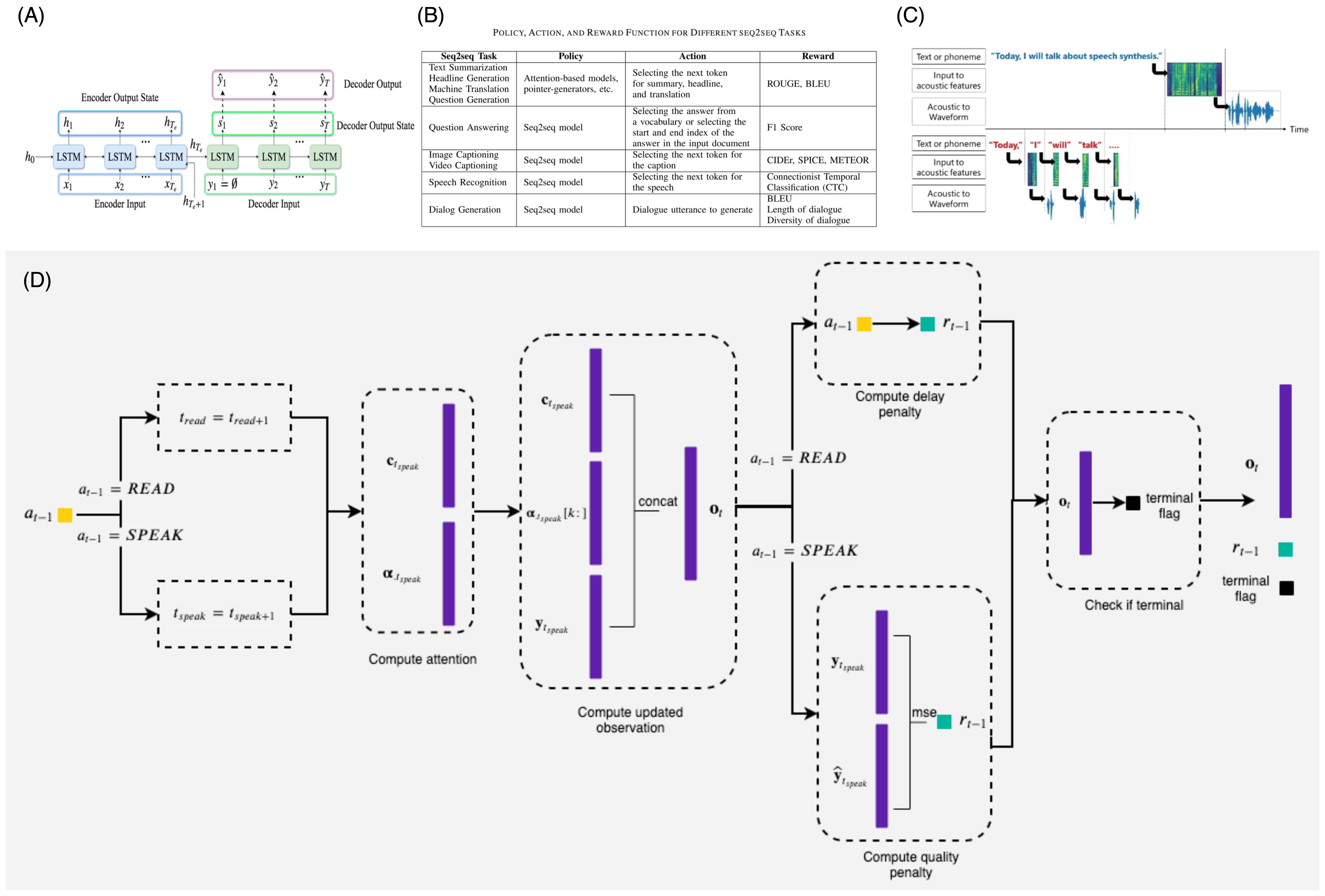}
\vspace{-1em}
\caption{Examples on sequence generation and text-to-speech (TTS) synthesis: (A) Seq2Seq model and (B) the reinforcement learning formulations of Seq2Seq models in sequence generation \cite{keneshloo2019deep}. (C) The incremental text-to-speech (TTS) problem and (D) its reinforcement learning solution \cite{mohan2020incremental}.
}\label{fig:sg}
\end{figure}





Sequence generation is an important task used in a wide range of applications such as text-to-speech (TTS) synthesis, language translation, and image captioning. However, there are several challenges that can make it difficult to achieve accurate and reliable performance.
One major challenge is that many Seq2Seq models suffer from exposure bias, which refers to the error accumulation during the output generation at test time since the model has never been exclusively exposed to its own predictions during training. This can result in poor performance and inaccurate results, particularly in scenarios where the output sequence is long and complex.
In addition, in the evaluation phase, many Seq2Seq models can suffer from an inconsistency between the training and testing datasets or domains. This can make it difficult to accurately generalize the model to new data or scenarios, which can result in poor performance and inaccurate results.
Finally, deep learning-based solutions in sequence generation can be slow in both training and deployment, which can make them unsuitable for real-time applications such as incremental text-to-speech (TTS) problems.

Reinforcement learning can be a useful tool for addressing some of these challenges in sequence generation and text-to-speech synthesis. For example, one approach is to use reinforcement learning to provide a more generalizable solution to sequence generation models by remembering long-term memories with state information. This can help improve the accuracy and effectiveness of the model, particularly in scenarios where the output sequence is long and complex.
Another approach is to use reinforcement learning to allocate resources in a dynamic way, which can enable high-quality real-time deployment of sequence generation and TTS models. This involves training the model to optimize a specific reward function, which can help improve the efficiency and effectiveness of the system in real-time scenarios.
Here are some case studies:

\textbf{Case study} \cite{keneshloo2019deep}. In this work, the authors argue that deep reinforcement learning can address two common issues in training sequence-to-sequence (Seq2Seq) models (Figure \ref{fig:sg}A). The first issue is exposure bias, which is the error accumulation during the sequence output generation during the test phase. This bias exists due to the difference of feeding sequence in the training and testing phases, as the model usually isn't exposed to its own predictions during training. The second issue is the inconsistency between the training and testing measurements and objectives. Reinforcement learning solves them by remembering long-term memories with a state representation, and thus removing the ground-truth dependency during training and using only the model distribution to minimize the objective (or equivalently, maximizing the rewards). It provides a nice overview of the related sequence generation applications in machine translation, text summarization, dialogue generation and many more. Figure \ref{fig:sg}B provide a few examples of formulating common sequence generation tasks into the policy, actions and rewards in reinforcement learning. Some common reinforcement learning agents used in this domain are policy gradient, REINFORCE, actor-critic, and Q-learning.


\textbf{Case study} \cite{mohan2020incremental}. In this work, the authors study the incremental text-to-speech (TTS) synthesis problem (Figure \ref{fig:sg}C), where instead of outputting the acoustic waveform after the entire text sentence is fed into the TTS model, the goal is to synthesis the acoustic waveform as soon as possible in real-time (i.e. synthesis by each word or phoneme). Since the text-to-acoustic features and speech synthesis both take time, the computing device can only perform one task at the same moment, reading the text, or speaking the speech. This is a challenging task for two reasons. First, since the TTS model cannot see the full sentence before the synthesis, the model has to have some sort of intelligence to predict future texts in order to generate speech that makes sense in its intonations and accents. Second, most state-of-the-art speech synthesis engines are usually deep learning-based, which can create latency that doesn't meet the real-time requirement. As a result, for time-sensitive tasks like simultaneous interpretation, solutions usually emphasize non-neural and more traditional architectures. As in Figure \ref{fig:sg}D, deciding when to read and when to speak can be a well-defined action space for a reinforcement learning solution. The reinforcement learning agent then learns from the reward function which is a mixed combination that trades off between the latency incurred during the synthesis and the quality of the synthesised speech output. 

A drawback of using RL for real-time incremental TTS synthesis is the trade-off between latency and quality. Optimizing the reward function to strike the right balance between generating speech quickly and maintaining high-quality output can be non-trivial. The RL agent may need to explore a wide range of latency-quality trade-offs before converging to an optimal solution, which makes certain application scenarios impractical.

\textbf{Other works}. \cite{liu2021reinforcement} proposes an interactive training paradigm that updates a reinforcement learning-based emotional text-to-speech synthesis model rewarded by high emotion discriminability measured by a speech emotion recognition system. \cite{gibson2022reinforcement} uses a multistage reinforcement learning improves the sample-by-sample tree coding of speech by modulating the exploration vs exploitation tradeoff in training the speech analysis and synthesis processes. \cite{jiang2020rl} uses deep reinforcement learning to generate music accompaniment as a duet.

\textbf{Practicality and limitations}.
Reinforcement learning offers potential solutions to challenges in sequence generation and text-to-speech synthesis tasks, such as exposure bias, inconsistency, and real-time deployment. However, the design and optimization of reward functions can be complex and require careful consideration. Balancing competing objectives, such as generating high-quality output while minimizing latency, can be challenging. Additionally, the effectiveness of reinforcement learning in addressing these challenges depends on the specific task and domain, and careful experimentation and tuning may be necessary to achieve optimal results.

 \subsection{Natural language generation (NLG)}
 
\begin{figure}[tb]
\centering
    \includegraphics[width=\linewidth]{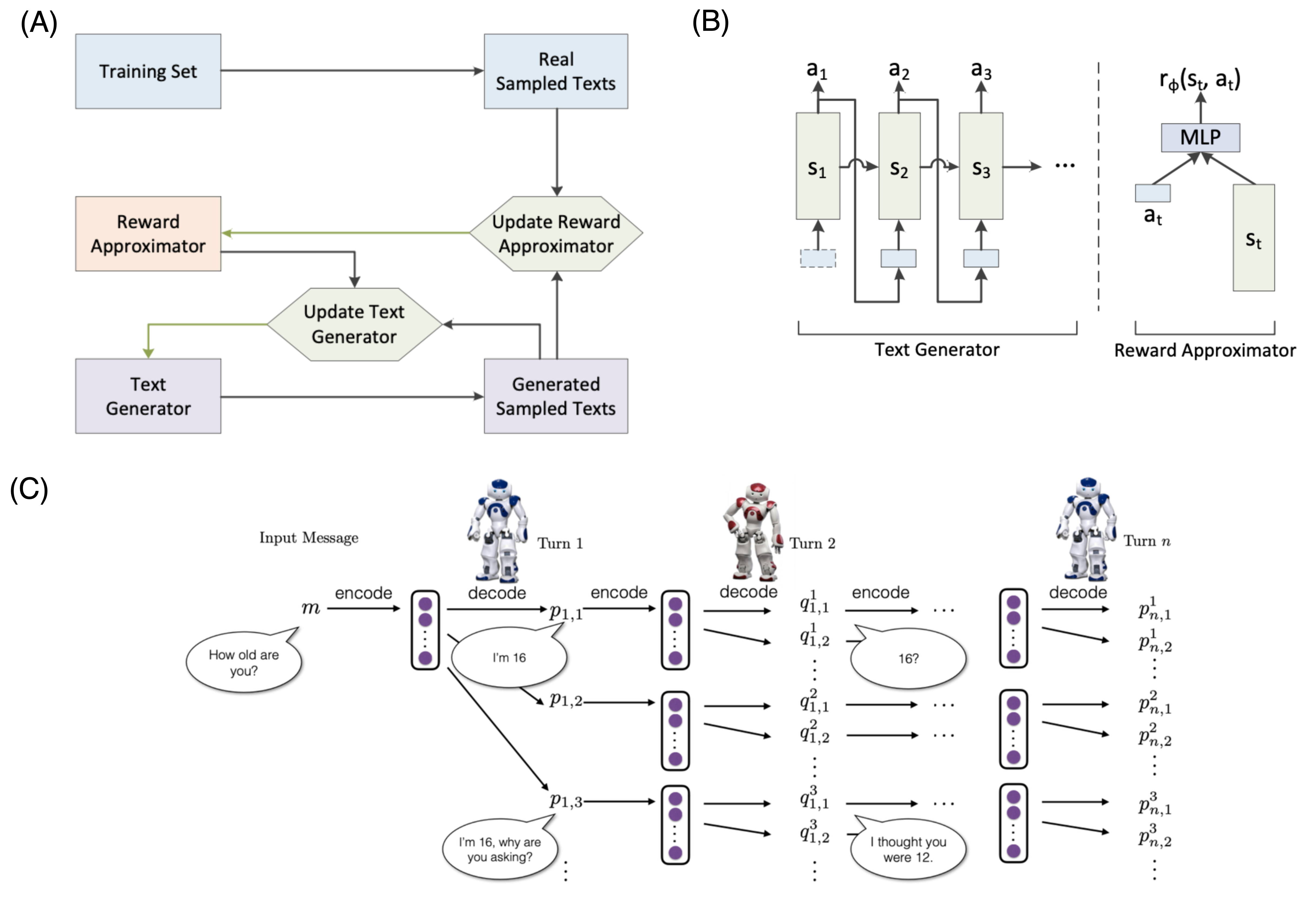}
\vspace{-1em}
\caption{Examples on natural language generation: (A) The inverse reinforcement learning framework of text generation \cite{shi2018toward} and (B) its sub-components. (C) The reinforcement learning problem in dialogue generation \cite{li2016deep}.
}\label{fig:nlg}
\end{figure}





Natural language generation (NLG) is a critical task used to generate human-readable text sequences from non-linguistic statistical representations of information. However, there are several challenges that can make it difficult to achieve accurate and reliable performance.
One major challenge is that text generation engines like Seq2Seq models can generate short and dull responses, such as ``idk'' or ``not sure.'' This can make it difficult to generate interesting and engaging responses, particularly in scenarios where the output needs to be engaging and informative.
In addition, NLG models can be short-sighted and only base their responses on the last few utterances. This can result in responses that are not fully contextualized or that lack nuance and depth.
Furthermore, the maximum likelihood objective is not necessarily representative of how humans converse with one another, and training NLG models can be expensive and time-consuming, requiring full supervision and labeled data. Finally, generated texts or dialogues can be repetitive and lack diversity, which can make it difficult to generate engaging and interesting content.

Reinforcement learning can be a useful tool for addressing some of these challenges in natural language generation. For example, one approach is to use inverse reinforcement learning to create more dense rewards and encourage diversity. This involves training the model to optimize a specific reward function, which can help improve the accuracy and effectiveness of the system in a wide range of applications and use cases.
Another approach is to use reinforcement learning to train dialogue models with customized rewards based on the problem being solved. This can help improve the accuracy and relevance of the responses, particularly in scenarios where the output needs to be tailored to the specific needs and preferences of the user or application.
Here are some case studies:

\textbf{Case study} \cite{shi2018toward}. In this work, the authors uses the inverse reinforcement learning to learn the underlying reward function or driving forces of the generative process of a text corpus and use it to generate realistic natural language samples (Figure \ref{fig:nlg}A). Existing NLG solutions are mostly based on adversarial generative models (e.g. SeqGAN \cite{yu2017seqgan}) followed by a reinforcement learning model, because as we points out in the sequence generation section, they can avoid the problem of exposure bias. However, the sequence generated by these GAN + RL solutions usually suffer from mode collapse in the adversarial generator and reward sparsity due to a perfect discriminator at the end of the training. Inverse reinforcement learning solves both by producing more dense reward signals (from the inferred reward functions) and encouraging more diversified texts as it uses entropy-regularized policy gradient (e.g. MaxEnt). As we pointed out earlier, inverse reinforcement learning is connected to GAN and can be formulated in similar architecture (Figure \ref{fig:nlg}A, B) by using a reward approximated generator followed by a discriminator to distinguish synthetic and real on-policy trajectories (which are text sequences). Under this architecture, the reward functions and the text sequence generator are jointly trained in an adversarial way.

One potential drawback of using inverse RL for text generation is the difficulty of accurately inferring the underlying reward function from the given text corpus, especially if the text corpus are not considered an expert policy and thus, can lead to equivalent solutions under a underspecified problem \cite{armstrong2018occam}. For instance, if the demonstration trajectories belong to human subjects with clinical conditions, as in \cite{lin2019split}, inverse RL can have a hard time reaching to an optimal solution. If the inferred reward function does not capture the nuances and complexities of human-generated text, it may lead to suboptimal results and fail to generate high-quality, realistic text samples.

\textbf{Case study} \cite{li2016deep}. When we think of dialogue generation, one might assume a necessity to use multiple (or two) agents to learn their interactions. However, due to the flexibility of the reward functions for policy gradient methods, in this work the authors aim to generate dialogue directly using a single deep reinforcement learning policy (Figure \ref{fig:nlg}C). The reinforcement learning problem is formulated the following way. The states are the concatenation of the previous two dialogue turns (which are also the inputs to the encoder in the Seq2Seq model). The actions are the dialogue utterances to generate (which have a infinite action space). The policy is a stochastic one, implicitly defined in the parameters of the encoder-decoder architecture. The reward has a combination of three desired properties: 1) the ease of answering (which is the the negative of likelihood of dull responses, such as ``Idk'' or ``no idea''); 2) the information flow or low repetitiveness in the generated dialogue (which is the information flow of the encoder representations); and 3) the semantic coherence (which is the pairwise mutual information between the two correspondents in a dialogue pair).

A potential drawback of training a single-agent dialogue generation model using deep RL is that it may not fully capture the complexities of multi-agent interactions in real-world dialogues. The simplified approach of using a single agent to generate dialogues might overlook the intricate dynamics that emerge from the interactions between multiple participants.

\textbf{Other works}. 
\cite{li2017paraphrase} proposes a paraphrase generation model that uses reinforcement learning to fine-tune deep networks of a generator and uses inverse reinforcement learning to train an evaluator on the similarity between phrases. \cite{dethlefs2011combining,dethlefs2011hierarchical} uses reinforcement learning to generate navigation instructions with rewards measured from either a hidden Markov model or a Bayesian network. \cite{riou2019reinforcement,riou2017online} proposes to use an adversarial bandit to adapt the NLG model online through direct interactions with user feedbacks. \cite{zhong2017seq2sql} proposes a reinforcement learning solution to generate structured queries from natural language with rewards from in-the-loop query execution over the database. 

\textbf{Practicality and limitations}.
As in the case studies, reinforcement learning offers promising solutions to challenges in NLG, such as generating engaging responses and ensuring context-awareness. However, using inverse reinforcement learning to infer reward functions for text generation may be prone to inaccuracies, potentially leading to suboptimal results. Similarly, training single-agent dialogue generation models may not fully capture the complexities of multi-agent interactions that occur in real-world conversations. These challenges highlight the need for careful formulation of reward functions and model architectures to ensure the effectiveness and accuracy of reinforcement learning-based solutions for natural language generation tasks.

\subsection{Large language models (LLM)}

\begin{figure}[tb]
\centering
    \includegraphics[width=\linewidth]{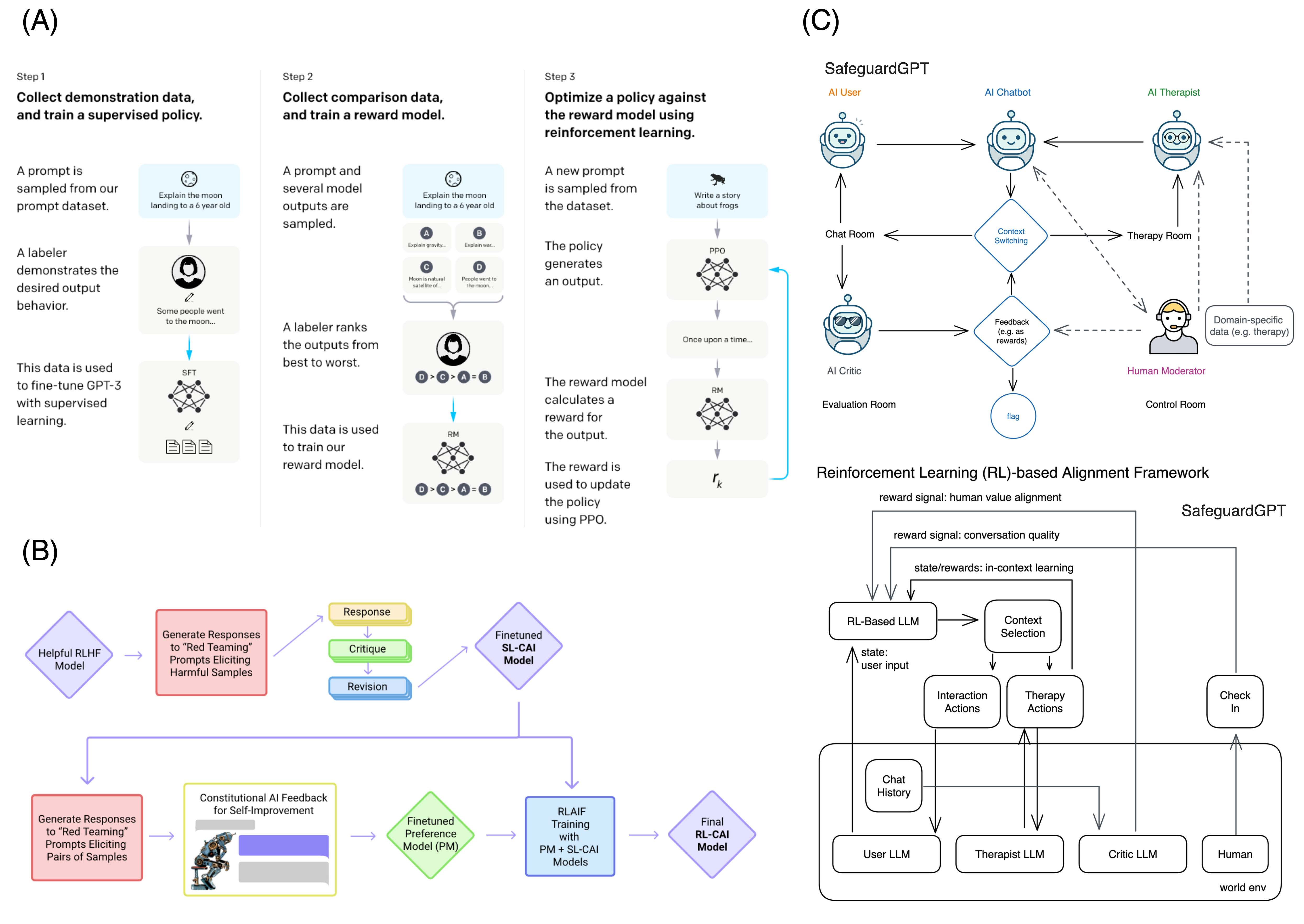}
\vspace{-1em}
\caption{Examples on large language models: (A) The three main steps of reinforcement learning with human feedback in InstructGPT \cite{ouyang2022training}. (B) The self-critique process of constitutional AI \cite{bai2022constitutional}. (C) The SafeguardGPT framework and its reinforcement learning alignment pipeline \cite{lin2023towards}.
}\label{fig:llm}
\end{figure}

There is a growing interest of large language models (LLMs) such as GPT-3 \cite{brown2020language}, PaLM \cite{chowdhery2022palm}, and ChatGPT which perform human-level performance in NLG task, which is why we separate it out as an individual section. These LLMs are typically trained using maximum likelihood estimation (MLE) to generate text that matches a given input. However, MLE-based methods suffer from various limitations, such as generating repetitive or uninteresting responses, and not taking into account the broader context of the conversation. 

RL provides a way to improve the performance of large language models by training them to optimize a specific reward function. In the context of NLG, the reward function can be defined based on various metrics, such as diversity, fluency, relevance, and engagement. For example, the reward function can be designed to encourage the model to generate diverse and engaging responses, while avoiding repetitive or irrelevant content. RL can also enable human feedbacks to fine-tune the LLMs to be more human-like and conversational in question answering type of scenarios.

\textbf{Case study} 
\cite{ouyang2022training}. With the introduction of human moderators or annotators, the LLM can be tuned with Reinforcement Learning from Human Feedback (RLHF) \cite{christiano2017deep,stiennon2020learning,ouyang2022training}, which involves using human feedback in the form of rewards to update the parameters of an LLM. As one of the first work in this direction, InstructGPT \cite{ouyang2022training} was proposed to align GPT-3 with users' intended outcomes. As in Figure \ref{fig:llm}A, the process involves assembling a collection of human-crafted instances showcasing the desired output behavior, and then, based on these demonstrations, GPT-3 is fine-tuned first using supervised learning. Following this, a reward model is constructed through the ranking of model-generated results ranging from the most favorable to the least. Employing this reward model, further refinement of the model is achieved via RL utilizing PPO. Results suggest that if trained with RLHF, smaller LLMs (e.g. 1.3B parameters) can generate more desirable results comparing to significantly larger ones (e.g. 175B).

A potential drawback of using RLHF for tuning LLMs is the need for human-generated rewards and rankings. Constructing a reliable reward model based on human preferences can be subjective and time-consuming. The quality of the reward signal heavily depends on the accuracy of human annotators, which may introduce bias or inconsistencies in the training process.

\textbf{Case study} 
\cite{bai2022constitutional}. Similar to RLHF, one can also tune LLMs using Reinforcement Learning from AI Feedback (RLAIF). Constitutional AI \cite{bai2022constitutional} refers to AI systems that are designed to comply with a set of ethical principles, similar to how democratic societies are governed by a constitution. The authors suggest using AI feedback as a mechanism for ensuring that the AI system remains within the boundaries of its ethical principles. Similar to RLHF, it involves both a supervised learning stage and a RL stage (Figure \ref{fig:llm}B). In the supervised stage, the model is refined based on the revisions generated alongside the output samples and self-critiques. In the RL stage, a secondary model is used to assess the qualities of both the original output sample and the output sample from the refined model. The difference between this two samples are treated as a guiding reward signal to train the LLM in a RL process.

In the case of RLAIF for ethical LLMs, a potential drawback is the challenge of defining the boundaries and principles that guide the AI system's behavior. Ensuring that the AI feedback accurately reflects the ethical principles can be complex and may require ongoing human oversight to prevent unintended consequences.

\textbf{Case study} 
\cite{lin2023towards}. 
As a special hybrid case of RLAIF and RLHF, SafeguardGPT \cite{lin2023towards} is proposed to adjust chatbot LLM behaviors by using psychotherapy as a framework to identify and mitigate toxicity. The SafeguardGPT framework involves human moderators and 4 different AI/LLM agents, including an AI Chatbot, an AI User, an AI Therapist, and an AI Critic. The Chatbot and User interact in the chat room, while the Therapist guides the Chatbot through a therapy session in the therapy room. Human moderators can control the sessions and diagnose the Chatbot’s state in the control room. Lastly, the Critic evaluates the quality of the conversation and provides feedback for improvement in the evaluation room as an RL process (Figure \ref{fig:llm}C), either entirely closed loop or human-in-the-loop). Since it involves an self-adaptive autonomous agent consisting of a group of AI agents, and thus, can benefit from group thinking and self-reflection through cross-talking among the agents. By incorporating psychotherapy and feedback mechanisms, results suggest that SafeguardGPT improves chatbots' communication skills, empathy, and emotional intelligence. 

While the SafeguardGPT framework offers a comprehensive approach to improve chatbot behavior, it involves multiple AI agents and human moderators, making it potentially complex to implement and manage. Coordinating interactions between different AI agents and ensuring their alignment with human moderators' intentions can be challenging and may require continuous monitoring and adjustment.

\textbf{Other works}. Other than using collaborative principles as in SafeguardGPT and Constitutional AI, one can train LLMs using adversarial techniques, as proposed in the Red Teaming \cite{perez2022red}, where one LLM is trained to identify and expose weaknesses in another LLM's language generation capabilities, as a more punitive approach. In this example, RL is used to maximize the expected harmfulness elicited in the Red LLM. 

In adversarial approaches like Red Teaming, where one LLM is trained to expose weaknesses in another LLM, there is a risk of generating harmful or inappropriate content. The adversarial nature of the training process might lead to unexpected and unintended negative behaviors, reducing the reliability and safety of the generated outputs.

In addition to these teaming and grouping of AI agents and human users, we can use RL to personalize LLMs with prior knowledge, such as existing datasets (e.g., psychotherapy transcripts, social forum interactions, online rating websites) to pre-train individual LLMs used as the AI Therapist, AI User, and AI Critic in the example work above. This can help develop more effective, safe, and ethical AI chatbots that can be integrated into various domains, such as customer service, education, and healthcare. 

\textbf{Practicality and limitations}.

While reinforcement learning offers promising ways to enhance the performance and behavior of large language models, there are potential drawbacks and challenges. These include the subjectivity and bias in human-generated rewards, the complexity of defining and enforcing ethical boundaries, the challenges of coordinating interactions among multiple AI agents and human moderators, and the potential for unintended negative outcomes in adversarial training approaches. Balancing these drawbacks with the benefits of improved LLM behavior requires careful consideration and continuous monitoring.

 \subsection{Conversational recommendation systems (CRS)}

\begin{figure}[tb]
\centering
    \includegraphics[width=\linewidth]{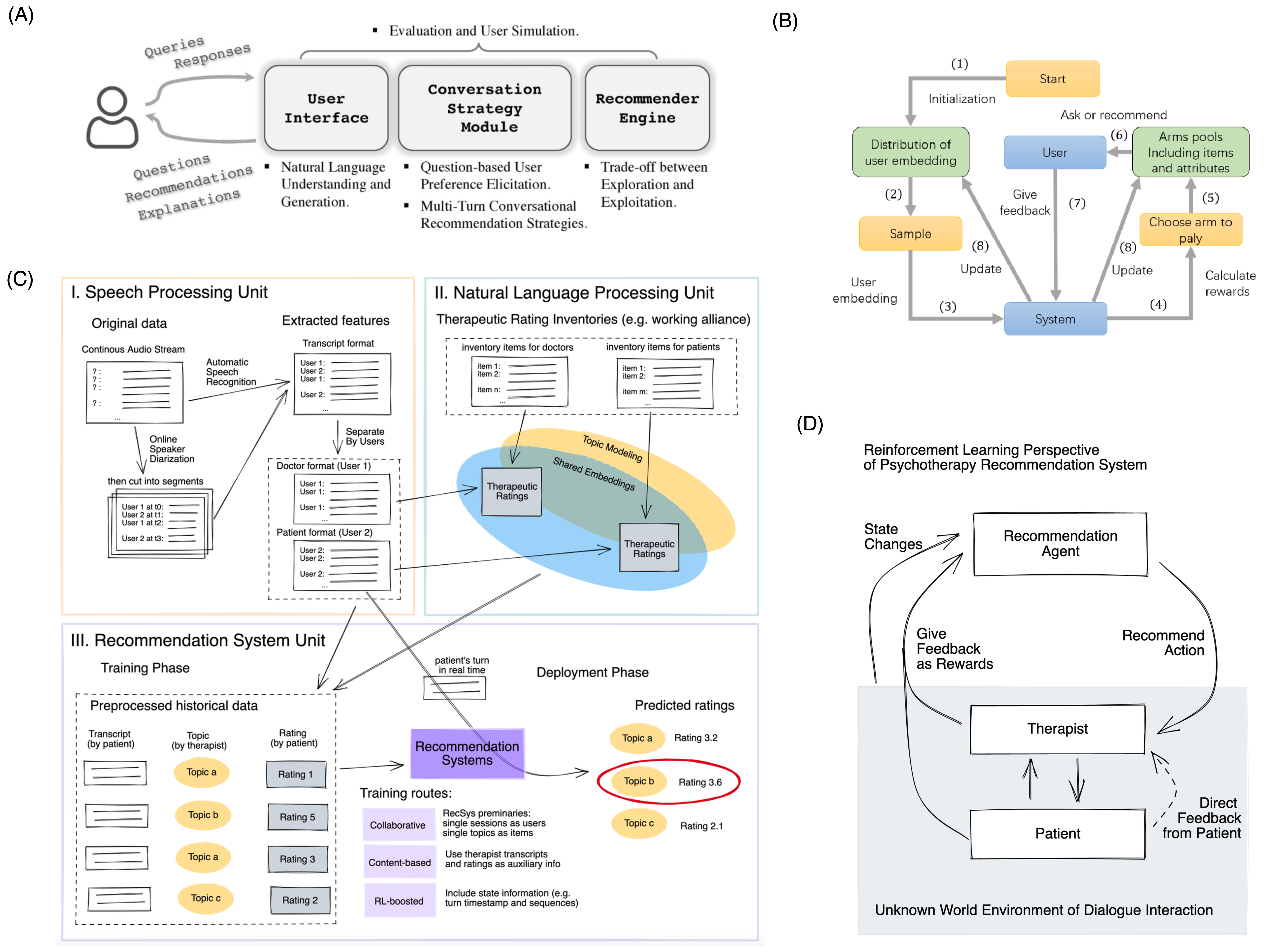}
\vspace{-1em}
\caption{Examples on conversational recommendation systems: (A) The major components of conversational recommendation systems \cite{gao2021advances} and (B) a bandit formulation of it \cite{li2021seamlessly}. (C) The major components of a dialogue topic recommendation in psychotherapy setting and (D) its reinforcement learning formulation \cite{lin2022supervisor}.
}\label{fig:rs}
\end{figure}





Conversational recommendation systems (CRS) are designed to enable dynamic communication between users and recommendation systems via natural language or speech interactions, and involve multiple language understanding components such as preference query, multi-turn conversational recommendations, and dialogue understanding and generation.
However, there are several challenges that can make it difficult to achieve accurate and reliable performance in conversational recommendation systems. One major challenge is that recommendation systems are often independently trained from natural language or speech components, which can limit their ability to take into account the contexts afforded by ambiguity. This can result in recommendations that are not tailored to the specific needs and preferences of the user.
Another challenge is that deep recommendation systems using existing natural language and speech components can be effective, but can be hard to generalize to cold-start users or items. This can limit the usefulness and applicability of the system, particularly in scenarios where new users or items are being added to the system over time.
Finally, directly applying the RL strategies in non-conversational recommendation systems to conversational ones may not handle sparse rewards very well in dialogue systems. This can make it difficult to accurately optimize the system to provide accurate and relevant recommendations to the user.

Reinforcement learning can be a useful tool for addressing some of these challenges in conversational recommendation systems. For example, one approach is to use bandits or reinforcement learning solutions to handle cold-start problems in recommendation systems. This involves training the system to optimize a specific reward function, which can help improve the accuracy and effectiveness of the system in a wide range of scenarios.
Another approach is to use speech and natural language components to parse real-time features as dense reward for reinforcement learning. This involves training the system to optimize a specific reward function, which can help improve the accuracy and effectiveness of the system in a wide range of scenarios, particularly in scenarios where the user is interacting with the system in real-time.
Here are some case studies:

\textbf{Case study} \cite{gao2021advances}. In this work, the authors summarize a bandit approach to tackle the conversational recommendation system setting (Figure \ref{fig:rs}A). In this example, the user interacts with a user interface powered by natural language understanding and generation units with queries and responses to the system. The recommendation model must strategize questions, recommendations and explanations in human-understandable multi-turn dialogues which balances the exploration vs exploitation trade-off. Intuitively, it can be formulated as a bandit problem where a specific set of questions or items can be selected as action arms in order to minimize the number of questions asked, maximize the click rate  or maximize the smooth experience in the interactions \cite{christakopoulou2016towards}. This objective relates to the knowledge of the user's preferences and the questions' quality in different aspects. A contextual bandit solution (such as LinUCB) can effectively a reward mapping from these properties by asking the user about one or more attributes and treating them as contexts \cite{zhang2020conversational}. Other than using reinforcement learning to select recommendation action, one can also use reinforcement learning to determine the timing to ask attributes or make recommendation. \cite{li2021seamlessly} proposes a bandit solution to dynamically and automatically alternate asking questions about attributes with recommending items given the user embeddings (Figure \ref{fig:rs}B).

Similar to the previous examples of bandit-based solutions in speaker diarization and SLU, a potential drawback of using a bandit approach to tackle conversational recommendation systems is that it may not fully capture the complexities of dynamic user preferences and evolving conversation contexts. The bandit model might struggle to adapt quickly to changing user preferences during the conversation, leading to suboptimal recommendations that do not align with the user's true interests. 

\textbf{Case study} \cite{lin2022supervisor,lin2023help,lin2023psychotherapy}. In this work, the authors propose a real-time conversational recommendation system in psychotherapy to recommend discussion topics to the therapist given previous patient-therapist dialogues. An unsupervised learning inference method \cite{lin2022deep,lin2022deep2,lin2022knowledge} is applied to annotate the therapeutic working alliance between the patient and therapist in each dialogue turn which predicts long-term clinical outcome and serve as the reward for each dialogue pair. The action space are the most common topics mined from neural topic models over historical data \cite{lin2022neural} and each dialogue turn are labelled by the topic label in a maximum likelihood principle (Figure \ref{fig:rs}C). They train a deep reinforcement learning policy to directly map from previous and current dialogue state to the best topic to recommend for the next turn, in order to maximize the cumulative therapeutic working alliance. As in Figure \ref{fig:rs}D, an interesting perspective to view this conversational recommendation system is that while the recommendation agent is driven by reinforcement learning, the therapist (and even patient) have their own agencies governed under the reinforcement learning principles. As such, the feedback loop are two folds: the patient can directly offer feedback to the therapists, given the feedback, the therapist may adjust his or her internal model to weigh on the quality of the suggestions made by the recommendation agent. 

While using RL to optimize the selection of discussion topics in psychotherapy is promising, it may not fully capture the nuances and complexities of therapeutic interactions and disease progression. The model's recommendations might not fully account for the sensitivity and emotional states of patients and therapists, potentially leading to recommendations that are technically relevant but emotionally inappropriate. Societal implications and ethical considerations \cite{lin2022computational} should be taken carefully when interpreting the insights from these clinically deployed RL models.

\textbf{Other works}. 
\cite{he2016deep} proposes a reinforcement learning solution that tackles a combinatorial recommendation problems by predicting the popularity of social media posts as the values of interdependent sub-actions. \cite{he2017reinforcement} is a follow up work that uses a two-stage Q learning approach and incorporate the global context represented by discussions in an external knowledge source into the state representations.

\textbf{Practicality and limitations}.
Reinforcement learning provides valuable solutions to challenges in conversational recommendation systems, including handling cold-start issues and real-time interactions. However, applying bandit or reinforcement learning strategies may struggle to fully capture the dynamics of user preferences and the emotional nuances in human conversations. The simplified decision-making process of reinforcement learning models might not always align with the complex and context-dependent nature of conversational interactions. These challenges emphasize the importance of integrating human-centric considerations and domain expertise when designing reinforcement learning solutions for conversational recommendation systems.

\section{Emerging Reinforcement Learning Strategies}

\subsection{Deep reinforcement learning and bandits}

Deep Reinforcement Learning methods combines the recent advancements of deep learning with reinforcement learning solutions. In the implementation level, these algorithms use deep neural networks to represent the value function, the policy function, or the world model. To optimize these models from the data, one can apply most deep learning optimization strategies, such as stochastic gradient descent, to optimize the value function, the policy function or the model in an end-to-end fashion.

For instance, the deep learning variant of the Q-learning algorithm is the deep Q networks (DQN) \cite{mnih2013playing}. Intuitively, it represent the Q value function by a deep Q network with weight $w$: $Q(s,a,w) \simeq Q^{\pi}(s,a)$. The objective function is defined by the mean squared errors in Q values:

\begin{equation}
    \mathcal{L}(w) = \mathbbm{E}[(r + \gamma \max_{a'} Q(s', a', w) - Q(s, a, w))^2]
\end{equation}

where $r + \gamma \max_{a'} Q(s', a', w)$ is the target that we want our Q function to converge to. Then, we can compute the gradient for Q-learning:

\begin{equation}
    \frac{\partial\mathcal{L}(w)}{\partial w} = \mathbbm{E}[(r + \gamma \max_{a'} Q(s', a', w) - Q(s, a, w))\frac{\partial Q(s,a,w)}{\partial w}]
\end{equation}

which can be optimized end-to-end with most deep learning optimization methods, such as the stochastic gradient descent. However, the naive Q-learning approach for deep Q network can be unstable and reach oscillatory or diverging solutions, for three reasons. First, the data is sequential, and in other words, the experience trajectories are non-iid and correlated with their successors. Second, the policy can change drastically with very small changes to the Q-values, and as a result, yielding oscillating policies and data distributions that swing back and forth. Third, the scale of the rewards and Q values are usually unknown, and thus can be amplified to very large values when trained with backpropagation.

There are a few solutions to solve these stability issues when training deep Q networks as Q-learning. Experience replay is a strategy that pools many episodes of experience at each time steps together as a replay memory to train the Q-learning in a off-policy way. By learning from all past policies with these replay memories, it uses iid samples and breaks the correlation of experience trajectories.
The second strategy is to freeze the target by using two networks, a Q evaluation network and a target network which we freeze during training. This strategy avoid the oscillation in our solutions and break the correlations between the Q-network and the target network. The third strategy is to clip the rewards or normalize the networks, such that the networks can learn from robust gradients.

There are many deep reinforcement learning algorithms growing as a popular field, so we will only briefly cover a few. The deterministic policy gradient (DPG, \cite{silver2014deterministic}) solves reinforcement learning problems with continuous action spaces using the deterministic policy gradient, i.e. the expected gradient of the action-value function integrated over the state space, or over both the state and action space in stochastic case. The deep deterministic policy gradient (DDPG, \cite{lillicrap2015continuous}) extends DQN and DPG into a model-free actor-critic method that replace DPG's stepwise optimization with experience replay and a ``soft'' target network which slowly tracks the learned network weights. Using an off-policy batch-based optimization approach, DDPG introduces exploration by injecting noise to the actor policy.
The trust region policy optimization (TRPO, \cite{schulman2015trust}) introduces the KL divergence between the new policy and the old policy as a trust region constraint, and solve the constrained optimization problem approximately using sample estimates from either single-path or rollout trajectories. The proximal policy optimization (PPO, \cite{schulman2017proximal}) improves upon TRPO by using KL-divergence as a penalty instead of a constraint. 

Similarly, we can also introduce deep learning into bandit solutions.  \cite{collier2018deep} proposes a deep contextual multi-armed bandit by binding a Thompson sampling mechanism on top of a Bayesian neural network such that the inference time dropout and weight posterior sampling are modeled as the exploration vs exploitation tradeoff.
\cite{guo2020deep} proposes a similar method of deep Bayesian bandit in recommendation system setting that approximates the uncertainty measurements of the bandit predictions by employing a bootstrapped neural network with multiple heads and dropout units.
Another approach is to directly use a neural network to represent the context.
For instance, \cite{zhou2020neural} proposes a neural contextual bandit which replaces LinUCB's linear mapping with a neural network-based random feature mapping.

For interested readers, \cite{li2017deep} is a good introduction to various types of deep reinforcement learning algorithms.



 

    
\subsection{Batched and offline reinforcement learning}

Most use scenarios of reinforcement learning that we introduce earlier are on-policy and purely interactive. However, many real-world reinforcement learning application systems have access to a large set of historical data, such as the prior user's behavioral trajectories, chat dialogues or purchase history. It would be a waste to not use them for off-policy training. Offline or batched reinforcement learning studies the reinforcement learning methods that use previously collected data without additional online data collection from on-policy interactions with the environment \cite{levine2020offline}. In this setting, usually we have access to the experience data collected once with some (potentially unknown) behavior policy (which we denote $\pi_{\beta}$, where $\beta$ refers to the data buffer, or also known as the replay buffer). In the training phase, we use the buffer data from the data buffer $\beta$ to learn a policy $\pi$. This training process doesn't have access to or interact with the MDP. This learned policy $\pi$ would only be only deployed into the task MDP after being fully trained. One relevant example is training a deep reinforcement learning from human dialogue data \cite{jaques2019way}. 

The offline reinforcement learning is the data-driven version of the reinforcement learning problem. And the optimization objective is still to maximize the expected future reward. What differentiate it with our reinforcement learning formulation earlier is that it doesn't have the ability to interact with the environment to collect additional transition experience using the demonstration behavioral policy. Instead, the transition data are given as a static historical records of experience for the agent to learn its best policy. On the first glimpse, it resembles the imitation learning which we discuss in earlier sections, which simply adopts a supervised learning approach to learn from the demonstration. Offline reinforcement learning, on the other hand, needs to first comprehend the dynamical system underlying the unknown MDP from the static transition data and then potentially use this world model to learn an optimal policy which can obtain the largest cumulative reward in deployment phase.

It is a nontrivial problem because existing reinforcement learning solutions which we previously introduce, despite its flexibility to learn from off-policy dataset, often fall short of their performance to learn entirely from offline data without online interaction. This is due to a few reasons. First, they cannot effectively explore the states that are rare or not available in the transition history and the actions that lead to those states. If the dataset doesn't contain connected state transition trajectories that touch high-reward regions, the agent would be unlikely to find those high-reward region. One possible way would be to manually inject exploration, but that would face our second issue: they cannot correct for out-of-distribution estimate via interactions. For an effective exploration, agents need to perform counterfactual queries, i.e. the question that what may happen if the agent were to take a series of actions different from the ones they have seen in the dataset. However, this is problematic, because as we discuss in the last section, deep reinforcement learning solutions use mechanisms like experience replay to create iid samples, which may limit the power of our learned policy to generate good yet potentially different actions, from those observed in our available dataset. In online setting, we can simply try it out and correct for it. In offline setting, we don't have this luxury. Third, like the challenge we discuss in imitation learning, we would face the problem of distributional shift when we perform a series of counterfactual queries. In other words, since we might not know the distribution of our behavioral policy in the dataset, it is likely that our agent (characterized by its policy function, value function and world model) is trained under one distribution, but evaluated in an entirely different distribution, since the small changes in visited states for the new policy can be amplified over time (or steps) such that the two distributions vary by a large degree. Lastly, the deep learning-based function approximation we use in these agents can exacerbate these issues due to their high-dimensional and expressive nature.

In order to evaluate how well a reinforcement learning perform given the historical data, importance sampling is used to perform offline evaluation with respect to either the return of a given policy, or the policy gradient of the offline policy gradient methods. In short, we use importance sampling to compute $J(\pi)$ with the trajectories sampled from $\pi_{\beta}$ as the off-policy evaluation. There are multiple variants of importance sampling estimators that people use, such as per-decision importance estimator normalized by the weights \cite{precup2000eligibility}, doubly robust estimator incorporating the function approximators \cite{jiang2016doubly} and marginalized importance sampling using the state-marginal importance ratio $\frac{d^{\pi_\theta}(s)}{d^{\pi_\beta}(s)}$ (where $\theta$ is the model parameter) \cite{sutton2016emphatic}. Similarly, to compute off-policy policy gradient, one can use above importance sampling estimators. Other than reducing the variance using self-normalization, one can apply regularization (such as softmax as in \cite{levine2013guided} or KL-divergence as in \cite{schulman2017proximal}) to constrain the learned policy $\pi_\theta$ to not deviate too far away from the behavioral policy $\pi_\beta$.

There are several strategies to solve offline reinforcement learning problems. One can perform off-policy value function estimation with linear least squared methods \cite{sutton2009fast,lagoudakis2003least} but they can suffer from distributional drift, unless we have some sort of policy constraint that the distribution over the actions $\pi_\theta(a'|s')$ that we compute the target value from should be close to the behavioral distribution $\pi_\beta(a'|s')$. This constraint makes sure that the Q function are only queried on in-distribution actions, such that generalization results should still hold as the errors in Q function are not accumulated. Although in this case, the Q function is evaluated on the same states as the ones in the training set, the action inputs are still flexible enough to be out of distribution. Methods such as 
Advantage Weighted Actor Critic (AWAC, \cite{nair2020accelerating}) uses the KL-divergence between the learned and behavioral policies as an implicit divergence constraints.
If using a policy penalty algorithm, the reward function can be considered to be augmented as $r^*(s,a)= r(s,a) - \alpha f(\pi_\theta, \pi_\beta)$ where $f(\cdot)$ is the divergence function of the polices. Similar to our discussion in bandits, we can also use uncertainty estimation as another constraint, because out-of-distribution actions tend to have large uncertainty and constraining the uncertainty can produce conservative target values, as in \cite{agarwal2020optimistic}. The conservative Q-learning (CQL, \cite{kumar2020conservative}) regularizes the value function or Q function directly to penalize large values and avoid overestimation for out-of-distribution actions. Lastly, model-based offline reinforcement learning (MoREL, \cite{kidambi2020morel}) modifies the MDP model learned from data to induce conservative behavior by penalizing the visitation of states under the model where the model is likely to be incorrect.

For interested readers, \cite{levine2020offline} is a good review on offline reinforcement learning techniques.

\subsection{Transfer learning in reinforcement learning}

Transfer learning is a set of training techniques to train models with greater generalization capabilities by utilizing resources from other domains or resources meant for other purposes \cite{pan2009survey}. The other tasks or purposes that collect the resources from are called source tasks, and the resources from other domains are called the source domains. Our task and domain of interest are called the target task or domain. The goal of transfer learning is to effectively perform a target task on a target domain dataset utilizing the features and weights learned from the source task or dataset. In our context of reinforcement learning, we aim to answer the question that, can reinforcement learning effectively use the prior knowledge to facilitate learning in a new task domain? The intuition is straightforward: if our models have solved prior tasks, they might acquire useful (i.e. reusable) knowledge or insights for solving a new task. The knowledge stored in a reinforcement learning model can come from different components: the Q value function informs us which actions or states are good comparing to the alternatives; the policy function informs us which actions are potentially useful comparing to the alternatives (since some actions in a given state are rarely useful); the world model informs us important understanding of the rules that govern the world (e.g. the laws of physics, which can be generalizable to other physically grounded system); and features, weights and hidden states, which inform us a good representation to use directly, or kick start the training as the initialization for fine-tuning.

Given these potentially useful knowledge in reinforcement learning systems, the transfer learning might help by using the past experience from one set of tasks for faster learning and better performance on a new task. In our context of reinforcement learning, the tasks are usually formulated as Markov decision processes. If we can directly run a policy trained in the source domain in the new domain, we call it a zero-shot transfer learning, where the ``shots'' refer to the number of attempts in the target task or domain. Similarly goes for trying the target task once or a few times, which correspond to one-shot or few-shot transfer learning.

There are two main classes of transfer learning methods, the transductive transfer learning and inductive transfer learning \cite{ruder2019neural}. In transductive transfer learning, the source and target tasks are the same, but labeled data is only available (or much more abundantly available) in the source domain. Here we translate it in reinforcement learning setting: perhaps the reward feedback is not available or only very sparsely revealed in the task in the target domain, but we have access to the historical trajectories or interaction environment of the same task but in the source domain. Domain adaption, as mentioned in one of our earlier examples, is a transductive transfer learning technique. In inductive transfer learning, the source and the target tasks are different and the labelled data is only available (or more abundantly available) in the target domain. Here we translate it in reinforcement learning setting: perhaps we have the historical action-state pairs in a source task, but don't have access to the reward feedback. The inductive transfer learning can be further separated into sequential transfer learning (STL, where multiple tasks are trained sequentially, either in a cross-domain or cross-task fashion), and the multi-task learning (MTL, where multiple tasks are trained at the same time, in a parallel fashion).

More specifically, here we outline a few common strategies. In forward or sequential transfer learning, we simply train the reinforcement learning model on one task, and directly transfer to a new task. One approach is to apply domain adaption in reinforcement learning training. Alternatively, we can also keep the reinforcement learning mechanism intact, but transfer the visual, speech or linguistic representations first. In multi-task transfer learning, we train the reinforcement learning model on many tasks at the same time and transfer to a new task. We can share the neural network representations and layers across tasks in multi-task learning. Alternatively, we can also use contextual policies by baking context-relevant information of the task or domain into the model. Finally, we can transfer the world models and value functions by either using model-based reinforcement learning (which innately serve as an mechanism for transfer), or using temporally correlated features such as successor features and representations.

For interested readers, \cite{taylor2009transfer} is a survey on transfer learning techniques applied in reinforcement learning, and  \cite{da2019survey,silva2016transfer} are two modern surveys on transfer learning in multi-agent reinforcement learning.

\section{Open Questions and Challenges}

\subsection{Multi-agent settings in speech and language}

\begin{figure}[tb]
\centering
    \includegraphics[width=\linewidth]{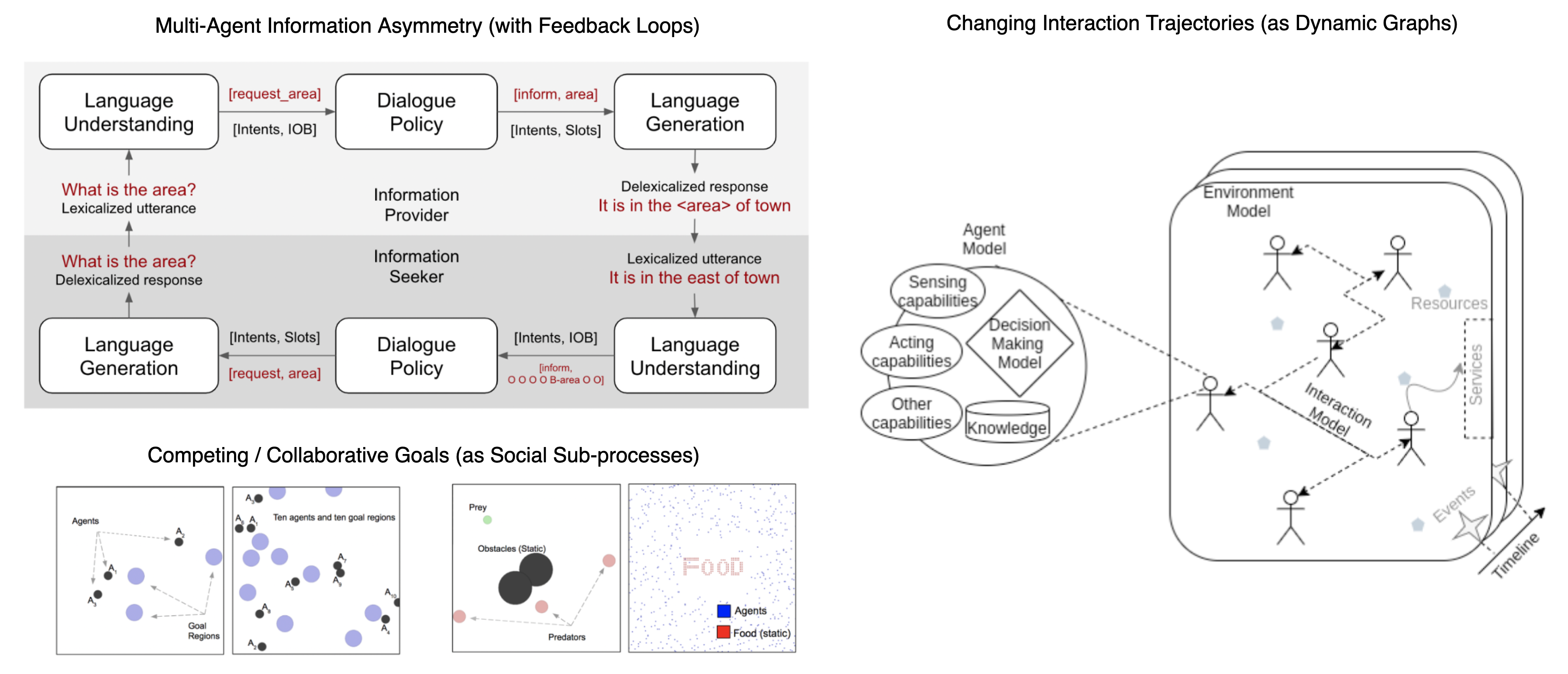}
\vspace{-1em}
\caption{Open question 1: multi-agent settings in speech and language.
}\label{fig:openq1}
\end{figure}

The end goal of the speech and language is to communicate. As we formulate the generative processes of these sequence-based signals as reinforcement learning agents, it is a natural step to simulate and model the mechanistic interactions of these agents in a multi-agent settings. We have shown in an earlier section the example of modeling the dialogue generation task as a single-agent reinforcement learning optimization problem \cite{li2016deep}, but in reality, there is an information asymmetry among the communicating individuals that might be more suitable for a multi-agent system (MAS) solution. The multiagent systems are self-organized systems with multiple interacting agents with computational intelligence \cite{wooldridge2009introduction}. For instance, each communicating individual undergoes an partially isolated information flow via multiple processing units (e.g. language understanding, dialogue policy, and language generation), and have feedback loops in multiple steps which might yield complicated dynamic interactions (Figure \ref{fig:openq1}). 

These interaction trajectories can also change over different time steps, and as a result, modeling the data-generating mechanism of such a system would require modeling a dynamic graphs with changing nodes (as communicating individuals) and edges (as communication types). There can be interesting group dynamics emerge from communications such as leadership, cohesion and conflicts \cite{murray2018nlp}. One can develop solutions to measure these social dynamics, common goals and collective intelligence in population and individual levels. 

From the RL point-of-view, these multi-agent interactions suggest that the actions of one agent can affect the rewards and outcomes of other agents. One challenge in multi-agent settings is that the agents can have conflicting objectives or preferences, which can lead to suboptimal outcomes. For example, in a dialogue system where the system and the user are both agents, the system may prioritize providing accurate recommendations, while the user may prioritize ease of use and naturalness of the conversation. Another challenge is that the agents may have different levels of information and understanding about the environment and each other, which can make it difficult to coordinate their actions and optimize their behaviors. For example, in a multi-speaker speech recognition system, different speakers may have different accents or speech patterns that can affect the performance of the system.

If we only consider the users as our agents to model, the agents in this human-only multi-agent system can be cooperative \cite{panait2005cooperative} or competitive \cite{bansal2017emergent}. Each communicating individuals, if modeled by reinforcement learning systems, might have different observation models and reward perceptions governed by competing or collaborative goals. Modeling (and potentially mimicking) these social sub-processes is a nontrivial task, as in \cite{lin2022online}, which compares the interaction trajectories of the human data and the reinforcement learning algorithms in social dilemma setting. \cite{das2017learning} demonstrates that deep reinforcement learning can interact with natural language in a visual task to share information cooperatively. 

To address these challenges, various approaches have been proposed to perform RL in multi-agent settings in NLP and speech tasks. One approach is to use coordination mechanisms to encourage agents to work together and achieve common goals. Another approach is to use adversarial training, where agents compete against each other in a zero-sum game to improve their performance and achieve optimal outcomes. New speech or language-related concepts can arise from multi-agent modeling. For instance, grounded compositional language can emerge from multi-agent populations of NLP agents, represented as segments of abstract discrete symbols uttered by the communicating agents \cite{mordatch2018emergence}, as further summarized in this review \cite{lin2023survey}. 
A follow up work suggests that these compositional languages has to be constrained in a specific curriculum in multi-agent setting in order to emerge \cite{kottur2017natural}. Lastly, computational techniques of multi-agent systems can also help improve temporal modeling of speech recognition systems, as in \cite{walsh2003multi,nagoev2018model}.

\subsection{Multi-objective training and human priors}
 
 \begin{figure}[tb]
\centering
    \includegraphics[width=\linewidth]{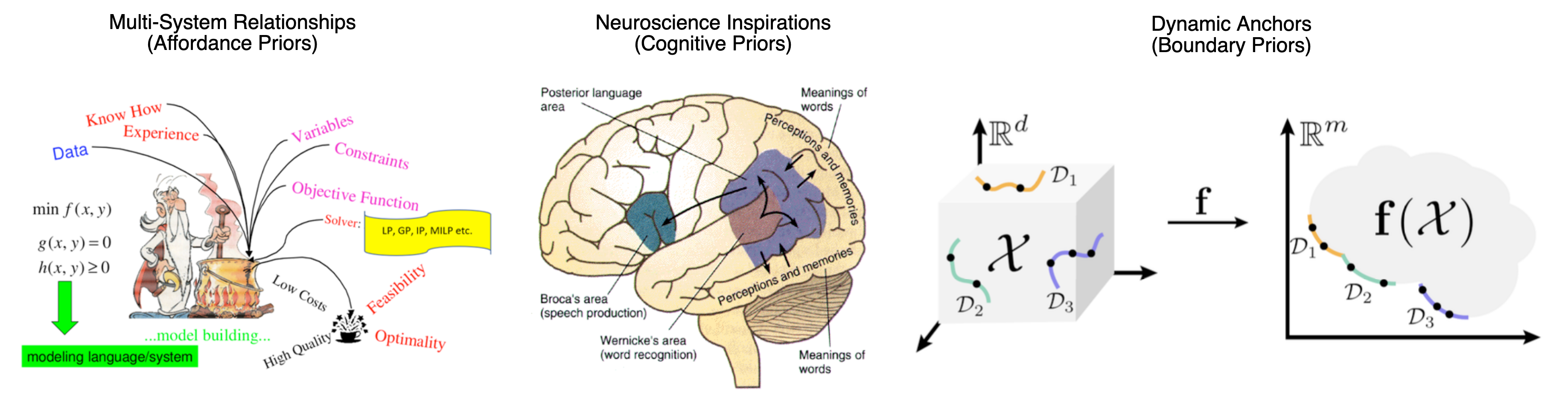}
\vspace{-1em}
\caption{Open question 2: multi-objective training and human priors.
}\label{fig:openq2}
\end{figure}
 
While we have covered many important sub-tasks in speech and language processes, many real-world applications involve the incorporation of multiple processing systems that interact with and depend on one another. For instance, a question answering system might need to understand the intent of the user to craft its objective functions, the descriptive entities as its retrieval constraints, and at the same time, analyze the contexts of the user scenario, reason the relationship of different entities, and learn a planning and prediction model to best interpolate or extrapolate users' past experience to maximize or minimize the estimated objective function given different dimensions of success criteria (Figure \ref{fig:openq2}). A conversational recommendation system, on the other hand, would need to understand the dialogue objective, infer the user's transactional state, estimate the user's dynamic propensities, reason the compatibility and sequential relationship of available inventories, recommend the best items or actions to guide the users towards their transactional destination, all at the same time. The sub-components of such hybrid systems would have different priors on their affordance, i.e. the definitive quality of this processing component about how it can or should be used. And thus, training them together in an end-to-end solution can potentially benefit from specific training curriculum to balance the exploration vs exploitation trade-off of all sub-components. One possible strategy would be to tune the reward function as a multi-objective optimization problem \cite{deb2014multi}, such that the learning objectives for certain sub-components can be emphasized at certain training phases. 


Another strategy would be to learn from human priors. 
Originally proposed as a neuroscience-inspired algorithm, the reinforcement learning solution has been widely studied over the years by neuroscientists and psychologists to further understand the biological constraints and mechanisms of this type of learning, in terms of reward processing \cite{mcclure2004neural}, anatomical separation \cite{lee2012neural}, abnormal states \cite{maia2011reinforcement} and many others. One direction is to consider these biological variants and constraints as a result from natural evolution, i.e. the algorithmic variants that serve a beneficial purpose (with a high fitness) at certain scenarios. Under this assumption, \cite{lin2020astory} studies the reward processing mechanism that distinguishes a series of human psychiatric conditions and introduces it into the reinforcement learning model directly. If given human behavioral data, one can use mechanistic simulation, hierarchical Bayesian inference, or inverse reinforcement learning to build neuromorphic reinforcement learning models that mimic human behaviors \cite{lin2019split}. 
\cite{lin2020unified,lin2021models} further unifies these human behavioral agents at all three levels of bandits, contextual bandits and reinforcement learning. Empirical results suggest that these biologically plausible models are advantageous over existing reinforcement learning models in AI tasks. Other than reward processing, reinforcement learning methods can also model other human priors, such as attention \cite{yang2020predicting}. From the anatomic separation of the biological brains, one can potentially map different speech and language processing components into different brain regions, and pre-program the information flow and training curriculum of these computational components, given the biological and cognitive priors from their corresponding neural correlates.


Applied systems with speech and language processing units can have different states and contexts, which governs the priority of the sub-components in the system. For instance, if the user asks the virtual assistant to play music, the priority of the music recommendation engine should precede that of the conversational engine, as opposed to the case where the user talks to the customer support. As a result, the dynamics and transition of the different contexts would serve as an anchor or boundary in the multi-objective optimization problem. in other words, we can potentially train a universal model that take into account all different contexts. And then, given a certain context, the model would attempt to find the pareto-optimal frontier by projecting affordance onto the context plane. 

Either creating an agent with generalized intelligence across these contexts, or creating specialized agent one context at a time, we need to deal with multi-objective training, the situation where multiple objectives need to be optimized simultaneously, which is a common scenario in NLP and speech tasks. For example, in machine translation, the system needs to generate translations that are both fluent and accurate, which are two separate objectives that need to be balanced. One challenge in multi-objective training is that the objectives can be in conflict with each other, making it difficult to find a single solution that optimizes all of them simultaneously. This can lead to suboptimal performance, where the system fails to fully optimize any of the objectives, or it prioritizes one objective over the other, leading to poor performance on the neglected objective. Another challenge is that the objectives can be difficult to quantify, and can vary depending on the context and task at hand. This can make it difficult to define the reward function that is used to train the RL model, which can affect the quality of the model and the effectiveness of the system. To address these challenges, various approaches have been proposed to perform multi-objective training in NLP and speech tasks using RL. One approach is to use a weighted sum of objectives to combine the multiple objectives into a single objective, which can simplify the optimization process. Another approach is to use a Pareto-based approach, which involves finding a set of solutions that represents the best trade-offs between the multiple objectives.

\section{Summary and Resources}

We starts our survey by motivating the exploration vs exploitation trade-off problem in real-world speech and natural language processing systems with innate data uncertainty and batch-based model training using large-scale relational databases. We formulate the reinforcement learning problem into five sub-classes of methodologies: 
Multi-armed bandits and contextual bandits are both optimization methods that aim to maximize rewards based on actions taken by an agent. While multi-armed bandits select the best action arm to maximize cumulative long-term rewards, contextual bandits use side information or features available to the agent to personalize their action strategies. Reinforcement learning algorithms, on the other hand, are formulated as Markov decision processes and rely on state representations to update policies based on reward feedback received from the environment. Inverse reinforcement learning algorithms learn the reward function of an unknown environment from demonstration trajectories to train reinforcement learning agents, while imitation learning and behavioral cloning use supervised learning to directly learn the mapping from state to action based on historical data of state-action pairs. These optimization methods are essential for improving the performance of agents in various NLP and speech tasks.


In the application domains, we have reviewed speech and language tasks including speech recognition, speaker diarization, spoken language understanding, natural language understanding, sequence generation, text-to-speech synthesis, natural language generation, large language models and conversational recommendation systems, as well as presented case studies regarding the following problems: 
The use of reinforcement learning (RL) has been shown to be effective in improving various aspects of natural language processing (NLP) and speech tasks. In automatic speech recognition (ASR), RL has been applied to improve low-resource training, speech enhancement, batch-wise adaptation, and model training using augmented unlabelled datasets. RL has also been applied to speaker diarization and spoken language understanding (SLU) to enable online interactive label speaker profiles and introduce slot stability by rewarding argmax policy. In NLP, RL has been applied to improve natural language understanding (NLU) in game settings and introduce personalization by adaptively selecting NLU interpretation. RL has also been shown to be useful in determining resource allocation in incremental text-to-speech problems and improving text generation and dialogue generation through inverse RL and combining rewards from two dialogue agents. RL with human or AI feedbacks can help train large language models to align with human values and adopt ethical principles, as well as reach better performance in generating realistic responses. Finally, RL has been applied in conversational recommendation systems to provide recommendations based on conversational dialogues and real-time NLP-parsed elements. These findings demonstrate the potential of RL in improving various NLP and speech tasks, and highlight the importance of combining NLP and RL techniques to achieve more effective and efficient systems.

In the emerging topics, we cover the advances in using deep learning techniques or representations in reinforcement learning and bandit settings, using historical data for off-policy training with offline reinforcement learning, and using transfer learning training techniques to effectively leverage resources or knowledge of other domains or tasks. 

In the open questions, we describe how the speech and language studies can potentially benefit from adopting a multi-agent perspective of a population of communicating individuals as well as several active research directions. We position existing real-world speech and language applications into the challenge of building a multi-system reasoning machine that have interacting sub-components with different training objectives. We propose to use insights from cognitive science or neuroscience as human priors to build better reinforcement learning models that can train faster and more human-like. We describe the possibility of unifying different task contexts as temporal states where pareto-optimal frontier can be located in the multi-objective optimization.


This survey is related to several fast growing fields. Some relevant reviews and textbooks include:

\begin{itemize}
    \item Introduction to reinforcement learning \cite{Sutton1998}
    \item Introduction to multi-armed bandits \cite{slivkins2019introduction}
    \item A survey of inverse reinforcement learning: Challenges, methods and progress \cite{arora2021survey}
    \item Imitation learning: A survey of learning methods \cite{hussein2017imitation} 
    \item Deep reinforcement learning: An overview \cite{li2017deep}
    \item Offline reinforcement learning: Tutorial, review, and perspectives on open problems \cite{levine2020offline}
    \item Survey on applications of multi-armed and contextual bandits \cite{bouneffouf2020survey}
    \item A survey on transfer learning for multiagent reinforcement learning systems \cite{da2019survey}
    \item Transfer learning for multiagent reinforcement learning systems \cite{silva2016transfer}
    \item Transfer learning for reinforcement learning domains: A survey \cite{taylor2009transfer}
    \item Deep reinforcement learning for sequence-to-sequence models \cite{keneshloo2019deep}
    \item Reinforcement learning based recommender systems: A survey \cite{afsar2021reinforcement}
    \item A survey of the usages of deep learning for natural language processing \cite{otter2020survey}
    \item Literature survey of statistical, deep and reinforcement learning in natural language processing \cite{sharma2017literature}
    \item Deep representation learning in speech processing: Challenges, recent advances, and future trends \cite{latif2020deep}
    \item Survey on reinforcement learning for language processing \cite{uc2023survey}
    \item A survey on compositional generalization in applications \cite{lin2023survey}
\end{itemize}

One good way to get onboard the research is to try out code examples. Some useful GitHub repositories include:

\begin{itemize}
    \item Bandit algorithms
  \begin{itemize}
        \item \href{https://github.com/doerlbh/BanditZoo}{https://github.com/doerlbh/BanditZoo} (Python package)
        \item \href{https://github.com/johnmyleswhite/BanditsBook}{https://github.com/johnmyleswhite/BanditsBook} (example)
    \end{itemize}
    \item Reinforcement learning algorithms
  \begin{itemize}
        \item \href{https://github.com/tensorflow/agents}{https://github.com/tensorflow/agents} (Python package)
        \item \href{https://github.com/facebookresearch/ReAgent}{https://github.com/facebookresearch/ReAgent} (Python package)
        \item \href{https://github.com/aikorea/awesome-rl}{https://github.com/aikorea/awesome-rl} (resources)
        \item \href{https://github.com/dennybritz/reinforcement-learning}{https://github.com/dennybritz/reinforcement-learning} (example)
        \item \href{https://github.com/udacity/deep-reinforcement-learning}{https://github.com/udacity/deep-reinforcement-learning} (example)
        \item \href{https://github.com/MorvanZhou/Reinforcement-learning-with-tensorflow}{https://github.com/MorvanZhou/Reinforcement-learning-with-tensorflow} (example)
        \item \href{https://github.com/p-christ/Deep-Reinforcement-Learning-Algorithms-with-PyTorch}{https://github.com/p-christ/Deep-Reinforcement-Learning-Algorithms-with-PyTorch} (example)
    \end{itemize}
    \item Offline Reinforcement learning algorithms
  \begin{itemize}
        \item \href{https://github.com/takuseno/d3rlpy}{https://github.com/takuseno/d3rlpy} (Python package)
        \item \href{https://github.com/hanjuku-kaso/awesome-offline-rl}{https://github.com/hanjuku-kaso/awesome-offline-rl} (resources)
    \end{itemize}
    \item Reinforcement learning applications
  \begin{itemize}
        \item \href{https://github.com/AI4Finance-Foundation/FinRL}{https://github.com/AI4Finance-Foundation/FinRL} (RL + finance)
        \item \href{https://github.com/microsoft/recommenders}{https://github.com/microsoft/recommenders} (recommendation systems)
        \item \href{https://github.com/doerlbh/MiniVox}{https://github.com/doerlbh/MiniVox} (RL + speaker diarization)
        \item \href{https://github.com/doerlbh/awesome-diarization}{https://github.com/doerlbh/awesome-diarization} (speaker diarization)
        \item \href{https://github.com/doerlbh/MentalRL}{https://github.com/doerlbh/MentalRL} (RL + psychiatry behavioral modeling)
        \item \href{https://github.com/doerlbh/DilemmaRL}{https://github.com/doerlbh/DilemmaRL} (RL + multi-agent behavioral modeling)
    \end{itemize}
\end{itemize}

\section{Note and Acknowledgements}

This survey accompanies the tutorial session ``Reinforcement Learning and Bandits for Speech and Language Processing'' held by the author at \textit{INTERSPEECH 2022}. We would like to thank the conference organizers and attended audience for valuable feedback and attention. The materials of the tutorial were inspired by and partially borrows from previous tutorials on reinforcement learning, offline reinforcement learning and recommendation systems by Sergey Levine, David Silver, Xiangyu Zhao, Andrea Barraza-Urbina, Dorota Glowacka, Chelsea Finn, Aviral Kumar, Lilian Weng and Emma Brunskill. The author would like to thank their inspirations to the creation of our tutorial and this survey.

\bibliographystyle{unsrt}
\bibliography{main}  

\begin{thebibliography}{100}

\bibitem{kober2013reinforcement}
Jens Kober, J~Andrew Bagnell, and Jan Peters.
\newblock Reinforcement learning in robotics: A survey.
\newblock {\em The International Journal of Robotics Research},
  32(11):1238--1274, 2013.

\bibitem{le2021deep}
Ngan Le, Vidhiwar~Singh Rathour, Kashu Yamazaki, Khoa Luu, and Marios Savvides.
\newblock Deep reinforcement learning in computer vision: a comprehensive
  survey.
\newblock {\em Artificial Intelligence Review}, pages 1--87, 2021.

\bibitem{fischer2018reinforcement}
Thomas~G Fischer.
\newblock Reinforcement learning in financial markets-a survey.
\newblock Technical report, FAU Discussion Papers in Economics, 2018.

\bibitem{yu2021reinforcement}
Chao Yu, Jiming Liu, Shamim Nemati, and Guosheng Yin.
\newblock Reinforcement learning in healthcare: A survey.
\newblock {\em ACM Computing Surveys (CSUR)}, 55(1):1--36, 2021.

\bibitem{uc2023survey}
Victor Uc-Cetina, Nicolas Navarro-Guerrero, Anabel Martin-Gonzalez, Cornelius
  Weber, and Stefan Wermter.
\newblock Survey on reinforcement learning for language processing.
\newblock {\em Artificial Intelligence Review}, 56(2):1543--1575, 2023.

\bibitem{zhang2019deep}
Zidong Zhang, Dongxia Zhang, and Robert~C Qiu.
\newblock Deep reinforcement learning for power system applications: An
  overview.
\newblock {\em CSEE Journal of Power and Energy Systems}, 6(1):213--225, 2019.

\bibitem{gawlikowski2021survey}
Jakob Gawlikowski, Cedrique Rovile~Njieutcheu Tassi, Mohsin Ali, Jongseok Lee,
  Matthias Humt, Jianxiang Feng, Anna Kruspe, Rudolph Triebel, Peter Jung,
  Ribana Roscher, et~al.
\newblock A survey of uncertainty in deep neural networks.
\newblock {\em arXiv preprint arXiv:2107.03342}, 2021.

\bibitem{Sutton1998}
Richard~S Sutton, Andrew~G Barto, et~al.
\newblock {\em Introduction to reinforcement learning}, volume 135.
\newblock MIT press Cambridge, 1998.

\bibitem{shen2015portfolio}
Weiwei Shen, Jun Wang, Yu-Gang Jiang, and Hongyuan Zha.
\newblock Portfolio choices with orthogonal bandit learning.
\newblock In {\em Twenty-fourth international joint conference on artificial
  intelligence}, 2015.

\bibitem{charpentier2021reinforcement}
Arthur Charpentier, Romuald Elie, and Carl Remlinger.
\newblock Reinforcement learning in economics and finance.
\newblock {\em Computational Economics}, pages 1--38, 2021.

\bibitem{lin2022optimal}
Baihan Lin and Djallel Bouneffouf.
\newblock Optimal epidemic control as a contextual combinatorial bandit with
  budget.
\newblock In {\em 2022 IEEE International Conference on Fuzzy Systems
  (FUZZ-IEEE)}, pages 1--8. IEEE, 2022.

\bibitem{lin2022evolutionary}
Baihan Lin.
\newblock Evolutionary multi-armed bandits with genetic thompson sampling.
\newblock In {\em 2022 IEEE Congress on Evolutionary Computation (CEC)}. IEEE,
  2022.

\bibitem{li2017hyperband}
Lisha Li, Kevin Jamieson, Giulia DeSalvo, Afshin Rostamizadeh, and Ameet
  Talwalkar.
\newblock Hyperband: A novel bandit-based approach to hyperparameter
  optimization.
\newblock {\em The Journal of Machine Learning Research}, 18(1):6765--6816,
  2017.

\bibitem{parker2020provably}
Jack Parker-Holder, Vu~Nguyen, and Stephen~J Roberts.
\newblock Provably efficient online hyperparameter optimization with
  population-based bandits.
\newblock {\em Advances in Neural Information Processing Systems},
  33:17200--17211, 2020.

\bibitem{yang2020exploring}
Liu Yang, Bo~Liu, Leyu Lin, Feng Xia, Kai Chen, and Qiang Yang.
\newblock Exploring clustering of bandits for online recommendation system.
\newblock In {\em Fourteenth ACM Conference on Recommender Systems}, pages
  120--129, 2020.

\bibitem{wang2017biucb}
Lu~Wang, Chengyu Wang, Keqiang Wang, and Xiaofeng He.
\newblock Biucb: A contextual bandit algorithm for cold-start and diversified
  recommendation.
\newblock In {\em 2017 IEEE International Conference on Big Knowledge (ICBK)},
  pages 248--253. IEEE, 2017.

\bibitem{aziz2021multi}
Maryam Aziz, Emilie Kaufmann, and Marie-Karelle Riviere.
\newblock On multi-armed bandit designs for dose-finding clinical trials.
\newblock {\em Journal of Machine Learning Research}, 22(1-38):4, 2021.

\bibitem{villar2015multi}
Sof{\'\i}a~S Villar, Jack Bowden, and James Wason.
\newblock Multi-armed bandit models for the optimal design of clinical trials:
  benefits and challenges.
\newblock {\em Statistical science: a review journal of the Institute of
  Mathematical Statistics}, 30(2):199, 2015.

\bibitem{lin2020unified}
Baihan Lin, Guillermo Cecchi, Djallel Bouneffouf, Jenna Reinen, and Irina Rish.
\newblock {Unified models of human behavioral agents in bandits, contextual
  bandits and RL}.
\newblock {\em arXiv preprint arXiv:2005.04544}, 2020.

\bibitem{lin2021models}
Baihan Lin, Guillermo Cecchi, Djallel Bouneffouf, Jenna Reinen, and Irina Rish.
\newblock Models of human behavioral agents in bandits, contextual bandits and
  rl.
\newblock In {\em Human Brain and Artificial Intelligence: Second International
  Workshop, HBAI 2020, Held in Conjunction with IJCAI-PRICAI 2020, Yokohama,
  Japan, January 7, 2021, Revised Selected Papers 2}, pages 14--33. Springer,
  2021.

\bibitem{bouneffouf2017bandit}
Djallel Bouneffouf, Irina Rish, and Guillermo~A Cecchi.
\newblock Bandit models of human behavior: Reward processing in mental
  disorders.
\newblock In {\em International Conference on Artificial General Intelligence},
  pages 237--248. Springer, 2017.

\bibitem{satyal2018ab}
Suhrid Satyal, Ingo Weber, Hye-young Paik, Claudio~Di Ciccio, and Jan Mendling.
\newblock Ab testing for process versions with contextual multi-armed bandit
  algorithms.
\newblock In {\em International Conference on Advanced Information Systems
  Engineering}, pages 19--34. Springer, 2018.

\bibitem{xiang2022multi}
Ding Xiang, Rebecca West, Jiaqi Wang, Xiquan Cui, and Jinzhou Huang.
\newblock Multi armed bandit vs. a/b tests in e-commence-confidence interval
  and hypothesis test power perspectives.
\newblock In {\em Proceedings of the 28th ACM SIGKDD Conference on Knowledge
  Discovery and Data Mining}, pages 4204--4214, 2022.

\bibitem{kaelbling1996reinforcement}
Leslie~Pack Kaelbling, Michael~L Littman, and Andrew~W Moore.
\newblock Reinforcement learning: A survey.
\newblock {\em Journal of artificial intelligence research}, 4:237--285, 1996.

\bibitem{cesa1998finite}
Nicolo Cesa-Bianchi and Paul Fischer.
\newblock Finite-time regret bounds for the multiarmed bandit problem.
\newblock In {\em ICML}, volume~98, pages 100--108. Citeseer, 1998.

\bibitem{Sutton98}
Richard~S. Sutton and Andrew~G. Barto.
\newblock {\em Introduction to Reinforcement Learning}.
\newblock MIT Press, Cambridge, MA, USA, 1st edition, 1998.

\bibitem{vermorel2005multi}
Joannes Vermorel and Mehryar Mohri.
\newblock Multi-armed bandit algorithms and empirical evaluation.
\newblock In {\em European conference on machine learning}, pages 437--448.
  Springer, 2005.

\bibitem{luce2012individual}
R~Duncan Luce.
\newblock {\em Individual choice behavior: A theoretical analysis}.
\newblock Courier Corporation, 2012.

\bibitem{shanks2002re}
David~R Shanks, Richard~J Tunney, and John~D McCarthy.
\newblock A re-examination of probability matching and rational choice.
\newblock {\em Journal of Behavioral Decision Making}, 15(3):233--250, 2002.

\bibitem{auer2002nonstochastic}
Peter Auer, Nicolo Cesa-Bianchi, Yoav Freund, and Robert~E Schapire.
\newblock The nonstochastic multiarmed bandit problem.
\newblock {\em SIAM Journal on Computing}, 32(1):48--77, 2002.

\bibitem{LaiRobbins1985}
T.~L. Lai and Herbert Robbins.
\newblock Asymptotically efficient adaptive allocation rules.
\newblock {\em Advances in Applied Mathematics}, 6(1):4--22, 1985.

\bibitem{T33}
W.R. Thompson.
\newblock On the likelihood that one unknown probability exceeds another in
  view of the evidence of two samples.
\newblock {\em Biometrika}, 25:285--294, 1933.

\bibitem{chapelle2011empirical}
Olivier Chapelle and Lihong Li.
\newblock An empirical evaluation of thompson sampling.
\newblock In {\em Advances in neural information processing systems}, pages
  2249--2257, 2011.

\bibitem{AgrawalG12}
Shipra Agrawal and Navin Goyal.
\newblock Analysis of thompson sampling for the multi-armed bandit problem.
\newblock In {\em {COLT} 2012 - The 25th Annual Conference on Learning Theory,
  June 25-27, 2012, Edinburgh, Scotland}, pages 39.1--39.26, 2012.

\bibitem{lazaric2014online}
Alessandro Lazaric, Emma Brunskill, et~al.
\newblock Online stochastic optimization under correlated bandit feedback.
\newblock In {\em International Conference on Machine Learning}, pages
  1557--1565. PMLR, 2014.

\bibitem{AuerC98}
Peter Auer and Nicol{\`o} Cesa-Bianchi.
\newblock On-line learning with malicious noise and the closure algorithm.
\newblock {\em Ann. Math. Artif. Intell.}, 23(1-2):83--99, 1998.

\bibitem{AuerCFS02}
Peter Auer, Nicol{\`o} Cesa-Bianchi, Yoav Freund, and Robert~E. Schapire.
\newblock The nonstochastic multiarmed bandit problem.
\newblock {\em SIAM J. Comput.}, 32(1):48--77, 2002.

\bibitem{BouneffoufF16}
Djallel Bouneffouf and Rapha{\"{e}}l F{\'{e}}raud.
\newblock Multi-armed bandit problem with known trend.
\newblock {\em Neurocomputing}, 205:16--21, 2016.

\bibitem{garivier2008upper}
Aur{\'e}lien Garivier and Eric Moulines.
\newblock On upper-confidence bound policies for non-stationary bandit
  problems.
\newblock {\em arXiv preprint arXiv:0805.3415}, 2008.

\bibitem{lin2018contextual}
Baihan Lin, Djallel Bouneffouf, Guillermo~A Cecchi, and Irina Rish.
\newblock Contextual bandit with adaptive feature extraction.
\newblock In {\em 2018 IEEE International Conference on Data Mining Workshops
  (ICDMW)}, pages 937--944. IEEE, 2018.

\bibitem{srinivas2009gaussian}
Niranjan Srinivas, Andreas Krause, Sham~M Kakade, and Matthias Seeger.
\newblock Gaussian process optimization in the bandit setting: No regret and
  experimental design.
\newblock {\em arXiv preprint arXiv:0912.3995}, 2009.

\bibitem{trovo2016budgeted}
Francesco Trov{\`o}, Stefano Paladino, Marcello Restelli, and Nicola Gatti.
\newblock Budgeted multi--armed bandit in continuous action space.
\newblock In {\em Proceedings of the Twenty-second European Conference on
  Artificial Intelligence}, pages 560--568, 2016.

\bibitem{wang2008algorithms}
Yizao Wang, Jean-Yves Audibert, and R{\'e}mi Munos.
\newblock Algorithms for infinitely many-armed bandits.
\newblock {\em Advances in Neural Information Processing Systems}, 21, 2008.

\bibitem{chen2013combinatorial}
Wei Chen, Yajun Wang, and Yang Yuan.
\newblock Combinatorial multi-armed bandit: General framework and applications.
\newblock In {\em International conference on machine learning}, pages
  151--159. PMLR, 2013.

\bibitem{lin2021optimal}
Baihan Lin and Djallel Bouneffouf.
\newblock Optimal epidemic control as a contextual combinatorial bandit with
  budget.
\newblock {\em arXiv preprint arXiv:2106.15808}, 2021.

\bibitem{lattimore2016regret}
Tor Lattimore.
\newblock Regret analysis of the finite-horizon gittins index strategy for
  multi-armed bandits.
\newblock In {\em Conference on Learning Theory}, pages 1214--1245. PMLR, 2016.

\bibitem{kocak2014efficient}
Tom{\'a}{\v{s}} Koc{\'a}k, Gergely Neu, Michal Valko, and R{\'e}mi Munos.
\newblock Efficient learning by implicit exploration in bandit problems with
  side observations.
\newblock {\em Advances in Neural Information Processing Systems}, 27, 2014.

\bibitem{berlinucb}
Baihan Lin.
\newblock Online semi-supervised learning in contextual bandits with episodic
  reward.
\newblock In {\em Australasian Joint Conference on Artificial Intelligence},
  pages 407--419. Springer, 2020.

\bibitem{ding2013multi}
Wenkui Ding, Tao Qin, Xu-Dong Zhang, and Tie-Yan Liu.
\newblock Multi-armed bandit with budget constraint and variable costs.
\newblock In {\em Twenty-Seventh AAAI Conference on Artificial Intelligence},
  2013.

\bibitem{badanidiyuru2018bandits}
Ashwinkumar Badanidiyuru, Robert Kleinberg, and Aleksandrs Slivkins.
\newblock Bandits with knapsacks.
\newblock {\em Journal of the ACM (JACM)}, 65(3):1--55, 2018.

\bibitem{slivkins2019introduction}
Aleksandrs Slivkins et~al.
\newblock Introduction to multi-armed bandits.
\newblock {\em Foundations and Trends{\textregistered} in Machine Learning},
  12(1-2):1--286, 2019.

\bibitem{langford2008epoch}
John Langford and Tong Zhang.
\newblock Epoch-greedy algorithm for multi-armed bandits with side information.
\newblock {\em Advances in Neural Information Processing Systems (NIPS 2007)},
  20:1, 2007.

\bibitem{ChuLRS11}
Wei Chu, Lihong Li, Lev Reyzin, and Robert~E. Schapire.
\newblock Contextual bandits with linear payoff functions.
\newblock In Geoffrey~J. Gordon, David~B. Dunson, and Miroslav Dudik, editors,
  {\em AISTATS}, volume~15 of {\em JMLR Proceedings}, pages 208--214. JMLR.org,
  2011.

\bibitem{AgrawalG13}
Shipra Agrawal and Navin Goyal.
\newblock Thompson sampling for contextual bandits with linear payoffs.
\newblock In {\em ICML (3)}, pages 127--135, 2013.

\bibitem{zhou2015survey}
Li~Zhou.
\newblock A survey on contextual multi-armed bandits.
\newblock {\em arXiv preprint arXiv:1508.03326}, 2015.

\bibitem{bertsekas1996neuro}
D.P. Bertsekas and J.N. Tsitsiklis.
\newblock {\em Neuro-dynamic programming}.
\newblock Athena Scientific, 1996.

\bibitem{weng2018PG}
Lilian Weng.
\newblock Policy gradient algorithms.
\newblock {\em lilianweng.github.io}, 2018.

\bibitem{schulman2015high}
John Schulman, Philipp Moritz, Sergey Levine, Michael Jordan, and Pieter
  Abbeel.
\newblock High-dimensional continuous control using generalized advantage
  estimation.
\newblock {\em arXiv preprint arXiv:1506.02438}, 2015.

\bibitem{sutton1999policy}
Richard~S Sutton, David McAllester, Satinder Singh, and Yishay Mansour.
\newblock Policy gradient methods for reinforcement learning with function
  approximation.
\newblock {\em Advances in neural information processing systems}, 12, 1999.

\bibitem{konda1999actor}
Vijay Konda and John Tsitsiklis.
\newblock Actor-critic algorithms.
\newblock {\em Advances in neural information processing systems}, 12, 1999.

\bibitem{russell1998learning}
Stuart Russell.
\newblock Learning agents for uncertain environments.
\newblock In {\em Proceedings of the eleventh annual conference on
  Computational learning theory}, pages 101--103, 1998.

\bibitem{ng2000algorithms}
Andrew~Y Ng, Stuart~J Russell, et~al.
\newblock Algorithms for inverse reinforcement learning.
\newblock In {\em Icml}, volume~1, page~2, 2000.

\bibitem{fu2020d4rl}
Justin Fu, Aviral Kumar, Ofir Nachum, George Tucker, and Sergey Levine.
\newblock D4rl: Datasets for deep data-driven reinforcement learning.
\newblock {\em arXiv preprint arXiv:2004.07219}, 2020.

\bibitem{armstrong2018occam}
Stuart Armstrong and S{\"o}ren Mindermann.
\newblock Occam's razor is insufficient to infer the preferences of irrational
  agents.
\newblock {\em Advances in neural information processing systems}, 31, 2018.

\bibitem{abbeel2004apprenticeship}
Pieter Abbeel and Andrew~Y Ng.
\newblock Apprenticeship learning via inverse reinforcement learning.
\newblock In {\em Proceedings of the twenty-first international conference on
  Machine learning}, page~1. ACM, 2004.

\bibitem{arora2021survey}
Saurabh Arora and Prashant Doshi.
\newblock A survey of inverse reinforcement learning: Challenges, methods and
  progress.
\newblock {\em Artificial Intelligence}, 297:103500, 2021.

\bibitem{ratliff2006maximum}
Nathan~D Ratliff, J~Andrew Bagnell, and Martin~A Zinkevich.
\newblock Maximum margin planning.
\newblock In {\em Proceedings of the 23rd international conference on Machine
  learning}, pages 729--736, 2006.

\bibitem{ziebart2008maximum}
Brian~D Ziebart, Andrew~L Maas, J~Andrew Bagnell, Anind~K Dey, et~al.
\newblock Maximum entropy inverse reinforcement learning.
\newblock In {\em Aaai}, volume~8, pages 1433--1438. Chicago, IL, USA, 2008.

\bibitem{goodfellow2020generative}
Ian Goodfellow, Jean Pouget-Abadie, Mehdi Mirza, Bing Xu, David Warde-Farley,
  Sherjil Ozair, Aaron Courville, and Yoshua Bengio.
\newblock Generative adversarial networks.
\newblock {\em Communications of the ACM}, 63(11):139--144, 2020.

\bibitem{finn2016connection}
Chelsea Finn, Paul Christiano, Pieter Abbeel, and Sergey Levine.
\newblock A connection between generative adversarial networks, inverse
  reinforcement learning, and energy-based models.
\newblock {\em arXiv preprint arXiv:1611.03852}, 2016.

\bibitem{ho2016generative}
Jonathan Ho and Stefano Ermon.
\newblock Generative adversarial imitation learning.
\newblock {\em Advances in neural information processing systems}, 29, 2016.

\bibitem{finn2016guided}
Chelsea Finn, Sergey Levine, and Pieter Abbeel.
\newblock Guided cost learning: Deep inverse optimal control via policy
  optimization.
\newblock In {\em International conference on machine learning}, pages 49--58.
  PMLR, 2016.

\bibitem{schulman2015trust}
John Schulman, Sergey Levine, Pieter Abbeel, Michael Jordan, and Philipp
  Moritz.
\newblock Trust region policy optimization.
\newblock In {\em International conference on machine learning}, pages
  1889--1897. PMLR, 2015.

\bibitem{hussein2017imitation}
Ahmed Hussein, Mohamed~Medhat Gaber, Eyad Elyan, and Chrisina Jayne.
\newblock Imitation learning: A survey of learning methods.
\newblock {\em ACM Computing Surveys (CSUR)}, 50(2):1--35, 2017.

\bibitem{osa2018algorithmic}
Takayuki Osa, Joni Pajarinen, Gerhard Neumann, J~Andrew Bagnell, Pieter Abbeel,
  Jan Peters, et~al.
\newblock An algorithmic perspective on imitation learning.
\newblock {\em Foundations and Trends{\textregistered} in Robotics},
  7(1-2):1--179, 2018.

\bibitem{ross2011reduction}
St{\'e}phane Ross, Geoffrey Gordon, and Drew Bagnell.
\newblock A reduction of imitation learning and structured prediction to
  no-regret online learning.
\newblock In {\em Proceedings of the fourteenth international conference on
  artificial intelligence and statistics}, pages 627--635. JMLR Workshop and
  Conference Proceedings, 2011.

\bibitem{lin2022ipd}
Baihan Lin, Djallel Bouneffouf, and Guillermo Cecchi.
\newblock Online learning in iterated prisoner's dilemma to mimic human
  behavior.
\newblock In {\em Pacific Rim International Conference on Artificial
  Intelligence}. Springer, 2022.

\bibitem{kumar2021should}
Aviral Kumar, Joey Hong, Anikait Singh, and Sergey Levine.
\newblock Should i run offline reinforcement learning or behavioral cloning?
\newblock In {\em International Conference on Learning Representations}, 2021.

\bibitem{shen2019reinforcement}
Yih-Liang Shen, Chao-Yuan Huang, Syu-Siang Wang, Yu~Tsao, Hsin-Min Wang, and
  Tai-Shih Chi.
\newblock Reinforcement learning based speech enhancement for robust speech
  recognition.
\newblock In {\em ICASSP 2019-2019 IEEE International Conference on Acoustics,
  Speech and Signal Processing (ICASSP)}, pages 6750--6754. IEEE, 2019.

\bibitem{kala2018reinforcement}
Taku Kala and Takahiro Shinozaki.
\newblock Reinforcement learning of speech recognition system based on policy
  gradient and hypothesis selection.
\newblock In {\em 2018 IEEE International Conference on Acoustics, Speech and
  Signal Processing (ICASSP)}, pages 5759--5763. IEEE, 2018.

\bibitem{chung2020semi}
Hoon Chung, Hyeong-Bae Jeon, and Jeon~Gue Park.
\newblock Semi-supervised training for sequence-to-sequence speech recognition
  using reinforcement learning.
\newblock In {\em 2020 International Joint Conference on Neural Networks
  (IJCNN)}, pages 1--6. IEEE, 2020.

\bibitem{rajapakshe2019pre}
Thejan Rajapakshe, Rajib Rana, Siddique Latif, Sara Khalifa, and Bj{\"o}rn~W
  Schuller.
\newblock Pre-training in deep reinforcement learning for automatic speech
  recognition.
\newblock {\em arXiv preprint arXiv:1910.11256}, 2019.

\bibitem{kuznetsova2021bandit}
Anastasia Kuznetsova, Anurag Kumar, and Francis~M Tyers.
\newblock A bandit approach to curriculum generation for automatic speech
  recognition.
\newblock {\em arXiv preprint arXiv:2102.03662}, 2021.

\bibitem{tjandra2018sequence}
Andros Tjandra, Sakriani Sakti, and Satoshi Nakamura.
\newblock Sequence-to-sequence asr optimization via reinforcement learning.
\newblock In {\em 2018 IEEE International Conference on Acoustics, Speech and
  Signal Processing (ICASSP)}, pages 5829--5833. IEEE, 2018.

\bibitem{tjandra2019end}
Andros Tjandra, Sakriani Sakti, and Satoshi Nakamura.
\newblock End-to-end speech recognition sequence training with reinforcement
  learning.
\newblock {\em IEEE Access}, 7:79758--79769, 2019.

\bibitem{karita2018sequence}
Shigeki Karita, Atsunori Ogawa, Marc Delcroix, and Tomohiro Nakatani.
\newblock Sequence training of encoder-decoder model using policy gradient for
  end-to-end speech recognition.
\newblock In {\em 2018 IEEE International Conference on Acoustics, Speech and
  Signal Processing (ICASSP)}, pages 5839--5843. IEEE, 2018.

\bibitem{lin2020voiceid}
Baihan Lin and Xinxin Zhang.
\newblock Voiceid on the fly: A speaker recognition system that learns from
  scratch.
\newblock In {\em INTERSPEECH}, 2020.

\bibitem{lin2020speaker}
Baihan Lin and Xinxin Zhang.
\newblock Speaker diarization as a fully online learning problem in minivox.
\newblock {\em arXiv preprint arXiv:2006.04376}, 2020.

\bibitem{lin2021speaker}
Baihan Lin and Xinxin Zhang.
\newblock Speaker diarization as a fully online bandit learning problem in
  minivox.
\newblock In {\em Asian Conference on Machine Learning}, pages 1660--1674.
  PMLR, 2021.

\bibitem{lin2022voice}
Baihan Lin.
\newblock {Voice2Alliance: automatic speaker diarization and quality assurance
  of conversational alignment}.
\newblock In {\em INTERSPEECH}, 2022.

\bibitem{lin2022rldiarization}
Baihan Lin and Xinxin Zhang.
\newblock A reinforcement learning framework for online speaker diarization.
\newblock {\em arXiv preprint arXiv:2302.10924}, 2023.

\bibitem{bai2018source}
He~Bai, Yu~Zhou, Jiajun Zhang, Liang Zhao, Mei-Yuh Hwang, and Chengqing Zong.
\newblock Source-critical reinforcement learning for transferring spoken
  language understanding to a new language.
\newblock {\em arXiv preprint arXiv:1808.06167}, 2018.

\bibitem{bellegarda2014spoken}
Jerome~R Bellegarda.
\newblock Spoken language understanding for natural interaction: The siri
  experience.
\newblock {\em Natural interaction with robots, knowbots and smartphones},
  pages 3--14, 2014.

\bibitem{ferreira2016adversarial}
Emmanuel Ferreira, Alexandre~Reiffers Masson, Bassam Jabaian, and Fabrice
  Lef{\`e}vre.
\newblock Adversarial bandit for online interactive active learning of
  zero-shot spoken language understanding.
\newblock In {\em 2016 IEEE International Conference on Acoustics, Speech and
  Signal Processing (ICASSP)}, pages 6155--6159. IEEE, 2016.

\bibitem{lin2022ispeak}
Baihan Lin and Xinxin Zhang.
\newblock ispeak: Interactive spoken language understanding system for children
  with speech and language disorders.
\newblock In {\em 2022 IEEE Spoken Language Technology Workshop (SLT)}. IEEE,
  2022.

\bibitem{lin2022ispeak_icassp}
Baihan Lin and Xinxin Zhang.
\newblock Interactive spoken language understanding system for children with
  speech and language disorders.
\newblock {\em arXiv preprint}, 2023.

\bibitem{chen2020deep}
Zhi Chen, Lu~Chen, Xiang Zhou, and Kai Yu.
\newblock Deep reinforcement learning for on-line dialogue state tracking.
\newblock {\em arXiv preprint arXiv:2009.10321}, 2020.

\bibitem{narasimhan2015language}
Karthik Narasimhan, Tejas Kulkarni, and Regina Barzilay.
\newblock Language understanding for text-based games using deep reinforcement
  learning.
\newblock {\em arXiv preprint arXiv:1506.08941}, 2015.

\bibitem{moerchen2020personalizing}
Fabian Moerchen, Patrick Ernst, and Giovanni Zappella.
\newblock Personalizing natural language understanding using multi-armed
  bandits and implicit feedback.
\newblock In {\em Proceedings of the 29th ACM International Conference on
  Information \& Knowledge Management}, pages 2661--2668, 2020.

\bibitem{luketina2019survey}
Jelena Luketina, Nantas Nardelli, Gregory Farquhar, Jakob Foerster, Jacob
  Andreas, Edward Grefenstette, Shimon Whiteson, and Tim Rockt{\"a}schel.
\newblock A survey of reinforcement learning informed by natural language.
\newblock {\em arXiv preprint arXiv:1906.03926}, 2019.

\bibitem{vogel2010learning}
Adam Vogel and Dan Jurafsky.
\newblock Learning to follow navigational directions.
\newblock In {\em Proceedings of the 48th annual meeting of the association for
  computational linguistics}, pages 806--814, 2010.

\bibitem{keneshloo2019deep}
Yaser Keneshloo, Tian Shi, Naren Ramakrishnan, and Chandan~K Reddy.
\newblock Deep reinforcement learning for sequence-to-sequence models.
\newblock {\em IEEE transactions on neural networks and learning systems},
  31(7):2469--2489, 2019.

\bibitem{mohan2020incremental}
Devang S~Ram Mohan, Raphael Lenain, Lorenzo Foglianti, Tian~Huey Teh, Marlene
  Staib, Alexandra Torresquintero, and Jiameng Gao.
\newblock Incremental text to speech for neural sequence-to-sequence models
  using reinforcement learning.
\newblock {\em arXiv preprint arXiv:2008.03096}, 2020.

\bibitem{liu2021reinforcement}
Rui Liu, Berrak Sisman, and Haizhou Li.
\newblock Reinforcement learning for emotional text-to-speech synthesis with
  improved emotion discriminability.
\newblock {\em arXiv preprint arXiv:2104.01408}, 2021.

\bibitem{gibson2022reinforcement}
Jerry Gibson and Hoontaek Oh.
\newblock A reinforcement learning approach to speech coding.
\newblock {\em Information}, 13(7):331, 2022.

\bibitem{jiang2020rl}
Nan Jiang, Sheng Jin, Zhiyao Duan, and Changshui Zhang.
\newblock Rl-duet: Online music accompaniment generation using deep
  reinforcement learning.
\newblock In {\em Proceedings of the AAAI Conference on Artificial
  Intelligence}, volume~34, pages 710--718, 2020.

\bibitem{shi2018toward}
Zhan Shi, Xinchi Chen, Xipeng Qiu, and Xuanjing Huang.
\newblock Toward diverse text generation with inverse reinforcement learning.
\newblock {\em arXiv preprint arXiv:1804.11258}, 2018.

\bibitem{li2016deep}
Jiwei Li, Will Monroe, Alan Ritter, Michel Galley, Jianfeng Gao, and Dan
  Jurafsky.
\newblock Deep reinforcement learning for dialogue generation.
\newblock {\em arXiv preprint arXiv:1606.01541}, 2016.

\bibitem{yu2017seqgan}
Lantao Yu, Weinan Zhang, Jun Wang, and Yong Yu.
\newblock Seqgan: Sequence generative adversarial nets with policy gradient.
\newblock In {\em Proceedings of the AAAI conference on artificial
  intelligence}, volume~31, 2017.

\bibitem{lin2019split}
Baihan Lin, Djallel Bouneffouf, and Guillermo Cecchi.
\newblock {Split Q Learning: Reinforcement Learning with Two-Stream Rewards}.
\newblock In {\em Proceedings of the Twenty-Eighth International Joint
  Conference on Artificial Intelligence, {IJCAI-19}}, pages 6448--6449. AAAI
  Press, International Joint Conferences on Artificial Intelligence
  Organization, 7 2019.

\bibitem{li2017paraphrase}
Zichao Li, Xin Jiang, Lifeng Shang, and Hang Li.
\newblock Paraphrase generation with deep reinforcement learning.
\newblock {\em arXiv preprint arXiv:1711.00279}, 2017.

\bibitem{dethlefs2011combining}
Nina Dethlefs and Heriberto Cuay{\'a}huitl.
\newblock Combining hierarchical reinforcement learning and bayesian networks
  for natural language generation in situated dialogue.
\newblock In {\em Proceedings of the 13th European Workshop on Natural Language
  Generation}, pages 110--120, 2011.

\bibitem{dethlefs2011hierarchical}
Nina Dethlefs and Heriberto Cuay{\'a}huitl.
\newblock Hierarchical reinforcement learning and hidden markov models for
  task-oriented natural language generation.
\newblock In {\em Proceedings of the 49th Annual Meeting of the Association for
  Computational Linguistics: Human Language Technologies}, pages 654--659,
  2011.

\bibitem{riou2019reinforcement}
Matthieu Riou, Bassam Jabaian, St{\'e}phane Huet, and Fabrice Lef{\`e}vre.
\newblock Reinforcement adaptation of an attention-based neural natural
  language generator for spoken dialogue systems.
\newblock {\em Dialogue \& Discourse}, 10:1--19, 2019.

\bibitem{riou2017online}
Matthieu Riou, Bassam Jabaian, St{\'e}phane Huet, and Fabrice Lef{\`e}vre.
\newblock Online adaptation of an attention-based neural network for natural
  language generation.
\newblock In {\em Conference of the International Speech Communication
  Association (Interspeech)}, 2017.

\bibitem{zhong2017seq2sql}
Victor Zhong, Caiming Xiong, and Richard Socher.
\newblock Seq2sql: Generating structured queries from natural language using
  reinforcement learning.
\newblock {\em arXiv preprint arXiv:1709.00103}, 2017.

\bibitem{ouyang2022training}
Long Ouyang, Jeff Wu, Xu~Jiang, Diogo Almeida, Carroll~L Wainwright, Pamela
  Mishkin, Chong Zhang, Sandhini Agarwal, Katarina Slama, Alex Ray, et~al.
\newblock Training language models to follow instructions with human feedback.
\newblock {\em arXiv preprint arXiv:2203.02155}, 2022.

\bibitem{bai2022constitutional}
Yuntao Bai, Saurav Kadavath, Sandipan Kundu, Amanda Askell, Jackson Kernion,
  Andy Jones, Anna Chen, Anna Goldie, Azalia Mirhoseini, Cameron McKinnon,
  et~al.
\newblock Constitutional ai: Harmlessness from ai feedback.
\newblock {\em arXiv preprint arXiv:2212.08073}, 2022.

\bibitem{lin2023towards}
Baihan Lin, Djallel Bouneffouf, Guillermo Cecchi, and Kush~R Varshney.
\newblock Towards healthy {AI}: Large language models need therapists too.
\newblock {\em arXiv preprint arXiv:2304.00416}, 2023.

\bibitem{brown2020language}
Tom Brown, Benjamin Mann, Nick Ryder, Melanie Subbiah, Jared~D Kaplan, Prafulla
  Dhariwal, Arvind Neelakantan, Pranav Shyam, Girish Sastry, Amanda Askell,
  et~al.
\newblock Language models are few-shot learners.
\newblock {\em Advances in neural information processing systems},
  33:1877--1901, 2020.

\bibitem{chowdhery2022palm}
Aakanksha Chowdhery, Sharan Narang, Jacob Devlin, Maarten Bosma, Gaurav Mishra,
  Adam Roberts, Paul Barham, Hyung~Won Chung, Charles Sutton, Sebastian
  Gehrmann, et~al.
\newblock Palm: Scaling language modeling with pathways.
\newblock {\em arXiv preprint arXiv:2204.02311}, 2022.

\bibitem{christiano2017deep}
Paul~F Christiano, Jan Leike, Tom Brown, Miljan Martic, Shane Legg, and Dario
  Amodei.
\newblock Deep reinforcement learning from human preferences.
\newblock {\em Advances in neural information processing systems}, 30, 2017.

\bibitem{stiennon2020learning}
Nisan Stiennon, Long Ouyang, Jeffrey Wu, Daniel Ziegler, Ryan Lowe, Chelsea
  Voss, Alec Radford, Dario Amodei, and Paul~F Christiano.
\newblock Learning to summarize with human feedback.
\newblock {\em Advances in Neural Information Processing Systems},
  33:3008--3021, 2020.

\bibitem{perez2022red}
Ethan Perez, Saffron Huang, Francis Song, Trevor Cai, Roman Ring, John
  Aslanides, Amelia Glaese, Nat McAleese, and Geoffrey Irving.
\newblock Red teaming language models with language models.
\newblock {\em arXiv preprint arXiv:2202.03286}, 2022.

\bibitem{gao2021advances}
Chongming Gao, Wenqiang Lei, Xiangnan He, Maarten de~Rijke, and Tat-Seng Chua.
\newblock Advances and challenges in conversational recommender systems: A
  survey.
\newblock {\em AI Open}, 2:100--126, 2021.

\bibitem{li2021seamlessly}
Shijun Li, Wenqiang Lei, Qingyun Wu, Xiangnan He, Peng Jiang, and Tat-Seng
  Chua.
\newblock Seamlessly unifying attributes and items: Conversational
  recommendation for cold-start users.
\newblock {\em ACM Transactions on Information Systems (TOIS)}, 39(4):1--29,
  2021.

\bibitem{lin2022supervisor}
Baihan Lin, Guillermo Cecchi, and Djallel Bouneffouf.
\newblock {SupervisorBot: {NLP}-Annotated Real-Time Recommendations of
  Psychotherapy Treatment Strategies with Deep Reinforcement Learning}.
\newblock In {\em Proceedings of the Thirty-Second International Joint
  Conference on Artificial Intelligence, {IJCAI-23}}. International Joint
  Conferences on Artificial Intelligence Organization, 8 2023.

\bibitem{christakopoulou2016towards}
Konstantina Christakopoulou, Filip Radlinski, and Katja Hofmann.
\newblock Towards conversational recommender systems.
\newblock In {\em Proceedings of the 22nd ACM SIGKDD international conference
  on knowledge discovery and data mining}, pages 815--824, 2016.

\bibitem{zhang2020conversational}
Xiaoying Zhang, Hong Xie, Hang Li, and John CS~Lui.
\newblock Conversational contextual bandit: Algorithm and application.
\newblock In {\em Proceedings of the web conference 2020}, pages 662--672,
  2020.

\bibitem{lin2023help}
Baihan Lin, Guillermo Cecchi, and Djallel Bouneffouf.
\newblock Helping therapists with {NLP}-annotated recommendation.
\newblock In {\em Joint Proceedings of the ACM IUI Workshops}, 2023.

\bibitem{lin2023psychotherapy}
Baihan Lin, Guillermo Cecchi, and Djallel Bouneffouf.
\newblock Psychotherapy {AI} companion with reinforcement learning
  recommendations and interpretable policy dynamics.
\newblock In {\em Proceedings of the Web Conference 2023}, 2023.

\bibitem{lin2022deep}
Baihan Lin, Guillermo Cecchi, and Djallel Bouneffouf.
\newblock Deep annotation of therapeutic working alliance in psychotherapy.
\newblock In {\em International Workshop on Health Intelligence}. Springer,
  2023.

\bibitem{lin2022deep2}
Baihan Lin, Guillermo Cecchi, and Djallel Bouneffouf.
\newblock Working alliance transformer for psychotherapy dialogue
  classification.
\newblock {\em arXiv preprint arXiv:2210.15603}, 2022.

\bibitem{lin2022knowledge}
Baihan Lin.
\newblock Knowledge management system with {NLP}-assisted annotations: A brief
  survey and outlook.
\newblock In {\em CIKM Workshops}, 2022.

\bibitem{lin2022neural}
Baihan Lin, Djallel Bouneffouf, Guillermo Cecchi, and Ravi Tejwani.
\newblock Neural topic modeling of psychotherapy sessions.
\newblock In {\em International Workshop on Health Intelligence}. Springer,
  2023.

\bibitem{lin2022computational}
Baihan Lin.
\newblock Computational inference in cognitive science: Operational, societal
  and ethical considerations.
\newblock {\em arXiv preprint arXiv:2210.13526}, 2022.

\bibitem{he2016deep}
Ji~He, Mari Ostendorf, Xiaodong He, Jianshu Chen, Jianfeng Gao, Lihong Li, and
  Li~Deng.
\newblock Deep reinforcement learning with a combinatorial action space for
  predicting popular reddit threads.
\newblock {\em arXiv preprint arXiv:1606.03667}, 2016.

\bibitem{he2017reinforcement}
Ji~He, Mari Ostendorf, and Xiaodong He.
\newblock Reinforcement learning with external knowledge and two-stage
  q-functions for predicting popular reddit threads.
\newblock {\em arXiv preprint arXiv:1704.06217}, 2017.

\bibitem{mnih2013playing}
Volodymyr Mnih, Koray Kavukcuoglu, David Silver, Alex Graves, Ioannis
  Antonoglou, Daan Wierstra, and Martin Riedmiller.
\newblock Playing atari with deep reinforcement learning.
\newblock {\em arXiv preprint arXiv:1312.5602}, 2013.

\bibitem{silver2014deterministic}
David Silver, Guy Lever, Nicolas Heess, Thomas Degris, Daan Wierstra, and
  Martin Riedmiller.
\newblock Deterministic policy gradient algorithms.
\newblock In {\em International conference on machine learning}, pages
  387--395. PMLR, 2014.

\bibitem{lillicrap2015continuous}
Timothy~P Lillicrap, Jonathan~J Hunt, Alexander Pritzel, Nicolas Heess, Tom
  Erez, Yuval Tassa, David Silver, and Daan Wierstra.
\newblock Continuous control with deep reinforcement learning.
\newblock {\em arXiv preprint arXiv:1509.02971}, 2015.

\bibitem{schulman2017proximal}
John Schulman, Filip Wolski, Prafulla Dhariwal, Alec Radford, and Oleg Klimov.
\newblock Proximal policy optimization algorithms.
\newblock {\em arXiv preprint arXiv:1707.06347}, 2017.

\bibitem{collier2018deep}
Mark Collier and Hector~Urdiales Llorens.
\newblock Deep contextual multi-armed bandits.
\newblock {\em arXiv preprint arXiv:1807.09809}, 2018.

\bibitem{guo2020deep}
Dalin Guo, Sofia~Ira Ktena, Pranay~Kumar Myana, Ferenc Huszar, Wenzhe Shi,
  Alykhan Tejani, Michael Kneier, and Sourav Das.
\newblock Deep bayesian bandits: Exploring in online personalized
  recommendations.
\newblock In {\em Fourteenth ACM Conference on Recommender Systems}, pages
  456--461, 2020.

\bibitem{zhou2020neural}
Dongruo Zhou, Lihong Li, and Quanquan Gu.
\newblock Neural contextual bandits with ucb-based exploration.
\newblock In {\em International Conference on Machine Learning}, pages
  11492--11502. PMLR, 2020.

\bibitem{li2017deep}
Yuxi Li.
\newblock Deep reinforcement learning: An overview.
\newblock {\em arXiv preprint arXiv:1701.07274}, 2017.

\bibitem{levine2020offline}
Sergey Levine, Aviral Kumar, George Tucker, and Justin Fu.
\newblock Offline reinforcement learning: Tutorial, review, and perspectives on
  open problems.
\newblock {\em arXiv preprint arXiv:2005.01643}, 2020.

\bibitem{jaques2019way}
Natasha Jaques, Asma Ghandeharioun, Judy~Hanwen Shen, Craig Ferguson, Agata
  Lapedriza, Noah Jones, Shixiang Gu, and Rosalind Picard.
\newblock Way off-policy batch deep reinforcement learning of implicit human
  preferences in dialog.
\newblock {\em arXiv preprint arXiv:1907.00456}, 2019.

\bibitem{precup2000eligibility}
Doina Precup.
\newblock Eligibility traces for off-policy policy evaluation.
\newblock {\em Computer Science Department Faculty Publication Series},
  page~80, 2000.

\bibitem{jiang2016doubly}
Nan Jiang and Lihong Li.
\newblock Doubly robust off-policy value evaluation for reinforcement learning.
\newblock In {\em International Conference on Machine Learning}, pages
  652--661. PMLR, 2016.

\bibitem{sutton2016emphatic}
Richard~S Sutton, A~Rupam Mahmood, and Martha White.
\newblock An emphatic approach to the problem of off-policy temporal-difference
  learning.
\newblock {\em The Journal of Machine Learning Research}, 17(1):2603--2631,
  2016.

\bibitem{levine2013guided}
Sergey Levine and Vladlen Koltun.
\newblock Guided policy search.
\newblock In {\em International conference on machine learning}, pages 1--9.
  PMLR, 2013.

\bibitem{sutton2009fast}
Richard~S Sutton, Hamid~Reza Maei, Doina Precup, Shalabh Bhatnagar, David
  Silver, Csaba Szepesv{\'a}ri, and Eric Wiewiora.
\newblock Fast gradient-descent methods for temporal-difference learning with
  linear function approximation.
\newblock In {\em Proceedings of the 26th annual international conference on
  machine learning}, pages 993--1000, 2009.

\bibitem{lagoudakis2003least}
Michail~G Lagoudakis and Ronald Parr.
\newblock Least-squares policy iteration.
\newblock {\em The Journal of Machine Learning Research}, 4:1107--1149, 2003.

\bibitem{nair2020accelerating}
Ashvin Nair, Murtaza Dalal, Abhishek Gupta, and Sergey Levine.
\newblock Accelerating online reinforcement learning with offline datasets.
\newblock {\em arXiv preprint arXiv:2006.09359}, 2020.

\bibitem{agarwal2020optimistic}
Rishabh Agarwal, Dale Schuurmans, and Mohammad Norouzi.
\newblock An optimistic perspective on offline reinforcement learning.
\newblock In {\em International Conference on Machine Learning}, pages
  104--114. PMLR, 2020.

\bibitem{kumar2020conservative}
Aviral Kumar, Aurick Zhou, George Tucker, and Sergey Levine.
\newblock Conservative q-learning for offline reinforcement learning.
\newblock {\em arXiv preprint arXiv:2006.04779}, 2020.

\bibitem{kidambi2020morel}
Rahul Kidambi, Aravind Rajeswaran, Praneeth Netrapalli, and Thorsten Joachims.
\newblock Morel: Model-based offline reinforcement learning.
\newblock {\em arXiv preprint arXiv:2005.05951}, 2020.

\bibitem{pan2009survey}
Sinno~Jialin Pan and Qiang Yang.
\newblock A survey on transfer learning.
\newblock {\em IEEE Transactions on knowledge and data engineering},
  22(10):1345--1359, 2009.

\bibitem{ruder2019neural}
Sebastian Ruder.
\newblock {\em Neural transfer learning for natural language processing}.
\newblock PhD thesis, NUI Galway, 2019.

\bibitem{taylor2009transfer}
Matthew~E Taylor and Peter Stone.
\newblock Transfer learning for reinforcement learning domains: A survey.
\newblock {\em Journal of Machine Learning Research}, 10(7), 2009.

\bibitem{da2019survey}
Felipe~Leno Da~Silva and Anna Helena~Reali Costa.
\newblock A survey on transfer learning for multiagent reinforcement learning
  systems.
\newblock {\em Journal of Artificial Intelligence Research}, 64:645--703, 2019.

\bibitem{silva2016transfer}
Felipe~Leno Silva and Anna Helena~Reali Costa.
\newblock Transfer learning for multiagent reinforcement learning systems.
\newblock In {\em Proceedings of the Twenty-Fifth International Joint
  Conference on Artificial Intelligence (IJCAI)}, pages 3982--3983. Springer,
  2016.

\bibitem{wooldridge2009introduction}
Michael Wooldridge.
\newblock {\em An introduction to multiagent systems}.
\newblock John wiley \& sons, 2009.

\bibitem{murray2018nlp}
Gabriel Murray, Giuseppe Carenini, and Shafiq Joty.
\newblock Nlp for conversations: Sentiment, summarization, and group dynamics.
\newblock In {\em Proceedings of the 27th International Conference on
  Computational Linguistics: Tutorial Abstracts}, pages 1--4, 2018.

\bibitem{panait2005cooperative}
Liviu Panait and Sean Luke.
\newblock Cooperative multi-agent learning: The state of the art.
\newblock {\em Autonomous agents and multi-agent systems}, 11(3):387--434,
  2005.

\bibitem{bansal2017emergent}
Trapit Bansal, Jakub Pachocki, Szymon Sidor, Ilya Sutskever, and Igor Mordatch.
\newblock Emergent complexity via multi-agent competition.
\newblock {\em arXiv preprint arXiv:1710.03748}, 2017.

\bibitem{lin2022online}
Baihan Lin, Djallel Bouneffouf, and Guillermo Cecchi.
\newblock Online learning in iterated prisoner's dilemma to mimic human
  behavior.
\newblock In {\em Pacific Rim International Conference on Artificial
  Intelligence}. Springer, 2022.

\bibitem{das2017learning}
Abhishek Das, Satwik Kottur, Jos{\'e}~MF Moura, Stefan Lee, and Dhruv Batra.
\newblock Learning cooperative visual dialog agents with deep reinforcement
  learning.
\newblock In {\em Proceedings of the IEEE international conference on computer
  vision}, pages 2951--2960, 2017.

\bibitem{mordatch2018emergence}
Igor Mordatch and Pieter Abbeel.
\newblock Emergence of grounded compositional language in multi-agent
  populations.
\newblock In {\em Proceedings of the AAAI Conference on Artificial
  Intelligence}, volume~32, 2018.

\bibitem{lin2023survey}
Baihan Lin, Djallel Bouneffouf, and Irina Rish.
\newblock A survey on compositional generalization in applications.
\newblock {\em arXiv preprint arXiv:2302.01067}, 2023.

\bibitem{kottur2017natural}
Satwik Kottur, Jos{\'e}~MF Moura, Stefan Lee, and Dhruv Batra.
\newblock Natural language does not emerge naturally in multi-agent dialog.
\newblock {\em arXiv preprint arXiv:1706.08502}, 2017.

\bibitem{walsh2003multi}
Michael Walsh, Robert Kelly, Greg~MP O'Hare, Julie Carson-Berndsen, and Tarek
  Abu-Amer.
\newblock A multi-agent computational linguistic approach to speech
  recognition.
\newblock In {\em The 18th International Joint Conference on Artificial
  Intelligence (IJCAI-03), 9th-15th August, Acapulco, Mexico, 2003}. IJCAI,
  2003.

\bibitem{nagoev2018model}
Zalimhan Nagoev, Larisa Lyutikova, and Irina Gurtueva.
\newblock Model for automatic speech recognition using multi-agent recursive
  cognitive architecture.
\newblock {\em Procedia computer science}, 145:386--392, 2018.

\bibitem{deb2014multi}
Kalyanmoy Deb.
\newblock Multi-objective optimization.
\newblock In {\em Search methodologies}, pages 403--449. Springer, 2014.

\bibitem{mcclure2004neural}
Samuel~M McClure, Michele~K York, and P~Read Montague.
\newblock The neural substrates of reward processing in humans: the modern role
  of {FMRI}.
\newblock {\em The Neuroscientist}, 10(3):260--268, 2004.

\bibitem{lee2012neural}
Daeyeol Lee, Hyojung Seo, and Min~Whan Jung.
\newblock Neural basis of reinforcement learning and decision making.
\newblock {\em Annual review of neuroscience}, 35:287, 2012.

\bibitem{maia2011reinforcement}
Tiago~V Maia and Michael~J Frank.
\newblock From reinforcement learning models to psychiatric and neurological
  disorders.
\newblock {\em Nature neuroscience}, 14(2):154--162, 2011.

\bibitem{lin2020astory}
Baihan Lin, Guillermo Cecchi, Djallel Bouneffouf, Jenna Reinen, and Irina Rish.
\newblock A story of two streams: Reinforcement learning models from human
  behavior and neuropsychiatry.
\newblock In {\em Proceedings of the Nineteenth International Conference on
  Autonomous Agents and Multi-Agent Systems, {AAMAS-20}}, pages 744--752,
  Auckland, New Zealand, 5 2020. International Foundation for Autonomous Agents
  and Multiagent Systems.

\bibitem{yang2020predicting}
Zhibo Yang, Lihan Huang, Yupei Chen, Zijun Wei, Seoyoung Ahn, Gregory Zelinsky,
  Dimitris Samaras, and Minh Hoai.
\newblock Predicting goal-directed human attention using inverse reinforcement
  learning.
\newblock In {\em Proceedings of the IEEE/CVF conference on computer vision and
  pattern recognition}, pages 193--202, 2020.

\bibitem{bouneffouf2020survey}
Djallel Bouneffouf, Irina Rish, and Charu Aggarwal.
\newblock Survey on applications of multi-armed and contextual bandits.
\newblock In {\em 2020 IEEE Congress on Evolutionary Computation (CEC)}, pages
  1--8. IEEE, 2020.

\bibitem{afsar2021reinforcement}
M~Mehdi Afsar, Trafford Crump, and Behrouz Far.
\newblock Reinforcement learning based recommender systems: A survey.
\newblock {\em ACM Computing Surveys (CSUR)}, 2021.

\bibitem{otter2020survey}
Daniel~W Otter, Julian~R Medina, and Jugal~K Kalita.
\newblock A survey of the usages of deep learning for natural language
  processing.
\newblock {\em IEEE transactions on neural networks and learning systems},
  32(2):604--624, 2020.

\bibitem{sharma2017literature}
Akanksha~Rai Sharma and Pranav Kaushik.
\newblock Literature survey of statistical, deep and reinforcement learning in
  natural language processing.
\newblock In {\em 2017 International Conference on Computing, Communication and
  Automation (ICCCA)}, pages 350--354. IEEE, 2017.

\bibitem{latif2020deep}
Siddique Latif, Rajib Rana, Sara Khalifa, Raja Jurdak, Junaid Qadir, and
  Bj{\"o}rn~W Schuller.
\newblock Deep representation learning in speech processing: Challenges, recent
  advances, and future trends.
\newblock {\em arXiv preprint arXiv:2001.00378}, 2020.

\end{thebibliography}

\end{document}